\documentclass[10pt,twocolumn,letterpaper]{article}

\usepackage[pagenumbers]{iccv} % To force page numbers, e.g. for an arXiv version

\def\paperID{14577}
\def\confName{ICCV}
\def\confYear{2025}

\def\paperTitle{IAP: Invisible Adversarial Patch Attack through Perceptibility-Aware Localization and Perturbation Optimization}

\def\authorBlock{
  \begin{minipage}[t]{0.45\linewidth}
    \centering
    Subrat Kishore Dutta \\
    CISPA Helmholtz Center for Information Security \\
    {\tt subrat.dutta@cispa.de}
  \end{minipage}
  \hfill
  \begin{minipage}[t]{0.45\linewidth}
    \centering
    Xiao Zhang \\
    CISPA Helmholtz Center for Information Security \\
    {\tt xiao.zhang@cispa.de}
  \end{minipage}
}

\newif\ifreview 
\newif\ifarxiv 
\newif\ifcamera 
\newif\ifrebuttal

\usepackage{graphicx}	
\usepackage{amsmath}	
\usepackage{amssymb}	
\usepackage{booktabs}
\usepackage{times}
\usepackage{microtype}
\usepackage{epsfig}
\usepackage{caption}
\usepackage{float}
\usepackage{placeins}
\usepackage{color, colortbl}
\usepackage{stfloats}
\usepackage{enumitem}
\usepackage{tabularx}
\usepackage{xstring}
\usepackage{multirow}
\usepackage{xspace}
\usepackage{url}
\usepackage{subcaption}
\usepackage{xcolor}
\usepackage{bm}
\usepackage{algorithm}
\usepackage{algorithmic}
\usepackage{setspace}
\usepackage[hang,flushmargin]{footmisc}
\usepackage{dsfont}
\usepackage{graphicx}
\ifcamera \usepackage[accsupp]{axessibility} \fi

% ...

\ifarxiv  \fi

\newcommand{\R}[1]{{%
    \textbf{%
        \ifstrequal{#1}{1}{\textcolor{red}{R#1}}{%
        \ifstrequal{#1}{2}{\textcolor{blue}{R#1}}{%
        \ifstrequal{#1}{3}{\textcolor{magenta}{R#1}}{%
        \ifstrequal{#1}{4}{\textcolor{teal}{R#1}}{%
                           \textcolor{cyan}{R#1}%
        }}}}%
    }%
}}

\DeclareMathOperator*{\argmax}{arg\,max}

\newcommand\shortsection[1]{\vspace{3pt}{\noindent\bf #1.}}

\usepackage[textsize=tiny]{todonotes}
\definecolor{ForestGreen}{cmyk}{0.864, 0.0, 0.429, 0.396}

\newcommand{\ifcomments}{\iftrue}

\definecolor{iccvblue}{rgb}{0.21,0.49,0.74}
\usepackage[pagebackref,breaklinks,colorlinks,allcolors=iccvblue]{hyperref}

\begin{document}

\title{\paperTitle}

\author{\authorBlock}

\maketitle

% \footnotetext[1]{This paper was accepted to ICCV 2025.}

\begin{abstract}
Despite modifying only a small localized input region, adversarial patches can drastically change the prediction of computer vision models. However, prior methods either cannot perform satisfactorily under targeted attack scenarios or fail to produce contextually coherent adversarial patches, causing them to be easily noticeable by human examiners and insufficiently stealthy against automatic patch defenses. In this paper, we introduce IAP, a novel attack framework that generates highly invisible adversarial patches based on perceptibility-aware localization and perturbation optimization schemes. Specifically, IAP first searches for a proper location to place the patch by leveraging classwise localization and sensitivity maps, balancing the susceptibility of patch location to both victim model prediction and human visual system, then employs a perceptibility-regularized adversarial loss and a gradient update rule that prioritizes color constancy for optimizing invisible perturbations. Comprehensive experiments across various image benchmarks and model architectures demonstrate that IAP consistently achieves competitive attack success rates in targeted settings with significantly improved patch invisibility compared to existing baselines. In addition to being highly imperceptible to humans, IAP is shown to be stealthy enough to render several state-of-the-art patch defenses ineffective.
% Additional evaluations suggest our method's adaptability to reduced patch size and black-box scenarios.
\end{abstract}

% We argue that current attack methods are not optimized for human imperceptibility and thus cannot bypass state-of-the-art patch defense techniques. 

\section{Introduction}
\label{sec:intro}

Deep neural networks (DNNs) are highly susceptible to inputs, often known as adversarial examples~\citep{szegedy2013intriguing,goodfellow2014explaining}, crafted with small perturbations designed to fool the model. Most existing attacks constrain perturbations using $\ell_p$-norm bounds to ensure imperceptibility~\citep{goodfellow2014explaining, moosavi2016deepfool, madry2017towards, carlini2017towards}, where any pixel of the entire input image is allowed to be modified to achieve the attack goal. Imposing an $\ell_p$-norm constraint limits perturbation size, ensuring generated perturbations remain visually imperceptible to humans. In contrast, a different but related line of work investigated adversarial patch attacks~\citep{brown2017adversarial, karmon2018lavan}, which modify localized regions without magnitude restrictions. The general framework of this approach facilitates adaptability to physical-world scenarios, increasing threats to security-critical machine learning systems.
Nevertheless, adversarial patches, though applicable in physical space, often attract suspicion due to their highly visible, salient nature and inability to maintain homogeneity with the host image. 

To overcome this, several techniques~\citep{eykholt2018robust, sharif2019general, liu2019perceptual, wang2023generating} have been proposed to generate adversarial patches that mimic realistic images, ensuring their placement does not raise suspicion. These approaches focus on maintaining context homogeneity and high realism, which are crucial for achieving imperceptibility to human perception. 
Despite these advancements, the perturbations required to sustain high attack efficacy, particularly in targeted scenarios, often result in large patches that are still visually noticeable. 
Defense mechanisms~\citep{liu2022segment, chen2022jujutsu, xu2023patchzero, tarchoun2023jedi, fu2024diffpad, kang2024diffender} have also been developed that leverage the contiguous high saliency of the patch region, proving successful in detecting adversarial patches.

The strict constraints on patch size and the need to avoid saliency make balancing stealth and targeted attack success particularly difficult. Therefore, a natural question arises:
\begin{center}
\emph{Are targeted attack goals at all achievable with visually imperceptible adversarial patches?}
\end{center}
In this paper, we answer the above question affirmatively by developing a general perceptibility-aware framework that can generate invisible patches with strong attack capabilities. Below, we summarize the key contributions of our work.

% Another line of research focused on developing methods to generate adversarial patches that are entirely invisible to humans~\citep{qian2020visually,bai2021inconspicuous,li2021generative}. 

% Most prior works either focus on untargeted attacks or face significant performance drops in targeted scenarios. While high imperceptibility is achievable in untargeted settings, targeted attacks remain notably more challenging and less explored. 

% Developed defense methods have increased the relevance of exploring adversarial patches in digital spaces. These defenses typically rely on complex, multi-stage preprocessing operations, often involving computationally intensive components like autoencoders~\cite{liu2022segment,fu2024diffpad,kang2024diffender,xu2023patchzero}. While effective, this makes them impractical for real-world applications with low-latency or resource-constrained environments, limiting their use to digital spaces. Additionally, in applications like face recognition, systems often process digital input along with real-time detection~\citep{mugalu2021face, wang2023privacy}, further reinforcing the need to focus on methods targeting the digital space while focusing on imperceptibility.

\begin{figure}[t]
  \centering
   \includegraphics[width=0.99\linewidth]{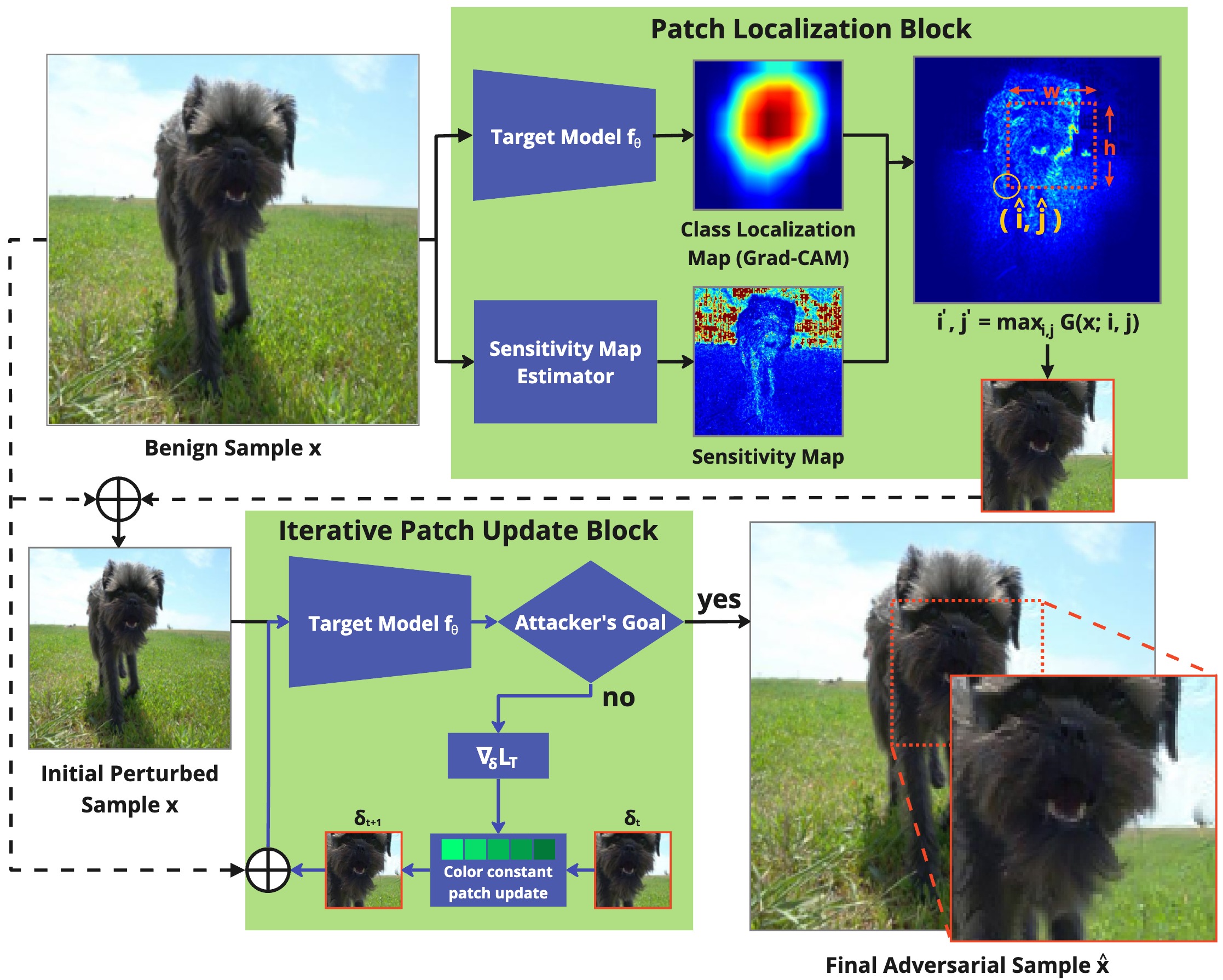}
    \vspace{-0.05in}
   \caption{The overall pipeline of IAP for conducting targeted attacks with imperceptible adversarial patches, consisting of both patch localization and iterative patch update blocks.}
   \vspace{-0.1in}
   \label{fig:flowchart}
\end{figure}

\shortsection{Contributions} We propose IAP\footnote{The code is available at \href{https://github.com/subratkishoredutta/IAP}{https://github.com/subratkishoredutta/IAP}.}, a novel patch attack framework based on a series of perceptibility-aware optimization schemes, which can realize targeted attack goals with \textbf{I}nvisible \textbf{A}dversarial \textbf{P}atches (see Figure \ref{fig:flowchart} for the overall pipeline). Most prior work overlooked human visual sensitivity in patch placement and perturbation, a key factor in our design's success. Specifically, IAP first locates a patch region in the host image that optimally balances the class localization and sensitivity scores, enabling a strategic advantage in both attack capabilities and imperceptibility (Section \ref{sec:patch placement}). After the localization step, it iteratively updates the patch by restricting the changes in base color and regularizing the adversarial loss using a human perception-based distance, which reduces the saliency of the resulting patch and further promotes imperceptibility (Section \ref{sec:perturbation opt}). 

Through extensive experiments on benchmark datasets for image classification and recognition, we show that IAP matches or surpasses state-of-the-art patch attacks in success rate while significantly enhancing invisibility across various targeted scenarios (Section \ref{sec:main attack capability}). Moreover, we demonstrate that adversarial patches generated by IAP can successfully bypass various patch defenses, confirming the stealthiness of our attack (Section \ref{Adefense}). The high stealthiness and strong capabilities of the adversarial patches generated by our method call for an urgent need to develop better defense strategies to detect and mitigate such stealthy attacks.

% To insert a figure: \input{figs/template}
% Or table: \input{tables/template}

\section{Related Work}
\label{sec:related}

\shortsection{Adversarial Patches}
\citet{brown2017adversarial} and \citet{karmon2018lavan} showed that small yet prominent adversarial patches can mislead classifiers, with the former introducing universal patches and the latter extending them to image-specific attacks. Later works focused on blending patches into their environment for stealth. \citet{eykholt2018robust} concealed perturbations in traffic signs, while \citet{sharif2016accessorize}, \citet{liu2019perceptual}, and \citet{li2021generative} leveraged generative models for adversarial eyeglasses, high-fidelity stickers, and GDPA-based soft-masked blending. \citet{zolfi2021translucent} used blending to generate translucent patches for detection tasks, and VRAP~\citep{wang2023generating} refined perturbations using $\ell_p$-norm constraints~\citep{madry2017towards}. Some aimed for complete invisibility, with \citet{bai2021inconspicuous} using cascaded GANs at high cost and \citet{qian2020visually} leveraging perceptual sensitivity but lacking spatial constraints. However, balancing stealth and attack efficacy remains a challenge, limiting most to untargeted attacks.

\shortsection{Imperceptibility in Adversarial Examples}
Imperceptibility in adversarial attacks ensures that adversarial examples closely resemble the original image, making them undetectable to humans. The human visual system is less sensitive to variations in high-textured regions than to perturbations on edges~\citep{liu2010just}. Studies have shown that perturbations in high-variance areas are less noticeable~\citep{luo2018towards}, while restricting modifications to horizontal and vertical edges enhances imperceptibility~\citep{croce2019sparse}. Additionally, altering only saturation and brightness to mimic different quantized levels of the same base color can effectively camouflage perturbations in textured regions~\citep{croce2019sparse,liu2010just,eckert1998perceptual}.
Common noise constraints, such as $\epsilon$-budgets and $\ell_p$-norm regularization, help limit perturbation magnitudes~\citep{szegedy2013intriguing, goodfellow2014explaining, moosavi2016deepfool, madry2017towards, eykholt2018robust}, but they do not account for human perception. To address this, \citet{luo2018towards} introduced a perception-aware distance metric that considers human sensitivity when curating perturbations. Given the space constraints of adversarial patch attacks, achieving targeted goals with limited perturbation is non-trivial. With our proposed method, IAP, we argue that human perception-oriented strategies, primarily studied in perturbation-restricted global attacks, are essential to store strategic large perturbations, which we hypothesize are crucial for stronger and stealthier targeted adversarial patch attacks.

\shortsection{Adversarial Examples in Face Recognition}
Adversarial attacks on facial recognition systems have shown that even subtle perturbations can undermine system reliability. \citet{goswami2018unravellingrobustnessdeeplearning} highlighted this with imperceptible attacks and proposed saliency-based defenses. \citet{sharif2016accessorize} demonstrated physical adversarial eyeglasses for identity spoofing. More recent works like \citet{Komkov_2021} introduced patch-based attacks that generalize across subjects. \citet{zolfi2022adversarialmaskrealworlduniversal} extended these attacks with adversarial masks, underscoring the practical threat such methods pose to biometric systems.

\shortsection{Patch Defenses} 
Recent adversarial patch defenses focus on detecting and mitigating high-saliency regions. Hayes~\cite{hayes2018visible} used saliency maps for patch localization and reconstruction, later enhanced by Jujutsu~\cite{chen2022jujutsu} with a two-stage GAN-based framework. Tarchoun et al.\cite{tarchoun2023jedi} improved localization via entropy analysis and autoencoder-based patch completion. Liu et al.\cite{liu2022segment} introduced SAC, leveraging U-Net-based segmenters and self-adversarial training, while PatchZero~\cite{xu2023patchzero} used pixel-level detection and adversarial training.
Adversarial purification, initially for global attacks, employed GANs like DefenseGAN~\cite{samangouei2018defenseganprotectingclassifiersadversarial}. With diffusion models~\cite{ho2020denoising,dhariwal2021diffusion}, GDMP~\cite{wang2022guided}, and DiffPure~\cite{nie2022diffusionmodelsadversarialpurification} introduced noise-based purification, while Xiao et al.\cite{xiao2023densepure} enhanced robustness through majority voting on generated reverse samples. However, these are optimized for $\ell_p$-norm perturbations, not patches. Kang et al.\cite{kang2024diffender} and Fu et al.~\cite{fu2024diffpad} adapted diffusion models for patch attacks, with DIFFender using text-guided localization and restoration, and DiffPAD combining super-resolution, binarization, and inpainting for decontamination.

\section{Problem Formulation and Motivation}
\label{sec:problem}

\subsection{Preliminaries}

We consider targeted, white-box settings for adversarial patch attacks. Assume the attacker has the full knowledge of a victim model $f_\theta$ with model parameters $\theta$. Let an RGB image $\bm{x}\in\mathcal{X} \subseteq \mathbb{R}^{W \times H \times C }$ be a correctly classified benign sample, $y\in\mathcal{Y}$ be the ground-truth class label of $\bm{x}$, and $y_\mathrm{targ}$ be the class label that the attacker aims to target for.
The adversarial input $\hat{\bm{x}}$ is generated by placing an adversarial patch of width $w$ and height $h$ on the benign sample $x$ at a certain localized region indexed by $(i,j)$ such that $f_\theta(\hat{\bm{x}}) = y_\mathrm{targ}$. More rigorously, $\hat{\bm{x}}$ is defined as:
\begin{equation}
\label{eq:1}
    \hat{\bm{x}} = (1 - \bm{m}) \odot \bm{x} + \bm{m} \odot \bm\delta,
\end{equation}
where $\bm{m} \in \{0,1\}^{W \times H \times C}$ is a location mask such that $m_{k,l,c} = 1$ if $i\leq k\leq i+w$, $j\leq l\leq j+h$ and $c\in C$, and $0$ otherwise, $\bm\delta \in \mathbb{R}^{W \times H \times C }$, and $\odot$ stands for the element-wise multiplication. 
For clarity, we use the following simplified notation to denote an adversarial input and the attached adversarial patch throughout the paper:
\begin{equation}
\label{eq:2}
    \hat{\bm{x}} = \bm{x} + _{i,j}\bm\delta,
\end{equation}
where $+_{i,j}\bm\delta$ means that we place the patch $\bm\delta$ at the pixel location $(i,j)$ with respect to the original input $\bm{x}$.

\subsection{Limitations of Existing Methods}

\shortsection{Lack of Attack Stealthiness}
% Early adversarial patch attacks expose the limitations of adversarial training on small perturbations and the lack of manual inspection, allowing overt yet effective attacks.
Initial works on adversarial patch attacks \cite{brown2017adversarial, karmon2018lavan} highlighted the limitations of adversarial training on small perturbations and the lack of manual inspection. These attacks enable visually overt yet effective attacks, but they overlook the patch's highly salient nature. 
This also applies to PS-GAN \cite{liu2019perceptual} and GDPA \cite{li2021generative} in their peak attack settings. Recent defenses \cite{tarchoun2023jedi,chen2022jujutsu,hayes2018visible,liu2022segment,fu2024diffpad,kang2024diffender} leverage this trait to detect and mitigate adversarial patches, rendering highly salient attacks ineffective (see Section \ref{Adefense}). Although these defenses are computationally intensive and primarily limited to digital spaces, they remain highly relevant given their applicability in security-critical systems, such as facial recognition, which operates extensively in digital environments~\cite{mugalu2021face,wang2023privacy}, thereby highlighting the importance of exploring attack methods within this domain. It is, therefore, essential to prioritize the design of adversarial attacks that generate inconspicuous non-salient patches capable of bypassing these systems.

\shortsection{Decreased Targeted Attack Performance}
Although works like PS-GAN \cite{liu2019perceptual} and GDPA \cite{li2021generative} in classification and \citep{zolfi2021translucent} detection tasks attempted to form the configuration where the saliency and conspicuousness of the applied sticker can be reduced, a trade-off between attack efficacy and imperceptibility is often observed. This also applies to targeted attack scenarios, where larger perturbations are often required to achieve good attack performance. As a result, prior works, focusing centrally on achieving complete invisibility, limit themselves to untargeted attack scenarios~\cite{qian2020visually,bai2021inconspicuous,wang2023generating}.

Contrary to the conventional approach of achieving imperceptibility by constraining perturbation magnitudes, we argue that unbounded perturbations are essential for realizing targeted adversarial goals. Instead, we advocate for strategically placing and curating large perturbations by leveraging insights from human visual perception, allowing substantial changes without compromising stealthiness or detectability. In Section \ref{sec:methodology}, we present a general framework that employs perceptibility-aware optimization to strategically curate large perturbations, enabling highly effective targeted attacks.

\section{IAP: Invisible Adversarial Patch Attack}
\label{sec:methodology}

This section explains the design of our attack, as illustrated in Figure \ref{fig:flowchart} and detailed in Algorithm \ref{alg:attack}. In particular, IAP first determines the optimal location of patch placement by considering attackability and imperceptibility to the human visual system (Section \ref{sec:patch placement}), and then optimizes the perturbations by minimizing a regularized targeted adversarial loss using color constant gradient updates (Section \ref{sec:perturbation opt}).

\subsection{Optimization of Patch Placement}
\label{sec:patch placement}

The central idea of this optimization step is to place the patch at a location that is highly vulnerable to adversarial perturbations and can host high perturbations, such that targeted goals can be achieved with adversarial patches without being visually salient. 
Previous studies have shown that models contain visually sensitive zones that play a significant role in their predictions \citep{zeiler2014visualizing, cao2015look}, making them more susceptible to attacks in these regions \citep{liu2019perceptual, qian2020visually}. 
Despite being vulnerable, these regions are observed to be highly sensitive to human observers as well, limiting their capacity to accommodate large perturbations. 
Nevertheless, it can be argued that a vulnerable location might require less perturbation for an attack to be successful, resulting in a less noticeable final patch, and hence desirable. 

On the contrary, regions with high variance, particularly those with intricate textures, can accommodate significantly large perturbations that remain imperceptible to humans~\citep{liu2010just}. These areas have been favored for global adversarial attacks~\citep{luo2018towards,croce2019sparse}. However, the attack region is restricted for adversarial patches, thereby limiting the attack's effectiveness. Consequently, a reliance solely on high perturbation levels, without strategic placement to enhance attack ability, may not yield a desirable result. Hence, our patch localization step is designed to arrive at an equilibrium that balances the vulnerability of the location and the capability to accommodate large perturbations without being visually salient.
More specifically, we propose a notion of perturbation priority index $G(\bm{x}; i,j)$ for any possible location $(i,j)$ with respect to the victim model $f$ and $(\bm{x},y)$, where the optimal location $(i',j')$ is determined based on $G(\bm{x}; i,j)$:
\begin{equation}
\begin{aligned}
\label{eq:3}
    &(i', j') = \argmax _{i,j}G(\bm{x}; i,j), \\
    &\quad \text{ where } G(\bm{x}; i,j) = \sum_{k=0}^{w}\sum_{l=0}^{h} \frac{J_y(\bm{x}; i+k,j+l)}{\mathrm{Sens}(\bm{x}; i+k,j+l)},
\end{aligned}
\end{equation}
where $h$ and $w$ represent the height and width of the patch, $J_y(\cdot, \cdot)$ denotes the class localization map capturing the model susceptibility with respect to the ground-truth label class $y$, and $\mathrm{Sens}(\cdot, \cdot)$ is the sensitivity map capturing the perturbation sensitivity to the human visual system. The perturbation priority metric aims to strike a balance between two aspects, seeking the most optimal location $(i,j)$, the window from which has the highest value for $G(\bm{x}; i,j)$ such that it facilitates attack capabilities, while also showing a high affinity for accommodating large perturbations.

\shortsection{Class Localization Map} To obtain the class localization map $J_{y}(\cdot, \cdot)$, we employ Grad-CAM \citep{selvaraju2017grad}. Since we consider white-box settings, we can directly employ the parameters of the victim model $f_\theta$ to obtain model-specific attention maps for the given input image $\bm{x}$. The computation process involves computing the gradient of the last fully connected layer's output, denoted as $g_{\theta}(\bm{x}, y)$, where $y$ is the ground-truth class of the benign input $\bm{x}$. Let $\mathbf{A}^k$ be the $k$-th feature map of the model's last convolution layer and $\alpha_k^y$ be its weight that characterizes the importance of the $k$-th feature map in predicting class label $y$. To be more specific, $\alpha_k^y$ is calculated by taking a global average pool over its calculated gradient as follows:
\begin{equation}
\label{eq:4}
    \alpha_k^y = \frac{1}{u\times v} \sum_{i=0}^{u} \sum_{j=0}^{v} \frac{\partial g_\theta(\bm{x},y)}{\partial A^k_{ij}},
\end{equation}
where $u$ and $v$ are the height and width of the feature map $A^k$. The final class localization map is calculated as the weighted sum of all feature maps followed by a ReLU function given by: for any $(i,j)$,
\begin{equation}
\label{eq:5}
    J_y(\bm{x}; i, j) = \mathrm{ReLU} \bigg( \sum_k \alpha^y_k \cdot 
A_{ij}^k \bigg).
\end{equation}

\shortsection{Sensitivity Map} 
Following prior works~\citep{luo2018towards,croce2019sparse}, we aim to position the patch in regions of high variance, while avoiding placement on object edges that are aligned with the coordinate axes. To ensure both factors when defining the sensitivity map $\mathrm{Sens}(\cdot,\cdot)$, we calculate the mean standard deviation of the pixel across the color channels along the horizontal and vertical axes, considering adjacent pixels, denoted as $\sigma_{ij}^{x}$ and $\sigma_{ij}^{y}$ respectively. Finally, the value of the sensitivity map at $(i,j)$ is computed as the reciprocal of the standard deviation given by:
\begin{equation}
\label{eq:6}
    \mathrm{Sens}(\bm{x}; i,j) = \frac{1}{\sigma_{ij}+\lambda}, \text{ where } \sigma_{ij}=\sqrt{\min(\sigma_{ij}^{x},\sigma_{ij}^{y})},
\end{equation}
% Finally, we consider $\sigma_{ij}=\sqrt{\min(\sigma_{ij}^{x},\sigma_{ij}^{y})}$ as the standard deviation value for the pixel at $(i,j)$. 
where $\lambda>0$ is a small value chosen to prevent division by zero. The sensitivity map induces a human perspective to the perturbation priority measure $G(\bm{x})$ in terms of perturbation sensitivity, such that those locations that cannot host large perturbations have a higher sensitivity than their counterparts. In the following, we use $J_y(\bm{x})$ and $\mathrm{Sens}(\bm{x})$ for the class localization and sensitivity maps of $\bm{x}$.

% =================================================
% ================== Our Algorithm ================
% =================================================

\begin{tiny}
\begin{algorithm*}[t]
\caption{Invisible Adversarial Patches (IAP)
}
\label{alg:attack}
\setstretch{1.1}
\begin{algorithmic}[1]
    \STATE \textbf{Input:} benign example $(\bm{x}, y)$, target class $y_{\mathrm{targ}}$, victim model $f_\theta$, and parameters $s,T,\eta, w, h$
    \STATE \quad $J_y(\bm{x}) \leftarrow$ compute the class localization map of $\bm{x}$ based on Equation \ref{eq:5}
    \STATE \quad $\mathrm{Sens}(\bm{x}) \leftarrow$ compute the sensitivity map of $\bm{x}$ based on Equation \ref{eq:6}
    \STATE \quad $(i', j') \leftarrow$ find the optimal patch location based on Equation \ref{eq:3}
    \STATE \quad $\bm{m} \leftarrow$ define the mask indexed by $(i',j')$ with patch size $w\times h$ 
    \STATE \quad Initialize $\bm{\delta}_0 \leftarrow \bm{x}$
    \STATE \quad \textbf{for} $t=0,1,\ldots T-1$ \textbf{do}
    \STATE \quad\quad \textbf{if} prediction confidence $f_{\theta}(y_{\mathrm{targ}}|\hat{\bm{x}}) \geq s$ \textbf{then}
    \STATE \quad\quad\quad \textbf{return} $\hat{\bm{x}}$
    \STATE \quad\quad \textbf{else}
    \STATE \quad\quad\quad $\mathcal{L}_T \leftarrow$ define the total adversarial loss function based on Equation \ref{eq:8}
    \STATE \quad\quad\quad $\bm\delta_{t+1} \leftarrow \bm\delta_t - \eta \cdot  \overline{\nabla_{\bm\delta}} \: \mathcal{L}_T(\bm\delta_t; \theta, \bm{x}, y) \odot \big(\bm\delta_{t} \oslash \mathrm{Sens}(\bm{x})\big)$
    \STATE \quad\quad\quad $\bm\delta_{t+1} \leftarrow \mathrm{clip}(\bm\delta_{t+1}, 0, 1)$
    \STATE \quad\quad $\hat{\bm{x}} \leftarrow \bm{x} + _{i', j'} \bm\delta_{t+1}$
    \STATE \textbf{Output:} $\hat{\bm{x}}$ 
\end{algorithmic}
\end{algorithm*}
\end{tiny}

\subsection{Optimization of Perturbation Update}
\label{sec:perturbation opt}

We start with initializing the patch with the original pixel values at the optimal location $(i', j')$ based on Equation \ref{eq:3}. We introduce a two-fold solution to integrate the human visual system into the perturbation crafting process. First, we introduce a regularization term to the adversarial loss to learn perturbations less salient to the human eye. Second, we utilize an update rule that considers human indifference to gray-level quantization as its basis to update the perturbation for visual advantage. In particular, we utilize the following distance metric, introduced in \cite{luo2018towards}, as the regularization term to penalize the visual distortion of $\hat{\bm{x}}$ with reference to $\bm{x}$:
\begin{equation}
\label{eq:7}
    D(\bm{x},\hat{\bm{x}}) = \frac{1}{h\times w}\sum_{k=i'}^{i'+w}\sum_{l=j'}^{j'+h} \mathrm{Sens}(\bm{x};k,l) \cdot \lvert x_{kl}-\hat{x}_{kl}\rvert,
\end{equation}
where $\mathrm{Sens}(\bm{x};k,l)$ is defined according to Equation \ref{eq:4}. $D(\bm{x},\hat{\bm{x}})$ incorporates the sensitivity of the human visual system in measuring the difference between the original and the adversarial example. We argue that employing such a distance metric as a regularization term in the final loss function will encourage the production of large perturbations at locations where the sensitivity of the human vision is limited while suppressing the perturbations in highly sensitive locations. Accordingly, we can achieve large perturbations favoring the attack while maintaining the overall insensitivity in appearance. 

The central part of the loss function remains consistent with existing attacks, comprising two cross-entropy loss terms with respect to the target class $y_{\mathrm{targ}}$ and the ground-truth class $y$, respectively. Specifically, the final adversarial loss objective that we aim to minimize is given by:
\begin{equation}
\begin{aligned}
\label{eq:8}
    \mathcal{L}_T(\bm\delta; &\theta, \bm{x}, y) = 
      w_1 \cdot \mathcal{L}_{\mathrm{CE}}(\hat{\bm{x}},y_{\mathrm{targ}}; \theta) \\
      &\quad - w_2 \cdot \mathcal{L}_{\mathrm{CE}}(\hat{\bm{x}},y;\theta)+ w_3 \cdot D(\bm{x},\hat{\bm{x}}),
\end{aligned}
\end{equation}
where $w_1, w_2$, and $w_3$ are weight parameters that regulate the contribution for each term and can be adjusted based on the preference of the attack.

Moreover, the loss function is accompanied by an update rule through which we aim to achieve two main objectives. Motivated by the fact that humans are highly indifferent to changes in brightness and saturation levels of the same base color, which is analogous to the behavior with lower quantization levels, we update the perturbation such that the base color of the pixel does not alter. Next, we aim to maximize the utility of gradient magnitude information of the loss function with respect to the input, while adhering to color constraints. Such a strategy is designed to minimize the number of iterations required for the attack. To be more specific, we propose the following gradient update rule: for $t=0,1,\ldots, T-1,$
\begin{align}
\label{eq:9}
    \bm\delta_{t+1} = \bm\delta_{t} - \eta \cdot \overline{\nabla_{\bm\delta}} \: \mathcal{L}_T(\bm\delta_t; \theta, \bm{x}, y) \odot \Big(\bm\delta_{t} \oslash \mathrm{Sens}(\bm{x})\Big), 
\end{align}
where $\odot$ (resp., $\oslash$) stands for element-wise multiplication (resp., division), $\overline{\nabla_{\bm\delta}}$ denotes the averaged gradient of the loss function $\mathcal{L}_T$ over the three color channels, and $\eta$ is the step size. Averaging over the three channels ensures that each pixel channel is updated by the same amount, thereby ensuring that the base color is not changed. We note that \cite{croce2019sparse} proposed a similar update rule, but their method does not enforce any color constraint. our proposed regularized loss and update rule contributes significantly to realizing targeted goals with imperceptible adversarial patches.

\section{Experiments}
\label{sec:experiment}

In this section, we evaluate the performance of IAP on image classification and face recognition tasks. We compare IAP against state-of-the-art attack methods, such as Google Patch \citep{brown2017adversarial}, LaVAN \citep{karmon2018lavan}, GDPA \citep{li2021generative}, and Masked Projected Gradient Descent (MPGD), an extension of standard PGD~\citep{madry2017towards}. 
Additionally, we assess its effectiveness against defense methods specifically designed for adversarial patch attacks~\citep{hayes2018visible,chen2022jujutsu,liu2022segment,tarchoun2023jedi,fu2024diffpad,kang2024diffender}. For GDPA, we set the visibility parameter $\alpha$ to $0.4$, and for MPGD, we use an $l_{\infty}$ perturbation bound of $\epsilon = 16/255$.

\subsection{Experimental Setup}

\shortsection{Dataset \& Configuration} 
We test IAP on a subset of the ILSVRC 2012 validation set~\citep{russakovsky2015imagenet} and VGG Face test set~\citep{Parkhi15,li2021generative}, respectively, for the image classification and face recognition task. We consider four target architectures: ResNet-50~\citep{he2016deep}, VGG16~\citep{simonyan2014very}, Swin Transformer Tiny, and Swin Transformer Base~\citep{liu2021swin}. For ImageNet, we use pre-trained weights, while for VGG Face, we re-train the models on its training set. All images are resized to $224\times224$ before the attack. We optimize the patch until the target class confidence reaches $0.9$ or up to $1000$ iterations with a patch size of $84\times84$, covering $14\%$ of the image. Although square patches are adopted following prior works, our attack framework supports arbitrary shapes. Upon failing, we reinitialize the step size up to $3$ times.
On average, IAP converges in 379 iterations (including possible reinitializations) and takes approximately 19 seconds per sample when targeting the Swin Transformer Base model using an NVIDIA A100 GPU. Most of the runtime overhead comes from patch localization, which can be reduced by increasing the stride of the sliding window used in the search.

\shortsection{Evaluation Metric}
We measure the performance of different methods based on attack success rate (ASR), which characterizes the ratio of instances that can be successfully attacked. Let $\mathcal{A}$ be the evaluated attack, $f_\theta$ be the victim model, and $\mathcal{S}$ be a test set of correctly classified images. The ASR of $\mathcal{A}$ with respect to $f_\theta$ and $\mathcal{S}$ is defined as:

\begin{equation}
\label{eq:10}
    \mathrm{ASR}(\mathcal{A}; f_\theta, \mathcal{S}) = \frac{1}{|\mathcal{S}|}\sum_{\bm{x} \in \mathcal{S}} \mathds{1}\big(f_\theta(\hat{\bm{x}}) = y_{\mathrm{targ}}\big),
\end{equation}
where $|\mathcal{S}|$ denotes the cardinality of $\mathcal{S}$, and $\hat{\bm{x}}$ is the adversarial example generated by $\mathcal{A}$ for $\bm{x}$.
In our evaluations, we assess patch imperceptibility using both traditional statistical methods and CNN-based measures. Traditional methods include SSIM~\citep{wang2004image}, UIQ~\citep{wang2002universal}, and SRE~\citep{lanaras2018super}, which evaluate structural similarity, image distortion, and error relative to signal power, respectively. CNN-based methods include CLIPScore~\citep{hessel2021clipscore} and LPIPS~\citep{zhang2018unreasonable}, which quantify perceptual similarity by capturing subtle visual features using pre-trained deep neural networks. Similarity is measured on two scales: globally, by comparing entire images, and locally, by focusing on the attacked region. We provide additional details on our experimental setup in Appendix \ref{appendix:detailed experimental settings}.

\begin{table*}[t]
    \centering
    \resizebox{\linewidth}{!}{
    \begin{tabular}{c|c|ccc|ccc|ccc|ccc}
        \toprule
        \midrule
        \multirow{2}{*}{\textbf{Dataset}} & \multirow{2}{*}{\textbf{Method}} & \multicolumn{3}{c|}{\textbf{ResNet-50}} & \multicolumn{3}{c|}{\textbf{VGG 16}} & \multicolumn{3}{c|}{\textbf{Swin Transformer Tiny}} & \multicolumn{3}{c}{\textbf{Swin Transformer Base}}\\
        \cmidrule{3-14}
        {} & {} &\textbf{ASR} & \textbf{LPIPS$_{L}$($\downarrow$)} & \textbf{SSIM$_{L}$($\uparrow$)}& \textbf{ASR} & \textbf{LPIPS$_{L}$($\downarrow$)} & \textbf{SSIM$_{L}$($\uparrow$)} & \textbf{ASR} & \textbf{LPIPS$_{L}$($\downarrow$)} & \textbf{SSIM$_{L}$($\uparrow$)} & \textbf{ASR} & \textbf{LPIPS$_{L}$($\downarrow$)} & \textbf{SSIM$_{L}$($\uparrow$)}\\ 
        
        \midrule
        \midrule
        
        \multirow{5}{*}{\textbf{ImageNet}} & Google Patch &  99.10 & 0.74 & 0.010 & 100.0 & 0.76 & 0.002 & 99.80 & 0.77 & 0.002 & 97.9 & 0.77 & 0.003\\

        {} & LaVAN &  100.0 & 0.78 & 0.010 & 93.60 & 0.79 & 0.002 & 99.70 & 0.78 & 0.005 & 100.0 & 0.78 & 0.004\\
        
        {} & GDPA &  93.70 & 0.57 & 0.350 & 89.20 & 0.61 & 0.310 & 83.70 & 0.54 & 0.390 & 85.10 & 0.54 & 0.360\\

        {} & MPGD &  97.80 & 0.24 & 0.790 & 96.50 & 0.32 & 0.810 & 98.80 & 0.19 & 0.800 & 70.50 & 0.20 & 0.800\\

        \cmidrule{2-14}
        {} & IAP &  99.50 & 0.12 & 0.940 & 99.10 & 0.23 & 0.900 & 99.60 & 0.06 & 0.980 & 99.40 & 0.07 & 0.970 \\
        \midrule
        \midrule
        \multirow{5}{*}{\textbf{VGG Face}} & Google Patch &  $97.73 \pm 1.56$ & $0.75 \pm 0.03$ & $0.01 \pm 0.00$ & $99.97 \pm 0.05$ & $0.87 \pm 0.01$ & $0.00 \pm 0.00$ & $99.13 \pm 0.17$ & $0.81 \pm 0.02$ & $0.01 \pm 0.02$ & $97.90 \pm 0.50$ & $0.89 \pm 0.04$ & $0.00 \pm 0.00$\\

        {} & LaVAN &  $99.00 \pm 1.41$ & $0.81 \pm 0.04$ & $0.01 \pm 0.00$ & $99.83 \pm 0.24$ & $0.86 \pm 0.01$ & $0.00 \pm 0.00$ & $100.0 \pm 0.00$ & $0.85 \pm 0.00$ & $0.01 \pm 0.00$ & $99.70 \pm 0.29$ & $0.85 \pm 0.00$ & $0.00 \pm 0.00$\\
        
        {} & GDPA &  $99.07 \pm 0.76$ & $0.62 \pm 0.03$ & $0.31 \pm 0.02$ & $95.71 \pm 3.28$ & $0.62 \pm 0.07$ & $0.31 \pm 0.12$ & $95.10 \pm 3.47$ & $0.61 \pm 0.03$ & $0.33 \pm 0.01$ & $72.41 \pm 12.6$ & $0.63 \pm 0.06$ & $0.29 \pm 0.10$\\
        
        {} & MPGD &  $67.11 \pm 10.8$ & $0.38 \pm 0.00$ & $0.61 \pm 0.00$ & $86.90 \pm 1.61$ & $0.42 \pm 0.02$ & $0.65 \pm 0.02$ & $95.52 \pm 0.54$ & $0.37 \pm 0.01$ & $0.64 \pm 0.00$ & $91.20 \pm 7.45$ & $0.38 \pm 0.01$ & $0.61 \pm 0.01$\\

        \cmidrule{2-14}
        {} & IAP &  $94.53 \pm 3.06$ & $0.21 \pm 0.03$ & $0.90 \pm 0.01$ & $99.44 \pm 0.49$ & $0.21 \pm 0.01$ & $0.92 \pm 0.01$ & $99.07 \pm 0.33$ & $0.26 \pm 0.01$ & $0.86 \pm 0.00$ & $98.20 \pm 0.86$ & $0.28 \pm 0.02$ & $0.86 \pm 0.02$ \\
        \midrule
        \bottomrule
    \end{tabular}
    }
    \vspace{-0.05in}
    \caption{Comparisons of ASR ($\%$) and imperceptibility between different adversarial patch attacks on VGG Face. For MPGD, we consider perturbations bounded by $\epsilon=16/255$ in $\ell_\infty$-norm. The subscripts $_L$ and $_G$ represent the imperceptibility measures at local and global scales, respectively. Note that a lower LPIPS score indicates the generated adversarial patches are less perceptible.}
    \vspace{-0.05in}
  \label{table: comparison main}
\end{table*}

\begin{figure}[t]
    \centering
    \begin{subfigure}[b]{1.0\linewidth}
        \centering
        \includegraphics[width=\linewidth]{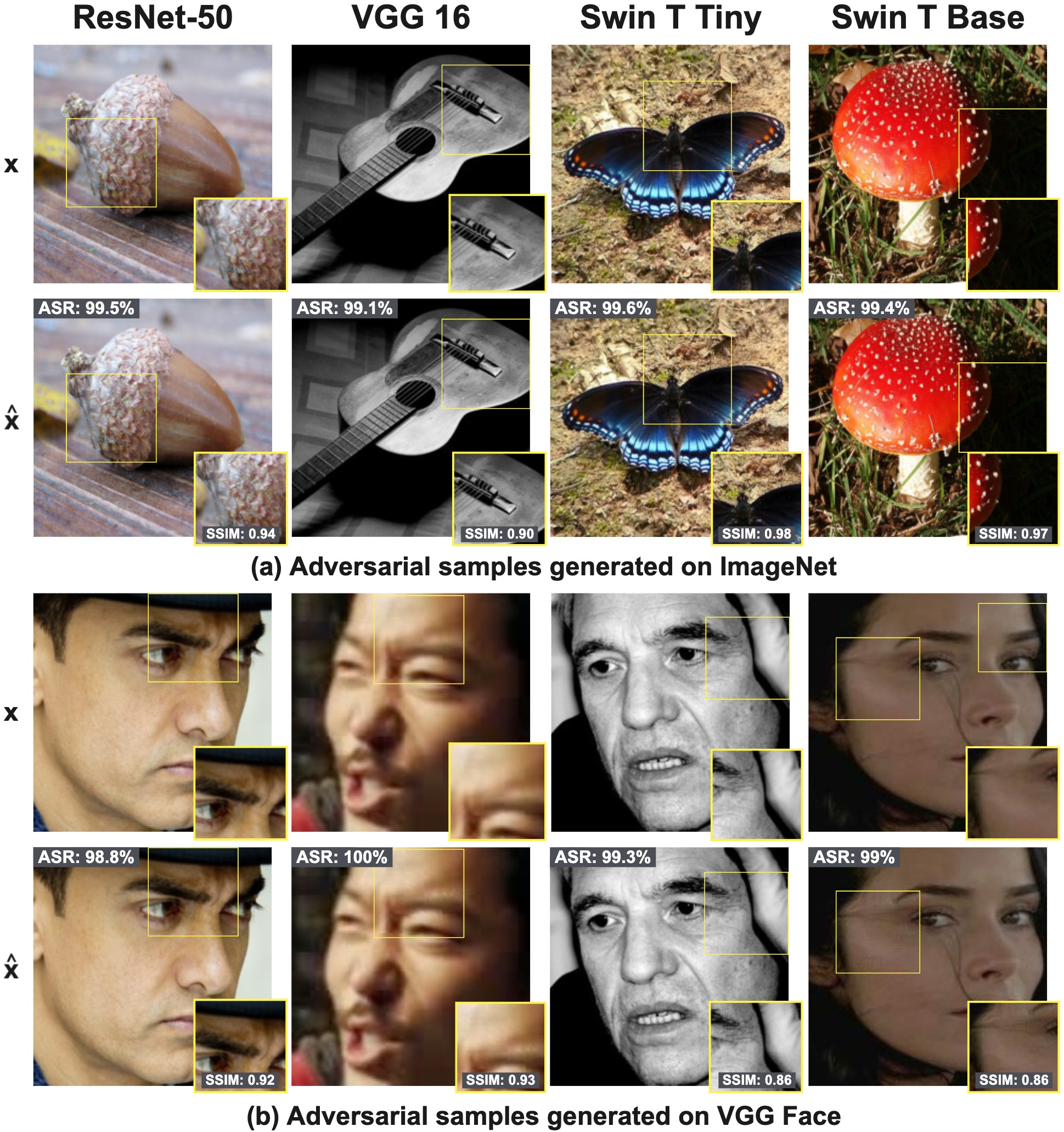}
    \end{subfigure}
    \vspace{-0.2in}
    \caption{Visualizations of original images (\textbf{$x$}) and their adversarial counterparts (\textbf{$\hat{x}$}) generated by IAP. The smaller images in the bottom-right corner indicate the optimal location $(i', j')$.}
    \label{fig:2}
    \vspace{-0.1in}
\end{figure}

\subsection{Main Results}
\label{sec:main attack capability}

We compare the performance of IAP with existing state-of-the-art adversarial patch attacks, including Google Patch, LAVAN, GDPA, and MPGD, on ImageNet for the classification tasks and VGG Face for the face recognition tasks. Table \ref{table: comparison main} presents a comprehensive comparison of attack success rates and patch imperceptibility across both the ImageNet and VGG Face datasets.
For simplicity, we only present the most widely adopted LPIPS and SSIM scores measured at the local scale in Table \ref{table: comparison main}, but defer the comparison of other imperceptibility metrics corresponding to different victim models in  Appendix \ref{appendix:additional experiments}, where similar trends on imperceptibility have been observed. 
For ImageNet, we choose ``Toaster'' as the target class that the adversary aims to trick the model predictions into, while for VGG Face, we repeat the attack $3$ times for $3$ different target classes, ``A. J. Buckley'', ``Aamir Khan'' and ``Aaron Staton'', and report the averaged statistics of our metrics. For both datasets, we fix the patch size as $84\times 84$, which covers $14\%$ of the total image. Due to space limits, we provide more detailed attack evaluation results of our method in Appendix \ref{appendix:additional results on Imagenet}, and \ref{appendix:additional results on VGG Face}.

Table \ref{table: comparison main} shows that the attack success rates achieved by IAP are comparable to or even higher than those of other state-of-the-art patch attacks. Considering all successful attack instances, the average target class prediction confidence is $ 0.84 \pm 0.03 $ for ImageNet and $0.86 \pm 0.01$ for VGG Face, aligned with our expectations since the confidence threshold is set as $s=0.9$. As for imperceptibility, it achieves significantly lower LPIPS scores and higher SSIM values compared to other methods. The achievement of the best performance in imperceptibility while maintaining high attack success rates validates the superiority of IAP in realizing targeted goals with imperceptible adversarial patches.

\begin{figure}[t]
    \centering
    \begin{subfigure}[b]{1.0\linewidth}
    \hspace{-0.25cm}
        \centering
        \includegraphics[width=\linewidth]{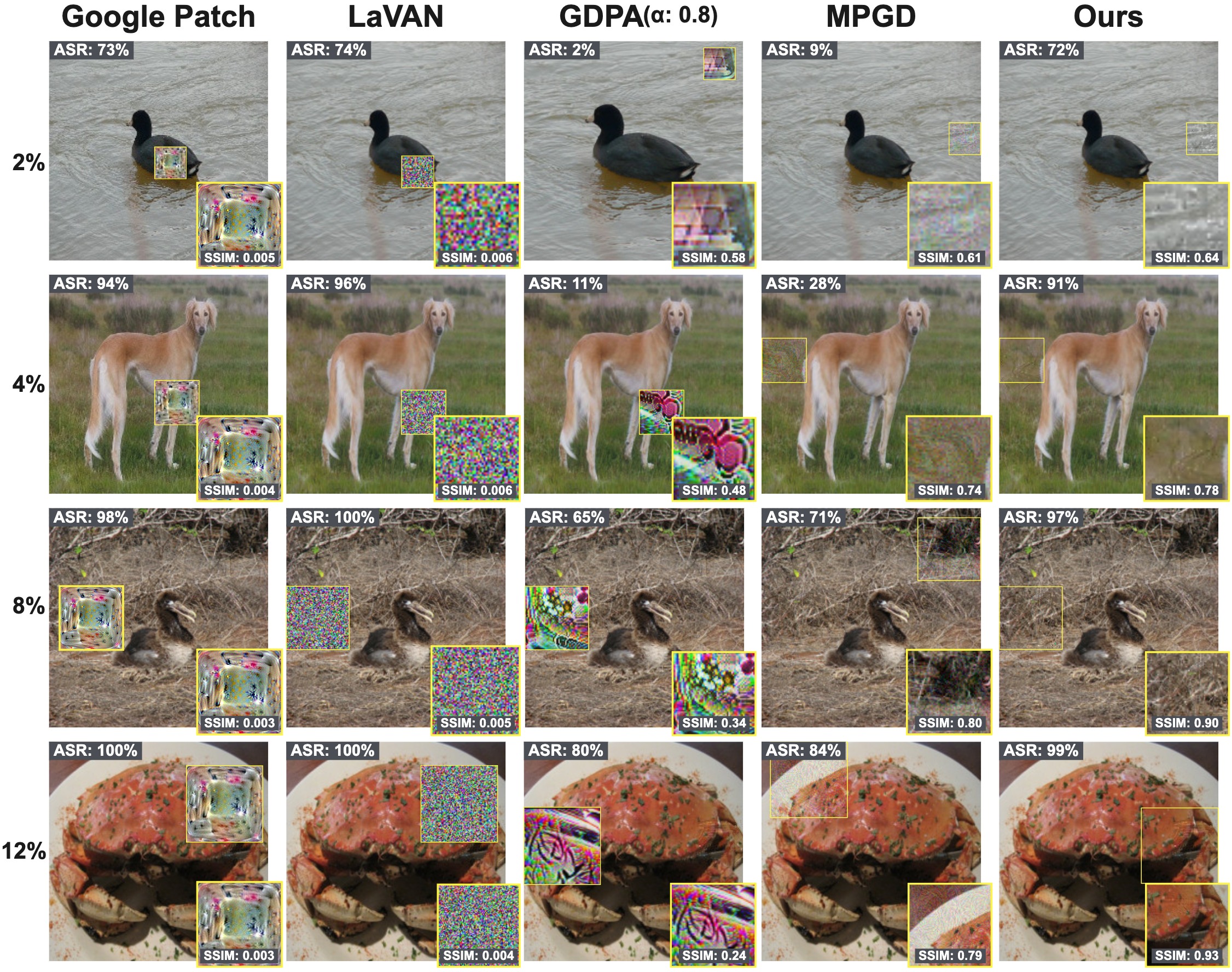}
    \end{subfigure}
    \vspace{-0.15in}
    \caption{The visualizations presented illustrate the adversarial samples generated by various methods at different patch sizes (expressed in \%). The corresponding ASR and SSIM$_L$ values for each method and patch size are shown. The smaller images in the lower-right corner represent the attack surface.}
    \label{fig:3}
    \vspace{-0.1in}
\end{figure}

\shortsection{Visualization} To illustrate the imperceptibility of our adversarial patches, we visualize the final adversarial samples $\hat{\bm{x}}$ alongside their original counterparts $\bm{x}$ in Figure \ref{fig:2} using randomly selected ImageNet and VGG Face images. Note that all adversarial samples are misclassified into the target label with prediction confidence above $0.9$. Recognizing that prior methods achieve high ASR even with smaller patches \citep{karmon2018lavan,brown2017adversarial}, we conducted evaluations across varying patch sizes to ensure a comprehensive comparison. As shown in Figure \ref{fig:3}, IAP not only matches or exceeds the ASR of competing approaches at all patch sizes but offers significantly superior imperceptibility. Additionally, to highlight IAP’s shape-agnostic design, we achieve $99.2\%$ ASR with $\mathrm{LPIPS}_L$ of $0.085$ using a circular patch covering $11\%$ of the host image (see Appendix \ref{append:flexibility in patch shape} for more details).

\subsection{Human Perceptibility Study}
\begin{figure}[t]
    \centering
    \begin{subfigure}[b]{\linewidth}
        \centering
        \includegraphics[width=\linewidth]{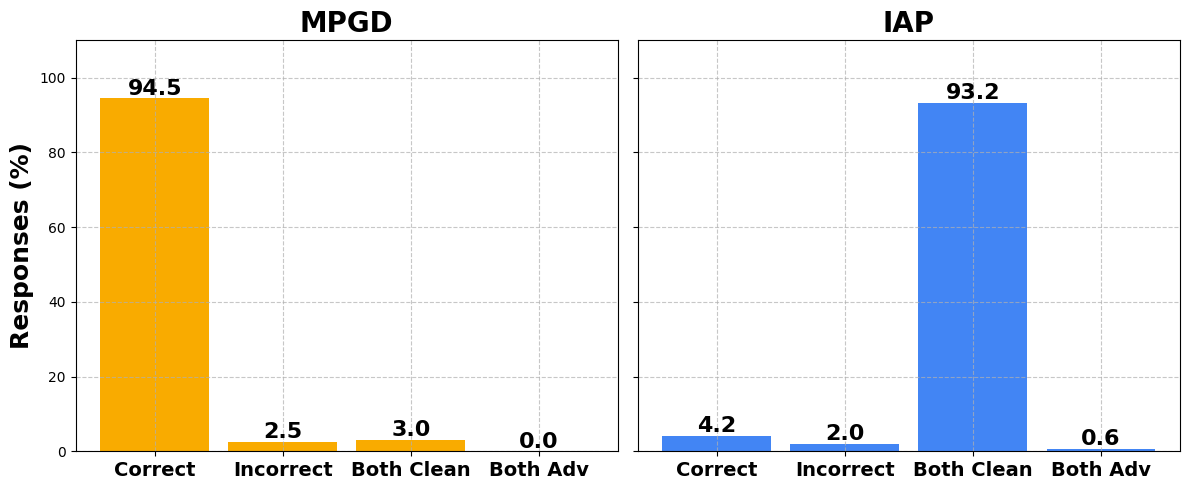}
    \end{subfigure}
    \vspace{-0.2in}
    \caption{Human perceptability study. ``Correct'' means correct selection, and ``Both Clean'' means considering both images clean.}
    \label{fig:humanstudy}
    \vspace{-0.05in}
\end{figure}

To assess how noticeable IAP-generated patches are to human observers, we conduct a user study with $28$ participants, all having an ML background and $82\%$ familiar with adversarial examples. Each participant was shown $20$ adversarial-clean image pairs ($10$ from MPGD, $10$ from IAP) and asked to choose from four options: (A) Left is adversarial, (B) Right is adversarial, (C) Both look clean, or (D) Both look adversarial. Options A and B assess detectability, while C and D reflect perceptual ambiguity. Figure \ref{fig:humanstudy} shows that IAP patches are highly imperceptible, with an average detection rate of just $4.2\%$, compared to $94.5\%$ for MPGD. Few users selected “Both adversarial” for IAP samples, which we attribute to accumulated suspicion over repeated comparisons.

\begin{figure}[t]
    \centering
    \begin{subfigure}[b]{1\linewidth}
        % \hspace{-0.15cm}
        \centering
        \includegraphics[width=1.0\textwidth]{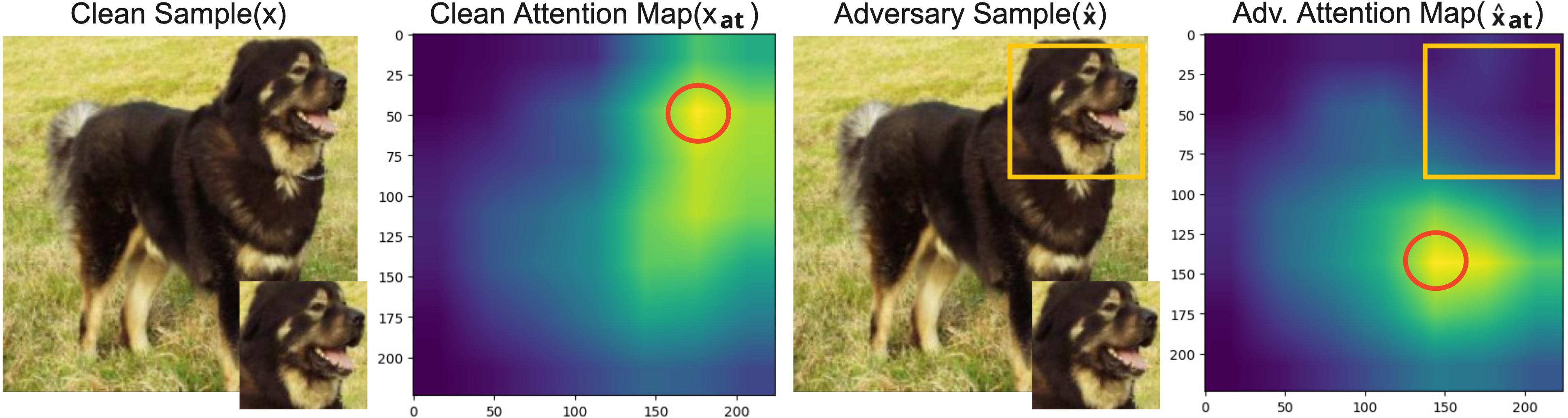}
    \end{subfigure}
    \vspace{-0.15in}
    \caption{Illustration of attention map changes from Grad-CAM for the clean sample $x$  and adversarial sample $\hat{x}$. The red circle highlights the highest attention region, while the yellow square represents the attack surface. Unlike \citep{brown2017adversarial}, IAP does not shift attention to the attack region.}
    \label{fig:4}
     \vspace{-0.1in}
\end{figure}

\subsection{Attack Stealthiness}
\label{Adefense}

Brown et al.\cite{brown2017adversarial} argued that adversarial patches are inherently salient, drawing a classifier’s attention and leading to misclassifications. Although this claim was found inconsistent by Karmon et al.\cite{karmon2018lavan}, it still remains a foundational assumption for many defense methods that localize patches based on saliency cues~\citep{hayes2018visible,chen2022jujutsu,xiang2021patchguardefficientprovableattack}. We tested whether this claim holds for IAP’s non-salient patches. Using Grad-CAM, we checked whether the point of maximum attention overlaps the patch. As illustrated in Figure \ref{fig:4}, they typically do not. Evaluating across multiple architectures, approximately $70 \%$ of IAP samples show no overlap between the highest-attention region and the attack surface (see Table \ref{table: gradatt} in Appendix \ref{append:GradCAM analysis of attention overlap}). This demonstrates the enhanced stealthiness of our approach.

\shortsection{Against Patch Defense} 
To evaluate our proposed method's robustness, we examine its ability to evade existing defense mechanisms. We assess our attack against six defense methods tailored for thwarting adversarial patch attacks: Jedi~\citep{tarchoun2023jedi}, Jujutsu~\citep{chen2022jujutsu}, SAC~\citep{liu2022segment}, DW~\citep{hayes2018visible}, DIFFender~\citep{kang2024diffender}, and DiffPAD~\citep{fu2024diffpad}. Operating under a white-box setting, we assume the attacker can access the underlying model parameters to launch the patch attack. 
In this task, we focus on ImageNet and employ ResNet-50 as the target model. Since patch defenses typically include a patch detection module as an initial step, we generate the adversarial samples on the target model for the target class ``toaster'' beforehand. 

We define an attack as successful if the adversarial sample, after going through the defense module, induces the target model to classify it as the target class correctly. Table \ref{table: defence comparison} demonstrates the effectiveness of different patch attacks in the presence of defenses. 
In all $6$ tested scenarios, our generated adversarial samples can bypass the defense module effectively with high attack success rates. In stark contrast, all the other attacks are ineffective against at least one of the evaluated defenses. For instance, IAP can achieve $100\%$ ASR against SAC, whereas the best performance attained by all the remaining patch attacks is as low as $11.6\%$. 

\begin{figure}[t]
    \centering
    \begin{subfigure}[b]{\linewidth}
        \centering
        \includegraphics[width=0.98\textwidth]{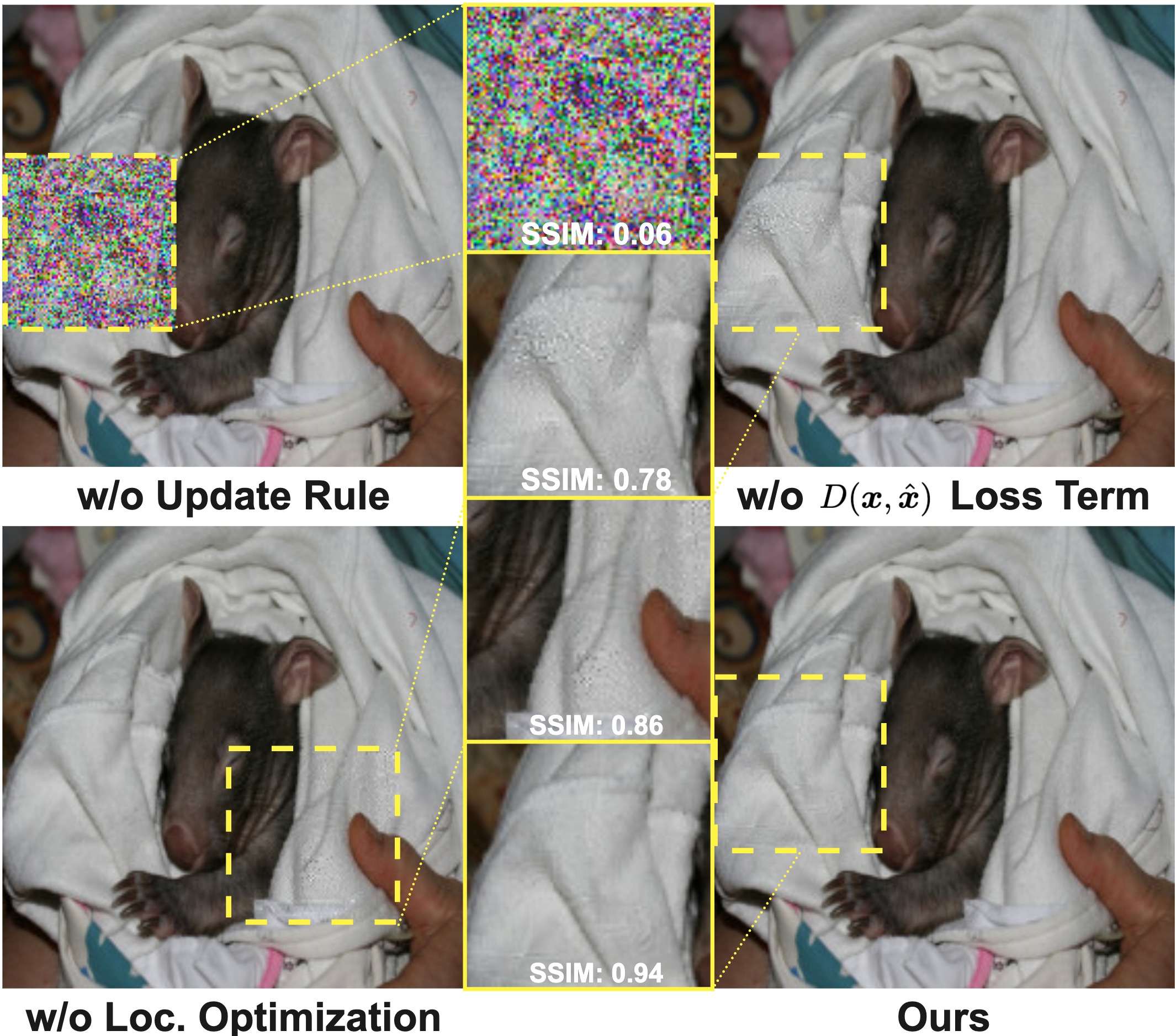}
    \end{subfigure}
    \vspace{-0.15in}
    \caption{Ablation study on the impact of IAP's components.}
    \label{fig:5}
    % \vspace{-0.1in}
\end{figure}

\begin{table}[t]
    \centering
    \resizebox{\linewidth}{!}{
    \begin{tabular}{lccccccc}
        \toprule
        \midrule
        \textbf{Method} & \textbf{Jedi} & \textbf{Jujutsu} & \textbf{SAC} & \textbf{DW} & \textbf{DIFFender} & \textbf{DiffPAD} \\
        \midrule
        \midrule
        Google Patch & $46.8$ & $0.0$ & $2.7$ & $1.4$ & $35.5$ & $33.2$ \\
        LaVAN & $50.9$ & $0.3$ & $3.8$ & $54.0$ & $53.2$ & $39.8$ \\
        GDPA & $67.1$ & $94.0$ & $7.4$ & $1.3$ & $57.0$ & $52.1$\\
         MPGD & $68.2$ & $95.1$ & $11.6$ & $79.0$ & $95.7$ & $92.1$\\
        \midrule
        IAP & $78.6$ & $99.8$ & $100$ & $89.8$ & $99.8$ & $98.6$\\
        \midrule
        \bottomrule
    \end{tabular}
    }
  \vspace{-0.1in}
  \caption{Comparisons of ASR ($\%)$ between different attack methods against various patch defenses.}
  \label{table: defence comparison}
  \vspace{-0.1in}
\end{table}

\section{Further Analyses}
\label{sec:further analysis}

\subsection{Black-Box \& Real-World Applicability}

To evaluate IAP’s generalizability beyond white-box settings, we assess both transferability and black-box performance. Adversarial patches generated on a surrogate model (e.g., ResNet-50) transfer reasonably well to unseen models like VGG and SqueezeNet, influenced by architectural similarity (see Figure \ref{fig:transfer} in Appendix \ref{append:transferability}). For true black-box adaptation, we approximate Grad-CAM using the surrogate and apply NES-based query optimization \cite{ilyas2018blackboxadversarialattackslimited}, achieving strong untargeted ASR with 12,000 queries on ImageNet (Table \ref{table: bbox}). Additionally, without specific adaptation, IAP demonstrates promising real-world physical attack abilities, achieving a 70\% average ASR on printed patches tested across images of varying viewpoints and object classes (see Figure \ref{fig:physical}), following PS-GAN's setup. Additional details are provided in Sections \ref{append:transferability}, \ref{append:black-box adaptation}, and \ref{append:physical} in the appendix.

\begin{table}[t]
    % \centering
    % \small
    \resizebox{\linewidth}{!}{
    \begin{tabular}{lccccc}
        \toprule
        {\textbf{Model}} & {\textbf{MobileNetV2}}& {\textbf{SqueezeNet}} & {\textbf{VGG13}} & {\textbf{DenseNet121}}& {\textbf{EfficientNet}}\\ 
        \midrule
        {\textbf{ASR}} & {$95\%$} & {$94\%$} & {$97\%$} & {$95\%$} & {$89\%$}\\
        {\textbf{LPIPS$_L$}} & {$0.24$} & {$0.22$} & {$0.27$} & {$0.25$} & {$0.26$}\\
        \bottomrule
    \end{tabular}}
\vspace{-0.1in}
\caption{Performance of IAP under black-box attack scenarios.}
\vspace{-0.1in}
\label{table: bbox}
\end{table}

\subsection{Ablation Studies}
We conduct ablation studies to evaluate the impact of key components in IAP, including patch size, the regularization coefficient $w_3$ in the human-oriented distance metric (Equation \ref{eq:7}), part of the total loss function (Equation \ref{eq:8}), and finally the update rule. Larger patch sizes significantly improve both attack efficacy and imperceptibility, increasing $w_3$ improves imperceptibility at the cost of slightly reduced attack success abilities, but excessively high values destabilize optimization and degrade both. Our update rule offers significantly better imperceptibility than Adam, with a minimal reduction in attack success rate. Detailed findings are elaborated in Appendix \ref{appendix:ablation studies}. Additionally, Figure \ref{fig:5} shows the impact of some of the key components of our attack methodology on the attack stealthiness, highlight the trade-offs and strengths of IAP’s design choices.

\section{Conclusion}
\label{sec:conclusion}

We introduced IAP, a perceptibility-aware optimization framework for generating imperceptible adversarial patches. Experimental results across various computer vision tasks demonstrated our method's superiority in maintaining high stealth and strong targeted attack efficacy. Beyond the white-box scenario, IAP also showed promise, both in black-box transferability and in the physical attack domain.

\shortsection{Limitation and Future Work}
Despite its strengths, IAP has several limitations. First, it does not account for local pixel context during perturbation updates; thus, there is potential for individual pixels to become unnaturally bright or dark relative to their neighbors, reducing imperceptibility. Second, while the perceptibility-aware patch placement improves stealth, it introduces additional computational overhead, making the attack slower. Future work could explore more efficient alternatives that preserve effectiveness. Consistent with other adversarial patch methods, its effectiveness diminishes for smaller patch sizes. While IAP generalizes across several tasks and domains, including physical and black-box scenarios, further adaptation is required to ensure robustness in fully black-box, query-limited, or physical-world settings. Finally, we believe that developing defense strategies that align machine perception with human vision~\cite{hua2021humanimperceptibleattacksapplications, geirhos2022imagenettrainedcnnsbiasedtexture} is crucial for improving generalizability and potentially mitigating invisible adversarial patches.

\newpage
\clearpage

{\small
\bibliographystyle{ieeenat_fullname}
\bibliography{11_references}

\begin{thebibliography}{55}
\providecommand{\natexlab}[1]{#1}
\providecommand{\url}[1]{\texttt{#1}}
\expandafter\ifx\csname urlstyle\endcsname\relax
  \providecommand{\doi}[1]{doi: #1}\else
  \providecommand{\doi}{doi: \begingroup \urlstyle{rm}\Url}\fi

\bibitem[Bai et~al.(2021)Bai, Luo, and Zhao]{bai2021inconspicuous}
Tao Bai, Jinqi Luo, and Jun Zhao.
\newblock Inconspicuous adversarial patches for fooling image-recognition systems on mobile devices.
\newblock \emph{IEEE Internet of Things Journal}, 9\penalty0 (12):\penalty0 9515--9524, 2021.

\bibitem[Brown et~al.(2017)Brown, Man{\'e}, Roy, Abadi, and Gilmer]{brown2017adversarial}
Tom~B Brown, Dandelion Man{\'e}, Aurko Roy, Mart{\'\i}n Abadi, and Justin Gilmer.
\newblock Adversarial patch.
\newblock \emph{arXiv preprint arXiv:1712.09665}, 2017.

\bibitem[Cao et~al.(2015)Cao, Liu, Yang, Yu, Wang, Wang, Huang, Wang, Huang, Xu, et~al.]{cao2015look}
Chunshui Cao, Xianming Liu, Yi Yang, Yinan Yu, Jiang Wang, Zilei Wang, Yongzhen Huang, Liang Wang, Chang Huang, Wei Xu, et~al.
\newblock Look and think twice: Capturing top-down visual attention with feedback convolutional neural networks.
\newblock In \emph{Proceedings of the IEEE international conference on computer vision}, pages 2956--2964, 2015.

\bibitem[Carlini and Wagner(2017)]{carlini2017towards}
Nicholas Carlini and David Wagner.
\newblock Towards evaluating the robustness of neural networks.
\newblock In \emph{2017 ieee symposium on security and privacy (sp)}, pages 39--57. Ieee, 2017.

\bibitem[Chen et~al.(2022)Chen, Dash, and Pattabiraman]{chen2022jujutsu}
Zitao Chen, Pritam Dash, and Karthik Pattabiraman.
\newblock Jujutsu: A two-stage defense against adversarial patch attacks on deep neural networks, 2022.

\bibitem[Croce and Hein(2019)]{croce2019sparse}
Francesco Croce and Matthias Hein.
\newblock Sparse and imperceivable adversarial attacks.
\newblock In \emph{Proceedings of the IEEE/CVF international conference on computer vision}, pages 4724--4732, 2019.

\bibitem[Dhariwal and Nichol(2021)]{dhariwal2021diffusion}
Prafulla Dhariwal and Alexander Nichol.
\newblock Diffusion models beat gans on image synthesis.
\newblock \emph{Advances in neural information processing systems}, 34:\penalty0 8780--8794, 2021.

\bibitem[Eckert and Bradley(1998)]{eckert1998perceptual}
Michael~P Eckert and Andrew~P Bradley.
\newblock Perceptual quality metrics applied to still image compression.
\newblock \emph{Signal processing}, 70\penalty0 (3):\penalty0 177--200, 1998.

\bibitem[Eykholt et~al.(2018)Eykholt, Evtimov, Fernandes, Li, Rahmati, Xiao, Prakash, Kohno, and Song]{eykholt2018robust}
Kevin Eykholt, Ivan Evtimov, Earlence Fernandes, Bo Li, Amir Rahmati, Chaowei Xiao, Atul Prakash, Tadayoshi Kohno, and Dawn Song.
\newblock Robust physical-world attacks on deep learning visual classification.
\newblock In \emph{Proceedings of the IEEE conference on computer vision and pattern recognition}, pages 1625--1634, 2018.

\bibitem[Fu et~al.(2024)Fu, Zhang, Pashami, Rahimian, and Holst]{fu2024diffpad}
Jia Fu, Xiao Zhang, Sepideh Pashami, Fatemeh Rahimian, and Anders Holst.
\newblock Diffpad: Denoising diffusion-based adversarial patch decontamination.
\newblock \emph{arXiv preprint arXiv:2410.24006}, 2024.

\bibitem[Geirhos et~al.(2022)Geirhos, Rubisch, Michaelis, Bethge, Wichmann, and Brendel]{geirhos2022imagenettrainedcnnsbiasedtexture}
Robert Geirhos, Patricia Rubisch, Claudio Michaelis, Matthias Bethge, Felix~A. Wichmann, and Wieland Brendel.
\newblock Imagenet-trained cnns are biased towards texture; increasing shape bias improves accuracy and robustness, 2022.

\bibitem[Goodfellow et~al.(2014)Goodfellow, Shlens, and Szegedy]{goodfellow2014explaining}
Ian~J Goodfellow, Jonathon Shlens, and Christian Szegedy.
\newblock Explaining and harnessing adversarial examples.
\newblock \emph{arXiv preprint arXiv:1412.6572}, 2014.

\bibitem[Goswami et~al.(2018)Goswami, Ratha, Agarwal, Singh, and Vatsa]{goswami2018unravellingrobustnessdeeplearning}
Gaurav Goswami, Nalini Ratha, Akshay Agarwal, Richa Singh, and Mayank Vatsa.
\newblock Unravelling robustness of deep learning based face recognition against adversarial attacks, 2018.

\bibitem[Hayes(2018)]{hayes2018visible}
Jamie Hayes.
\newblock On visible adversarial perturbations \& digital watermarking.
\newblock In \emph{Proceedings of the IEEE Conference on Computer Vision and Pattern Recognition Workshops}, pages 1597--1604, 2018.

\bibitem[He et~al.(2016)He, Zhang, Ren, and Sun]{he2016deep}
Kaiming He, Xiangyu Zhang, Shaoqing Ren, and Jian Sun.
\newblock Deep residual learning for image recognition.
\newblock In \emph{Proceedings of the IEEE Conference on Computer Vision and Pattern Recognition (CVPR)}, pages 770--778, 2016.

\bibitem[Hessel et~al.(2021)Hessel, Holtzman, Forbes, Bras, and Choi]{hessel2021clipscore}
Jack Hessel, Ari Holtzman, Maxwell Forbes, Ronan~Le Bras, and Yejin Choi.
\newblock Clipscore: A reference-free evaluation metric for image captioning.
\newblock \emph{arXiv preprint arXiv:2104.08718}, 2021.

\bibitem[Ho et~al.(2020)Ho, Jain, and Abbeel]{ho2020denoising}
Jonathan Ho, Ajay Jain, and Pieter Abbeel.
\newblock Denoising diffusion probabilistic models.
\newblock \emph{Advances in neural information processing systems}, 33:\penalty0 6840--6851, 2020.

\bibitem[Hua et~al.(2021)Hua, Xu, Blanchet, and Nguyen]{hua2021humanimperceptibleattacksapplications}
Xinru Hua, Huanzhong Xu, Jose Blanchet, and Viet Nguyen.
\newblock Human imperceptible attacks and applications to improve fairness, 2021.

\bibitem[Ilyas et~al.(2018)Ilyas, Engstrom, Athalye, and Lin]{ilyas2018blackboxadversarialattackslimited}
Andrew Ilyas, Logan Engstrom, Anish Athalye, and Jessy Lin.
\newblock Black-box adversarial attacks with limited queries and information, 2018.

\bibitem[Kang et~al.(2024)Kang, Dong, Wang, Ruan, Chen, Su, and Wei]{kang2024diffender}
Caixin Kang, Yinpeng Dong, Zhengyi Wang, Shouwei Ruan, Yubo Chen, Hang Su, and Xingxing Wei.
\newblock Diffender: Diffusion-based adversarial defense against patch attacks.
\newblock In \emph{European Conference on Computer Vision}, pages 130--147. Springer, 2024.

\bibitem[Karmon et~al.(2018)Karmon, Zoran, and Goldberg]{karmon2018lavan}
Danny Karmon, Daniel Zoran, and Yoav Goldberg.
\newblock Lavan: Localized and visible adversarial noise.
\newblock In \emph{International Conference on Machine Learning}, pages 2507--2515. PMLR, 2018.

\bibitem[Komkov and Petiushko(2021)]{Komkov_2021}
Stepan Komkov and Aleksandr Petiushko.
\newblock Advhat: Real-world adversarial attack on arcface face id system.
\newblock In \emph{2020 25th International Conference on Pattern Recognition (ICPR)}, page 819–826. IEEE, 2021.

\bibitem[Lanaras et~al.(2018)Lanaras, Bioucas-Dias, Galliani, Baltsavias, and Schindler]{lanaras2018super}
Charis Lanaras, Jos{\'e} Bioucas-Dias, Silvano Galliani, Emmanuel Baltsavias, and Konrad Schindler.
\newblock Super-resolution of sentinel-2 images: Learning a globally applicable deep neural network.
\newblock \emph{ISPRS Journal of Photogrammetry and Remote Sensing}, 146:\penalty0 305--319, 2018.

\bibitem[Li and Ji(2021)]{li2021generative}
Xiang Li and Shihao Ji.
\newblock Generative dynamic patch attack.
\newblock \emph{arXiv preprint arXiv:2111.04266}, 2021.

\bibitem[Liu et~al.(2010)Liu, Lin, Paul, Deng, and Zhang]{liu2010just}
Anmin Liu, Weisi Lin, Manoranjan Paul, Chenwei Deng, and Fan Zhang.
\newblock Just noticeable difference for images with decomposition model for separating edge and textured regions.
\newblock \emph{IEEE Transactions on Circuits and Systems for Video Technology}, 20\penalty0 (11):\penalty0 1648--1652, 2010.

\bibitem[Liu et~al.(2019)Liu, Liu, Fan, Ma, Zhang, Xie, and Tao]{liu2019perceptual}
Aishan Liu, Xianglong Liu, Jiaxin Fan, Yuqing Ma, Anlan Zhang, Huiyuan Xie, and Dacheng Tao.
\newblock Perceptual-sensitive gan for generating adversarial patches.
\newblock In \emph{Proceedings of the AAAI conference on artificial intelligence}, pages 1028--1035, 2019.

\bibitem[Liu et~al.(2022)Liu, Levine, Lau, Chellappa, and Feizi]{liu2022segment}
Jiang Liu, Alexander Levine, Chun~Pong Lau, Rama Chellappa, and Soheil Feizi.
\newblock Segment and complete: Defending object detectors against adversarial patch attacks with robust patch detection.
\newblock In \emph{Proceedings of the IEEE/CVF Conference on Computer Vision and Pattern Recognition}, pages 14973--14982, 2022.

\bibitem[Liu et~al.(2021)Liu, Lin, Cao, Hu, Wei, Zhang, Lin, and Guo]{liu2021swin}
Ze Liu, Yutong Lin, Yue Cao, Han Hu, Yixuan Wei, Zheng Zhang, Stephen Lin, and Baining Guo.
\newblock Swin transformer: Hierarchical vision transformer using shifted windows.
\newblock In \emph{Proceedings of the IEEE/CVF International Conference on Computer Vision (ICCV)}, pages 10012--10022, 2021.

\bibitem[Luo et~al.(2018)Luo, Liu, Wei, and Xu]{luo2018towards}
Bo Luo, Yannan Liu, Lingxiao Wei, and Qiang Xu.
\newblock Towards imperceptible and robust adversarial example attacks against neural networks.
\newblock In \emph{Proceedings of the AAAI Conference on Artificial Intelligence}, 2018.

\bibitem[Madry et~al.(2017)Madry, Makelov, Schmidt, Tsipras, and Vladu]{madry2017towards}
Aleksander Madry, Aleksandar Makelov, Ludwig Schmidt, Dimitris Tsipras, and Adrian Vladu.
\newblock Towards deep learning models resistant to adversarial attacks.
\newblock \emph{arXiv preprint arXiv:1706.06083}, 2017.

\bibitem[Moosavi-Dezfooli et~al.(2016)Moosavi-Dezfooli, Fawzi, and Frossard]{moosavi2016deepfool}
Seyed-Mohsen Moosavi-Dezfooli, Alhussein Fawzi, and Pascal Frossard.
\newblock Deepfool: a simple and accurate method to fool deep neural networks.
\newblock In \emph{Proceedings of the IEEE conference on computer vision and pattern recognition}, pages 2574--2582, 2016.

\bibitem[Mugalu et~al.(2021)Mugalu, Wamala, Serugunda, and Katumba]{mugalu2021face}
Ben~Wycliff Mugalu, Rodrick~Calvin Wamala, Jonathan Serugunda, and Andrew Katumba.
\newblock Face recognition as a method of authentication in a web-based system.
\newblock \emph{arXiv preprint arXiv:2103.15144}, 2021.

\bibitem[Nie et~al.(2022)Nie, Guo, Huang, Xiao, Vahdat, and Anandkumar]{nie2022diffusionmodelsadversarialpurification}
Weili Nie, Brandon Guo, Yujia Huang, Chaowei Xiao, Arash Vahdat, and Anima Anandkumar.
\newblock Diffusion models for adversarial purification, 2022.

\bibitem[Parkhi et~al.(2015)Parkhi, Vedaldi, and Zisserman]{Parkhi15}
Omkar~M. Parkhi, Andrea Vedaldi, and Andrew Zisserman.
\newblock Deep face recognition.
\newblock In \emph{British Machine Vision Conference}, 2015.

\bibitem[Qian et~al.(2020)Qian, Wang, Wang, Zeng, Gu, Ji, and Swaileh]{qian2020visually}
Yaguan Qian, Jiamin Wang, Bin Wang, Shaoning Zeng, Zhaoquan Gu, Shouling Ji, and Wassim Swaileh.
\newblock Visually imperceptible adversarial patch attacks on digital images.
\newblock \emph{arXiv preprint arXiv:2012.00909}, 2020.

\bibitem[Russakovsky et~al.(2015)Russakovsky, Deng, Su, Krause, Satheesh, Ma, Huang, Karpathy, Khosla, Bernstein, Berg, and Fei-Fei]{russakovsky2015imagenet}
Olga Russakovsky, Jia Deng, Hao Su, Jonathan Krause, Sanjeev Satheesh, Sean Ma, Zhiheng Huang, Andrej Karpathy, Aditya Khosla, Michael Bernstein, Alexander~C. Berg, and Li Fei-Fei.
\newblock Imagenet large scale visual recognition challenge, 2015.

\bibitem[Samangouei et~al.(2018)Samangouei, Kabkab, and Chellappa]{samangouei2018defenseganprotectingclassifiersadversarial}
Pouya Samangouei, Maya Kabkab, and Rama Chellappa.
\newblock Defense-gan: Protecting classifiers against adversarial attacks using generative models, 2018.

\bibitem[Selvaraju et~al.(2017)Selvaraju, Cogswell, Das, Vedantam, Parikh, and Batra]{selvaraju2017grad}
Ramprasaath~R Selvaraju, Michael Cogswell, Abhishek Das, Ramakrishna Vedantam, Devi Parikh, and Dhruv Batra.
\newblock Grad-cam: Visual explanations from deep networks via gradient-based localization.
\newblock In \emph{Proceedings of the IEEE international conference on computer vision}, pages 618--626, 2017.

\bibitem[Sharif et~al.(2016)Sharif, Bhagavatula, Bauer, and Reiter]{sharif2016accessorize}
Mahmood Sharif, Sruti Bhagavatula, Lujo Bauer, and Michael~K Reiter.
\newblock Accessorize to a crime: Real and stealthy attacks on state-of-the-art face recognition.
\newblock In \emph{Proceedings of the 2016 acm sigsac conference on computer and communications security}, pages 1528--1540, 2016.

\bibitem[Sharif et~al.(2019)Sharif, Bhagavatula, Bauer, and Reiter]{sharif2019general}
Mahmood Sharif, Sruti Bhagavatula, Lujo Bauer, and Michael~K Reiter.
\newblock A general framework for adversarial examples with objectives.
\newblock \emph{ACM Transactions on Privacy and Security (TOPS)}, 22\penalty0 (3):\penalty0 1--30, 2019.

\bibitem[Simonyan and Zisserman(2014)]{simonyan2014very}
Karen Simonyan and Andrew Zisserman.
\newblock Very deep convolutional networks for large-scale image recognition.
\newblock \emph{arXiv preprint arXiv:1409.1556}, 2014.

\bibitem[Szegedy et~al.(2013)Szegedy, Zaremba, Sutskever, Bruna, Erhan, Goodfellow, and Fergus]{szegedy2013intriguing}
Christian Szegedy, Wojciech Zaremba, Ilya Sutskever, Joan Bruna, Dumitru Erhan, Ian Goodfellow, and Rob Fergus.
\newblock Intriguing properties of neural networks.
\newblock \emph{arXiv preprint arXiv:1312.6199}, 2013.

\bibitem[Tarchoun et~al.(2023)Tarchoun, Ben~Khalifa, Mahjoub, Abu-Ghazaleh, and Alouani]{tarchoun2023jedi}
Bilel Tarchoun, Anouar Ben~Khalifa, Mohamed~Ali Mahjoub, Nael Abu-Ghazaleh, and Ihsen Alouani.
\newblock Jedi: entropy-based localization and removal of adversarial patches.
\newblock In \emph{Proceedings of the IEEE/CVF Conference on Computer Vision and Pattern Recognition}, pages 4087--4095, 2023.

\bibitem[Wang et~al.(2022)Wang, Lyu, Lin, Dai, and Fu]{wang2022guided}
Jinyi Wang, Zhaoyang Lyu, Dahua Lin, Bo Dai, and Hongfei Fu.
\newblock Guided diffusion model for adversarial purification.
\newblock \emph{arXiv preprint arXiv:2205.14969}, 2022.

\bibitem[Wang and Wang(2023)]{wang2023generating}
Xiaosen Wang and Kunyu Wang.
\newblock Generating visually realistic adversarial patch.
\newblock \emph{arXiv preprint arXiv:2312.03030}, 2023.

\bibitem[Wang and Bovik(2002)]{wang2002universal}
Zhou Wang and Alan~C Bovik.
\newblock A universal image quality index.
\newblock \emph{IEEE signal processing letters}, 9\penalty0 (3):\penalty0 81--84, 2002.

\bibitem[Wang et~al.(2004)Wang, Bovik, Sheikh, and Simoncelli]{wang2004image}
Zhou Wang, Alan~C Bovik, Hamid~R Sheikh, and Eero~P Simoncelli.
\newblock Image quality assessment: from error visibility to structural similarity.
\newblock \emph{IEEE transactions on image processing}, 13\penalty0 (4):\penalty0 600--612, 2004.

\bibitem[Wang et~al.(2023)Wang, Wang, Jin, Zhang, Hu, Wang, Sun, Yuan, Liu, and Ren]{wang2023privacy}
Zhibo Wang, He Wang, Shuaifan Jin, Wenwen Zhang, Jiahui Hu, Yan Wang, Peng Sun, Wei Yuan, Kaixin Liu, and Kui Ren.
\newblock Privacy-preserving adversarial facial features.
\newblock In \emph{Proceedings of the IEEE/CVF Conference on Computer Vision and Pattern Recognition}, pages 8212--8221, 2023.

\bibitem[Xiang and Mittal(2021)]{xiang2021patchguardefficientprovableattack}
Chong Xiang and Prateek Mittal.
\newblock Patchguard++: Efficient provable attack detection against adversarial patches, 2021.

\bibitem[Xiao et~al.(2023)Xiao, Chen, Jin, Wang, Nie, Liu, Anandkumar, Li, and Song]{xiao2023densepure}
Chaowei Xiao, Zhongzhu Chen, Kun Jin, Jiongxiao Wang, Weili Nie, Mingyan Liu, Anima Anandkumar, Bo Li, and Dawn Song.
\newblock Densepure: Understanding diffusion models for adversarial robustness.
\newblock In \emph{The Eleventh International Conference on Learning Representations}, 2023.

\bibitem[Xu et~al.(2023)Xu, Xiao, Zheng, Cai, and Nevatia]{xu2023patchzero}
Ke Xu, Yao Xiao, Zhaoheng Zheng, Kaijie Cai, and Ram Nevatia.
\newblock Patchzero: Defending against adversarial patch attacks by detecting and zeroing the patch.
\newblock In \emph{Proceedings of the IEEE/CVF Winter Conference on Applications of Computer Vision}, pages 4632--4641, 2023.

\bibitem[Zeiler and Fergus(2014)]{zeiler2014visualizing}
Matthew~D Zeiler and Rob Fergus.
\newblock Visualizing and understanding convolutional networks.
\newblock In \emph{Computer Vision--ECCV 2014: 13th European Conference, Zurich, Switzerland, September 6-12, 2014, Proceedings, Part I 13}, pages 818--833. Springer, 2014.

\bibitem[Zhang et~al.(2018)Zhang, Isola, Efros, Shechtman, and Wang]{zhang2018unreasonable}
Richard Zhang, Phillip Isola, Alexei~A Efros, Eli Shechtman, and Oliver Wang.
\newblock The unreasonable effectiveness of deep features as a perceptual metric.
\newblock In \emph{Proceedings of the IEEE conference on computer vision and pattern recognition}, pages 586--595, 2018.

\bibitem[Zolfi et~al.(2021)Zolfi, Kravchik, Elovici, and Shabtai]{zolfi2021translucent}
Alon Zolfi, Moshe Kravchik, Yuval Elovici, and Asaf Shabtai.
\newblock The translucent patch: A physical and universal attack on object detectors.
\newblock In \emph{Proceedings of the IEEE/CVF conference on computer vision and pattern recognition}, pages 15232--15241, 2021.

\bibitem[Zolfi et~al.(2022)Zolfi, Avidan, Elovici, and Shabtai]{zolfi2022adversarialmaskrealworlduniversal}
Alon Zolfi, Shai Avidan, Yuval Elovici, and Asaf Shabtai.
\newblock Adversarial mask: Real-world universal adversarial attack on face recognition model, 2022.

\end{thebibliography}
}

\newpage
\clearpage

\appendix 
% \section{Appendix}
% \label{sec:appendix_section}

\section{Detailed Experimental Settings}
\label{appendix:detailed experimental settings}

In this section, we give more details on the setup of our experiments. We evaluate the performance of our adversarial patch attack on image classification and face recognition tasks, with comparisons to state-of-the-art attack methods, such as Google Patch \citep{brown2017adversarial}, LaVAN \citep{karmon2018lavan}, GDPA \citep{li2021generative}, and Masked Projected Gradient Descent (MPGD), which is an extension of the standard PGD attack introduced in~\citet{madry2017towards}. In addition, we evaluate the effectiveness of our attack against existing defense methods designed specifically against adversarial patch attacks~\citep{hayes2018visible,chen2022jujutsu,liu2022segment,tarchoun2023jedi,fu2024diffpad,kang2024diffender}. For GDPA,  we balance attack efficacy and imperceptibility by setting the visibility parameter $\alpha$ to $0.4$, while for MPGD, we set the $l_{\infty}$ perturbation bound to $ \epsilon = 16/255$.

\shortsection{Dataset and Model Setup}
We consider a subset of the ILSVRC 2012 validation set~\citep{russakovsky2015imagenet} consisting of $1000$ correctly classified images, one from each class, for image classification. For face recognition, following \citet{li2021generative}, we use the test set of the VGG face dataset \citep{Parkhi15,li2021generative}, consisting of a total of $470$ images across 10 classes. We consider four target network architectures: ResNet-50 \citep{he2016deep}, VGG16 \citep{simonyan2014very}, Swin Transformer Tiny, and Swin Transformer Base \citep{liu2021swin}. For image classification, we use their pre-trained weights. For face recognition, we re-train them on the VGG Face dataset’s train set, which comprised $3,178$ images across $10$ classes. The retraining procedure follows the same specifications as used by \cite{li2021generative}. All the images in both tasks are resized to a dimension of $224 \times 224$ before being attacked.

\shortsection{Attack Configuration}
In our experiments, we optimize the patch until the target class confidence reaches $0.9$ or for a maximum of $1,000$ iterations. The patch size is fixed at $84\times 84$, covering $14\%$ of the image. While we use a square patch following prior works, our optimization framework can be generalized to other shapes. If the attack fails, we reinitialize the step size up to three times. All experiments are conducted on a single NVIDIA A100 GPU ($80$ GB), using PyTorch as the deep learning framework.

\shortsection{Attack Success Rate}
We evaluate the effectiveness of different attack methods based on targeted attack success rate, denoted as ASR, which characterizes the ratio of instances that can be successfully attacked using the evaluated method. Let $\mathcal{A}$ be the evaluated attack, $f_\theta$ be the victim model, and $\mathcal{S}$ be a test set of correctly classified images. The ASR of $\mathcal{A}$ with respect to $f_\theta$ and $\mathcal{S}$ is defined as:
\begin{equation}
\label{eq:10}
    \mathrm{ASR}(\mathcal{A}; f_\theta, \mathcal{S}) = \frac{1}{|\mathcal{S}|}\sum_{\bm{x} \in \mathcal{S}} \mathds{1}\big(f_\theta(\hat{\bm{x}}) = y_{\mathrm{targ}}\big),
\end{equation}
where $|\mathcal{S}|$ denotes the cardinality of $\mathcal{S}$, and $\hat{\bm{x}}$ is the adversarial example generated by $\mathcal{A}$ for $\bm{x}$.

\shortsection{Imperceptibility}
To measure patch imperceptibility, we use similarity matrices, incorporating both traditional statistical methods and convolutional neural network (CNN) based measures. The former measures involve Structural Similar Index Measure (SSIM)~\citep{wang2004image}, Universal Image Quality index (UIQ)~\citep{wang2002universal}, and Signal to Reconstruction Error ratio (SRE)~\citep{lanaras2018super}, while the latter involves CLIPScore~\citep{hessel2021clipscore}, and Learned Perceptual Image Patch Similarity (LPIPS) metric~\citep{zhang2018unreasonable}. 
SSIM measures structural similarity, while UIQ evaluates distortion based on correlation, luminance, and contrast, yielding a single index within $[-1, 1]$. SRE, akin to PSNR, measures error relative to the signal's power, ensuring consistency across different brightness levels. CLIPScore and LPIPS assess perceptual similarity using pre-trained DNNs, capturing subtle visual features. We evaluate similarity between adversarial and original samples on two scales: globally, by comparing entire images, and locally, by analyzing the similarity within the attacked region.

\section{Additional Experiments}
\label{appendix:additional experiments}

\subsection{Additional Results on ImageNet}
\label{appendix:additional results on Imagenet}

In this section, we present additional detailed results corresponding to every victim model considered, with ``Toaster'' as the target class. The results include a comprehensive analysis of our attack's stealthiness, including all the imperceptibility metrics considered and mentioned earlier. 

\begin{figure}[t]
  \centering
  \includegraphics[width=1\linewidth]{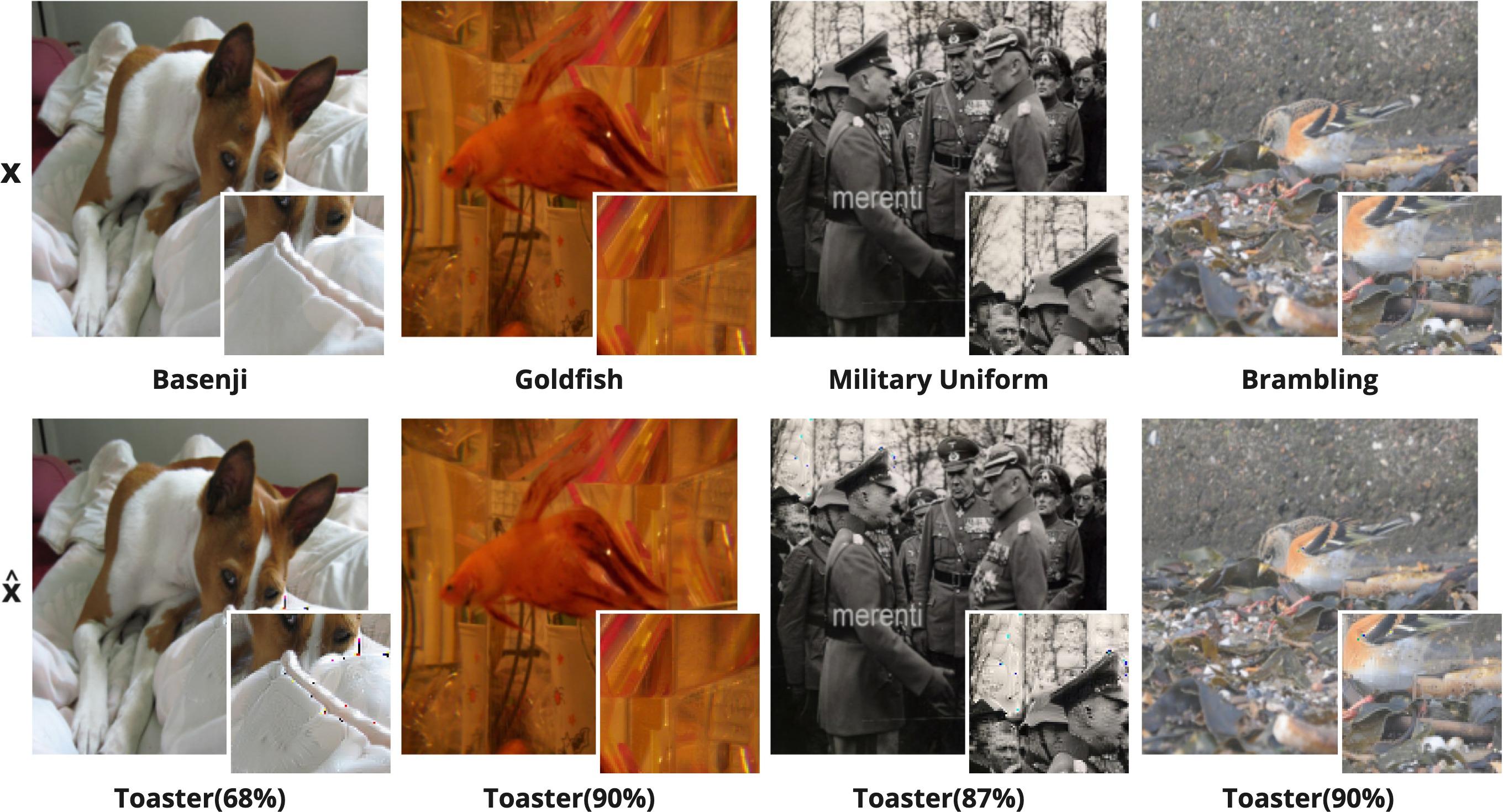}
  \vspace{-0.1in} % Adjust space if needed
  \caption{Visualizations of the original images and their adversarial counterparts produced by IAP corresponding to the target class on the ImageNet Dataset with \textbf{VGG16} as the victim model. \( x \) represents the benign sample, and \( \hat{x} \) represents the adversarial samples with the generated adversarial patch corresponding to the target class. The smaller images at the right-bottom corner correspond to the optimal location $(i', j')$.}
  \label{fig:vgg16img}
  \vspace{-0.1in}
\end{figure}

\begin{figure}[t]
  \centering
  \includegraphics[width=1\linewidth]{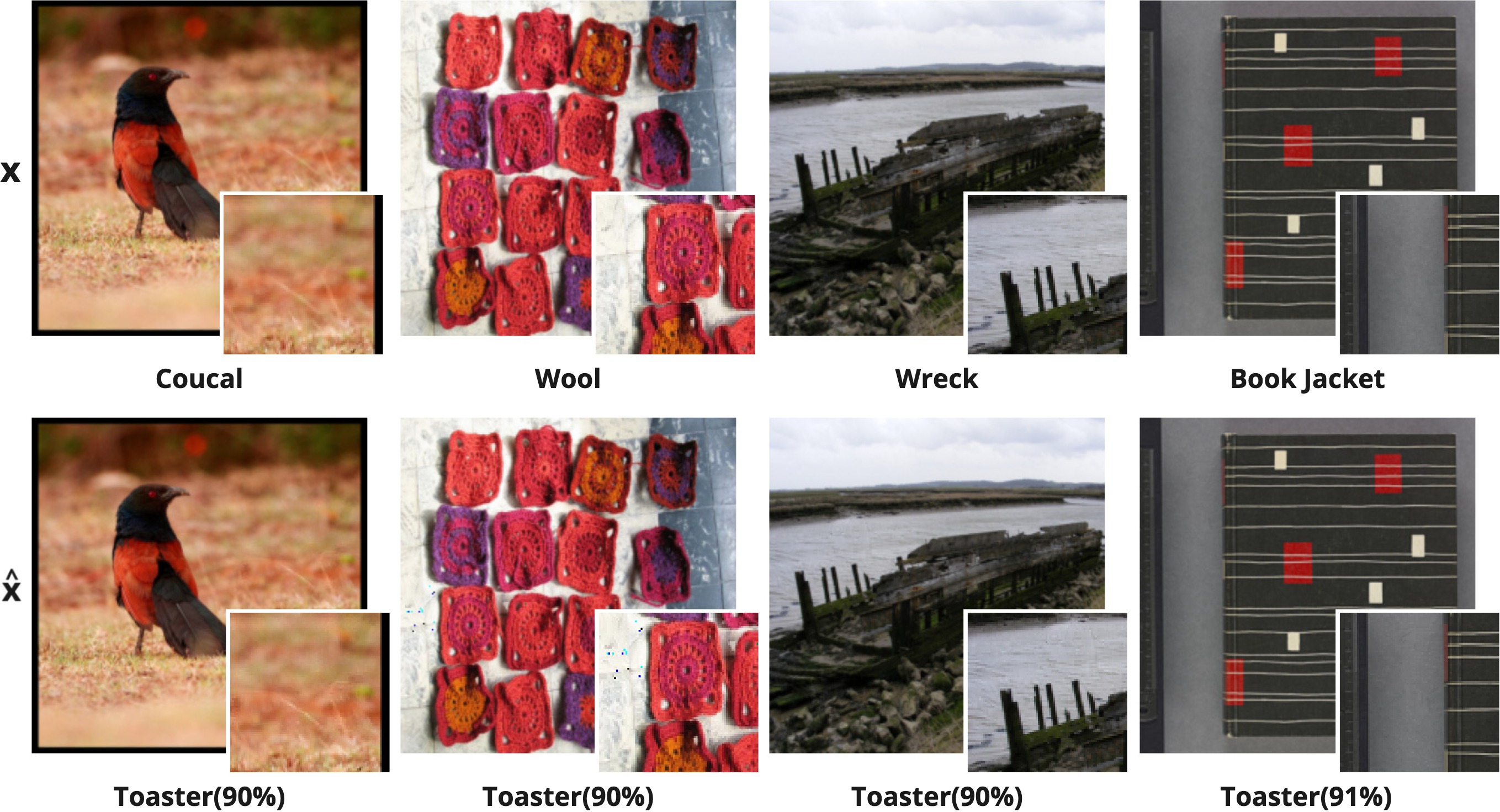}
  \vspace{-0.1in} % Adjust space if needed
  \caption{Visualizations of the original images and their adversarial counterparts produced by IAP corresponding to the target class on the ImageNet Dataset with \textbf{Swin Transformer Tiny} as the victim model. \( x \) represents the benign sample, and \( \hat{x} \) represents the adversarial samples with the generated adversarial patch corresponding to the target class. The smaller images at the right-bottom corner correspond to the optimal location $(i', j')$.}
  \label{fig:swintimg}
  \vspace{-0.1in} 
\end{figure}

\begin{figure}[t]
  \centering
  \includegraphics[width=1\linewidth]{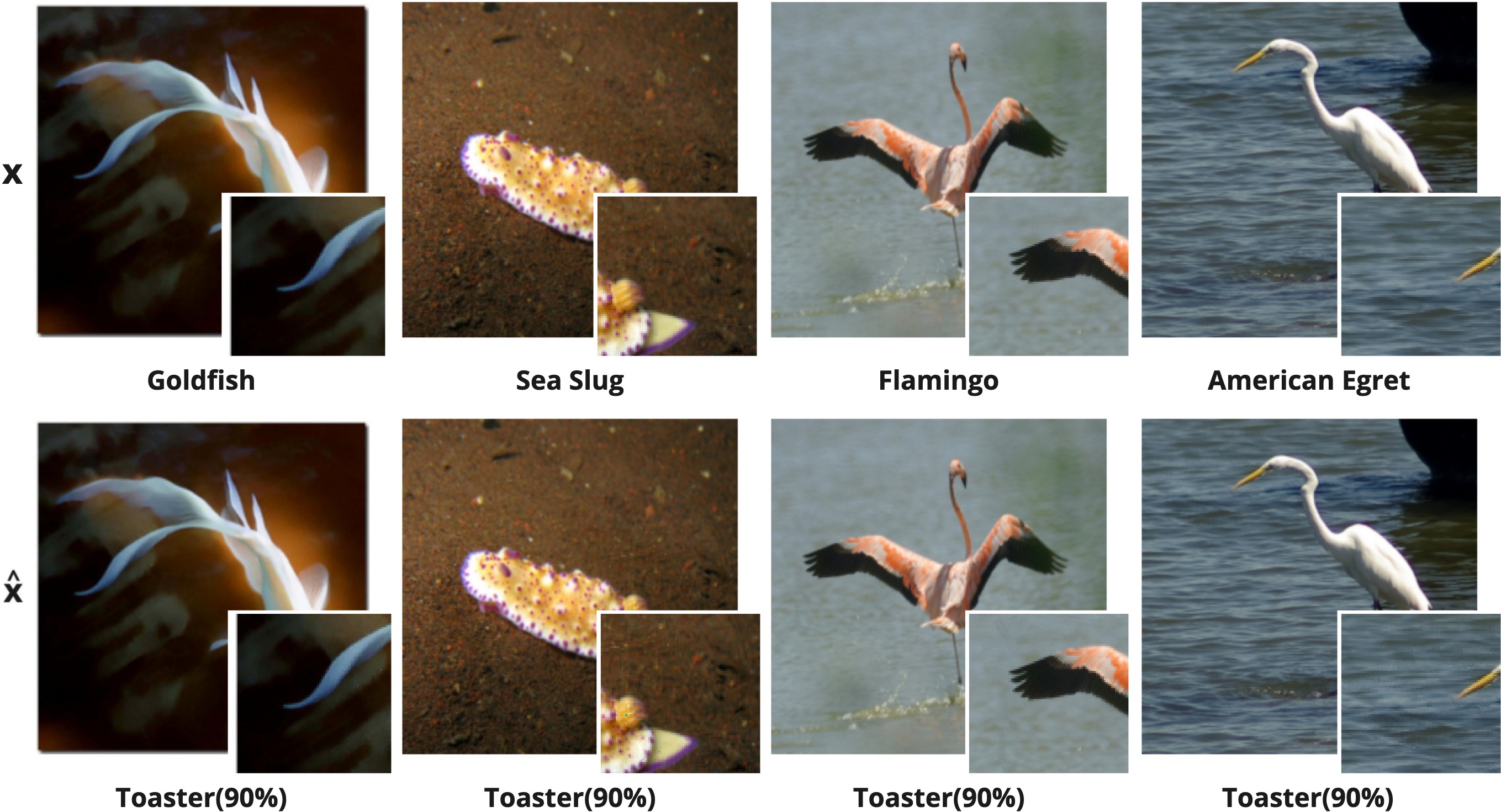}
  \vspace{-0.1in} % Adjust space if needed
  \caption{Visualizations of the original images and their adversarial counterparts produced by IAP corresponding to the target class on the ImageNet Dataset with \textbf{Swin Transformer Base} as the victim model. \( x \) represents the benign sample, and \( \hat{x} \) represents the adversarial samples with the generated adversarial patch corresponding to the target class. The smaller images at the right-bottom corner correspond to the optimal location $(i', j')$.}
  \label{fig:swinbimg}
  \vspace{-0.1in} 
\end{figure}

The evaluations across victim models, VGG16, ResNet-50, Swin Transformer Tiny, and Swin Transformer Base, are presented in Tables \ref{table: VGG16results}-\ref{table: SwinBresults} respectively. The results account for the stability of IAP across architectures in terms of ASR, which is either on par with or exceeds the baseline methods considered. As evident from the analysis, we achieve state-of-the-art performance in imperceptibility, further demonstrating its stability. The adversarial samples created corresponding to each victim architecture are shown along with their target class confidence in Figures \ref{fig:vgg16img}-\ref{fig:swinbimg}.

\begin{figure}[t]
  \centering
  \includegraphics[width=1\linewidth]{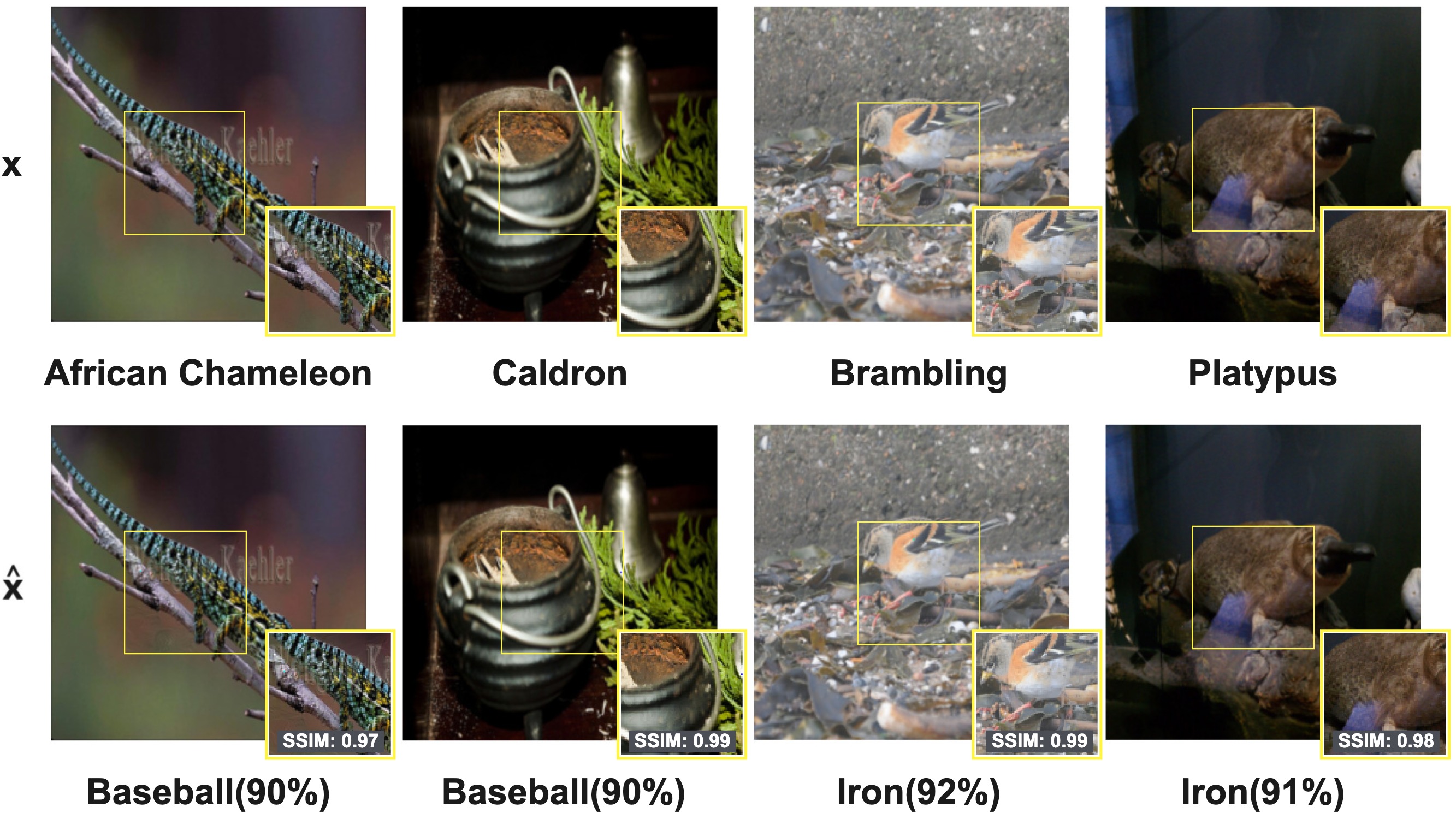}
  \vspace{-0.1in} % Adjust space if needed
  \caption{Visualizations of the original images and their adversarial counterparts produced by IAP corresponding to the target class on the ImageNet Dataset with \textbf{ResNet-50} as the victim model. \( x \) represents the benign sample, and \( \hat{x} \) represents the adversarial samples with the generated adversarial patch corresponding to the target class. The smaller images at the right-bottom corner correspond to the optimal location $(i', j')$.}
  \label{fig:othertarget}
\vspace{-0.1in}  
\end{figure}

\shortsection{Cross-Class Attack Stability}
To assess the effectiveness of our attack across multiple target classes, we extend our evaluation beyond the ``Toaster'' class to include ``Baseball'' and ``Iron'' as additional targets. We employ ResNet-50 as the victim model while maintaining all other attack configurations consistent with previous experiments. The results, summarized in Table~\ref{table: multiTarget results}, demonstrate that IAP achieves consistently high ASR across different target classes while preserving its imperceptibility, as illustrated in Figure~\ref{fig:othertarget}. Notably, our approach exhibits stability across classes, achieving an ASR of $99.47 \pm 0.13$ and maintaining high imperceptibility, exemplified by a local SSIM of $0.94\pm ± 0.005$.

\subsection{Additional Results on VGG Face}
\label{appendix:additional results on VGG Face}

\begin{figure}[t]
  \centering
  \includegraphics[width=1\linewidth]{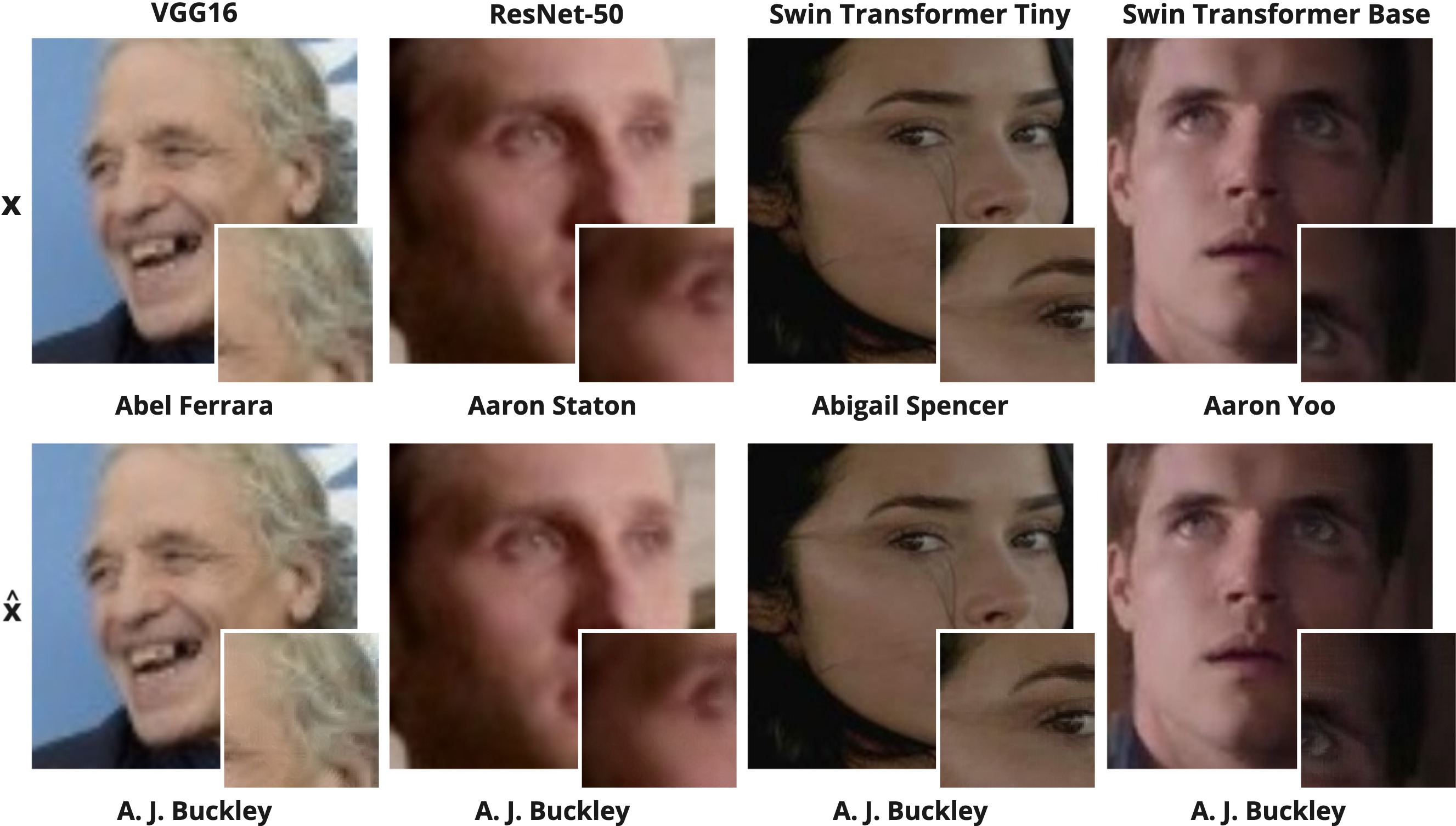}
  \vspace{-0.1in} % Adjust space if needed
  \caption{Visualizations of the original images and their adversarial counterparts with IAP and the target class \textbf{``A. J. Buckley''} on the VGG Face Dataset. \( x \) represents the benign sample, and \( \hat{x} \) represents the adversarial samples with the generated adversarial patch corresponding to the target class. The smaller images at the right-bottom corner correspond to the optimal location $(i', j')$.}
  \label{fig:vgg16vgg0}
  \vspace{-0.1in}
\end{figure}

\begin{figure}[t]
  \centering
  \includegraphics[width=1\linewidth]{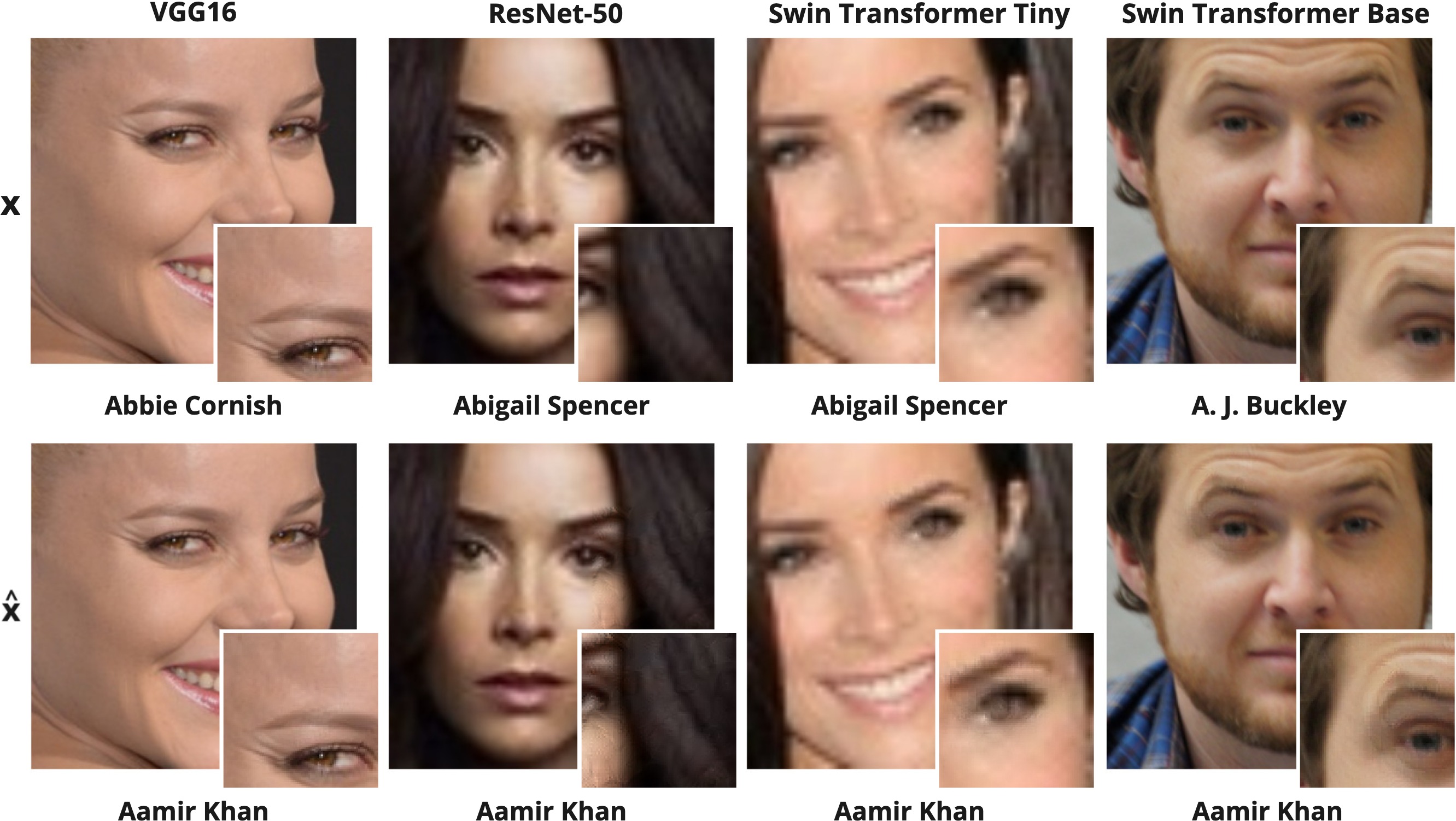}
  \vspace{-0.1in} % Adjust space if needed
  \caption{Visualizations of the original images and their adversarial counterparts with IAP and the target class \textbf{``Aamir Khan''} on the VGG Face Dataset. \( x \) represents the benign sample, and \( \hat{x} \) represents the adversarial samples with the generated adversarial patch corresponding to the target class. The smaller images at the right-bottom corner correspond to the optimal location $(i', j')$.}
  \label{fig:vgg16vgg2}
  \vspace{-0.1in} 
\end{figure}

\begin{figure}[t]
  \centering
  \includegraphics[width=1\linewidth]{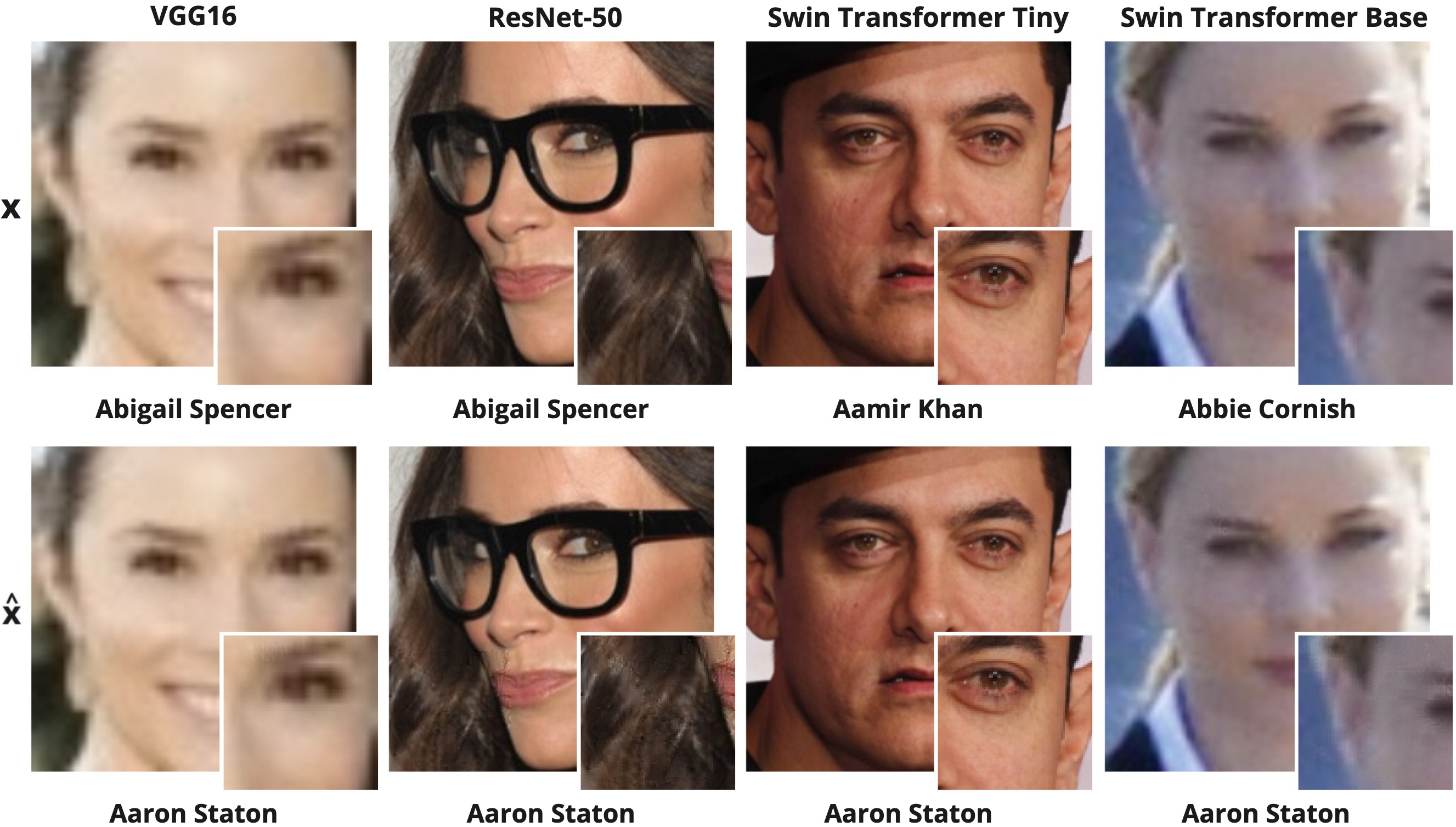}
  \vspace{-0.1in} % Adjust space if needed
  \caption{Visualizations of the original images and their adversarial counterparts with IAP and the target class \textbf{``Aaron Staton''} on the VGG Face Dataset. \( x \) represents the benign sample, and \( \hat{x} \) represents the adversarial samples with the generated adversarial patch corresponding to the target class. The smaller images at the right-bottom corner correspond to the optimal location $(i', j')$.}
  \label{fig:vgg16vgg3}
\end{figure}

Here, we present detailed results explicitly corresponding to every victim model corresponding to the three target classes considered. Tables \ref{table: VGG16_0_results}-\ref{table: VGG16_3_results}, \ref{table: resnet_0_results}-\ref{table: resnet_3_results}, \ref{table: swint_0_results}-\ref{table: swint_3_results}, and \ref{table: swinb_0_results}-\ref{table: swinb_3_results} summarize the results for the three target classes using VGG16, ResNet-50, Swin Transformer Tiny, and Swin Transformer Base as victim models, respectively. The results show consistent attack performance across the criteria considered, as well as achieving state-of-the-art imperceptibility performance, which further demonstrates its efficiency. The adversarial samples corresponding to the target classes ``A. J. Buckley'', ``Aamir Khan'', and ``Aaron Staton'' are shown in Figures \ref{fig:vgg16vgg0}-\ref{fig:vgg16vgg3} respectively.

\section{Additional Analyses}
\label{appendix:further analysis}

\subsection{Ablation Studies}
\label{appendix:ablation studies}

\begin{figure}[t]
  \centering
  \includegraphics[width=1\linewidth]{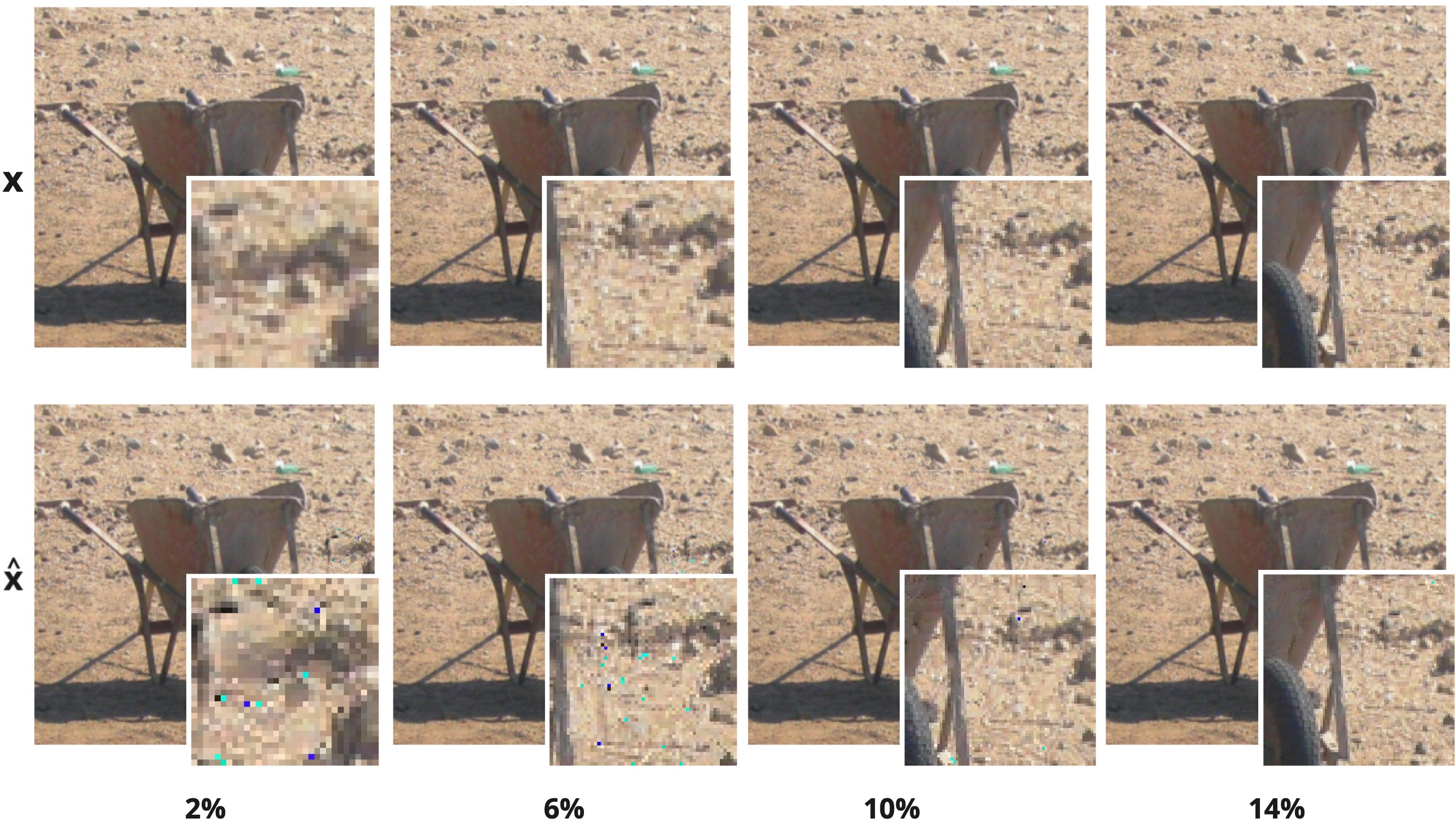}
  \vspace{-0.1in} % Adjust space if needed
  \caption{Visualizations of the impact of the patch sizes on attack imperceptibility. \( x \) represents the benign sample, and \( \hat{x} \) represents the adversarial samples with the generated adversarial patch corresponding to the target class. The smaller images at the right-bottom corner correspond to the optimal location $(i', j')$.}
  \label{fig:PSimg}
\end{figure}

\begin{figure}[t]
  \centering
  \includegraphics[width=1\linewidth]{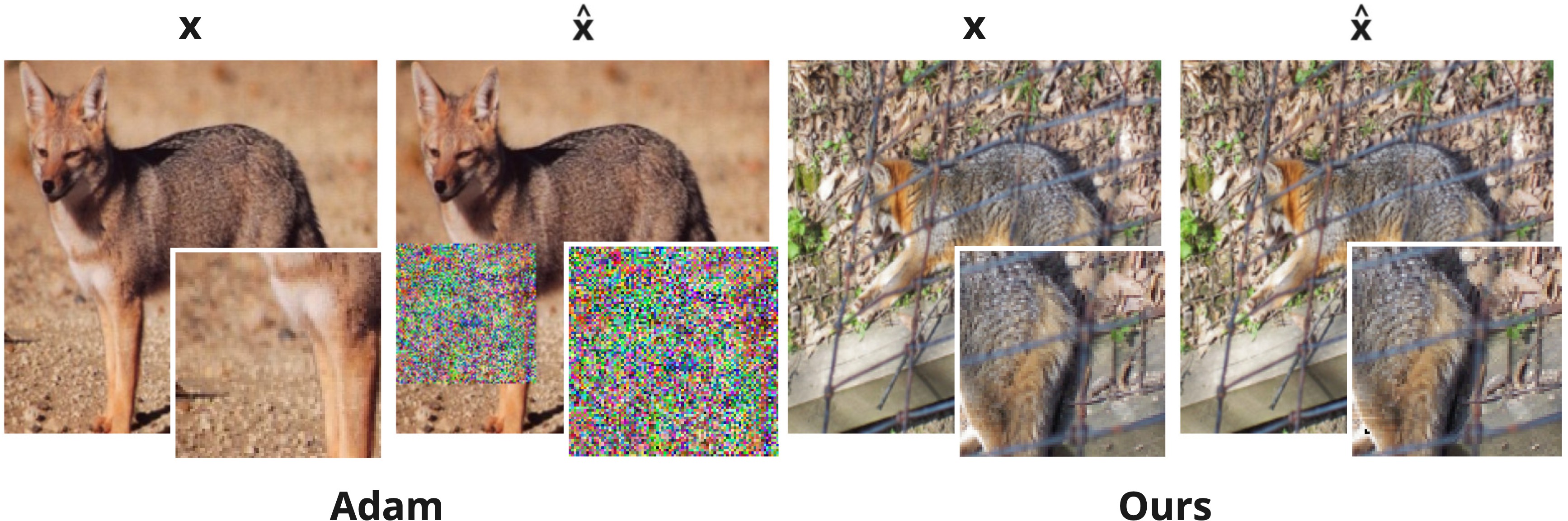}
  \vspace{-0.1in} % Adjust space if needed
  \caption{Visualizations of adversarial patch generated by update rule from Adam optimizer vs IAP. \( x \) represents the benign sample, and \( \hat{x} \) represents the adversarial samples with the generated adversarial patch corresponding to the target class. The smaller images at the right-bottom corner correspond to the optimal location $(i', j')$.}
  \label{fig:URimg}
\end{figure}

We perform ablation studies to assess key components of IAP, including patch size, update iterations, and the regularization coefficient in the loss function (Equation \ref{eq:8}). We compare our update rule with the Adam optimizer and test the assumption that adversarial patches attract classifier's attention. All experiments use ImageNet with Swin Transformer Base as the victim model. Here, we detail the comprehensive evaluation corresponding to both attack efficacy and imperceptibility. Table \ref{table: PSabla} demonstrates that as the patch size increases, the imperceptibility improves. Figure \ref{fig:PSimg} validates this as we see that the attack area becomes smoother with the increase in the size.  Aligned with our hypothesis, initial increase in $w_3$ improved the imperceptibility of the generated patches as presented in Table \ref{table: w3abla}. We studied the impact of the update rule proposed by our method, IAP, by altering it with the update rule corresponding to the Adam optimizer. As shown in Table \ref{table: URabla} and visualized in Figure \ref{fig:URimg}, we achieve most of our imperceptibility because of the update rule we utilize for updating the perturbation. In addition, we also considered the effect of the number of optimization steps on the ASR and imperceptibility of the attack.

\begin{figure}[t]
  \centering
  \includegraphics[width=1\linewidth]{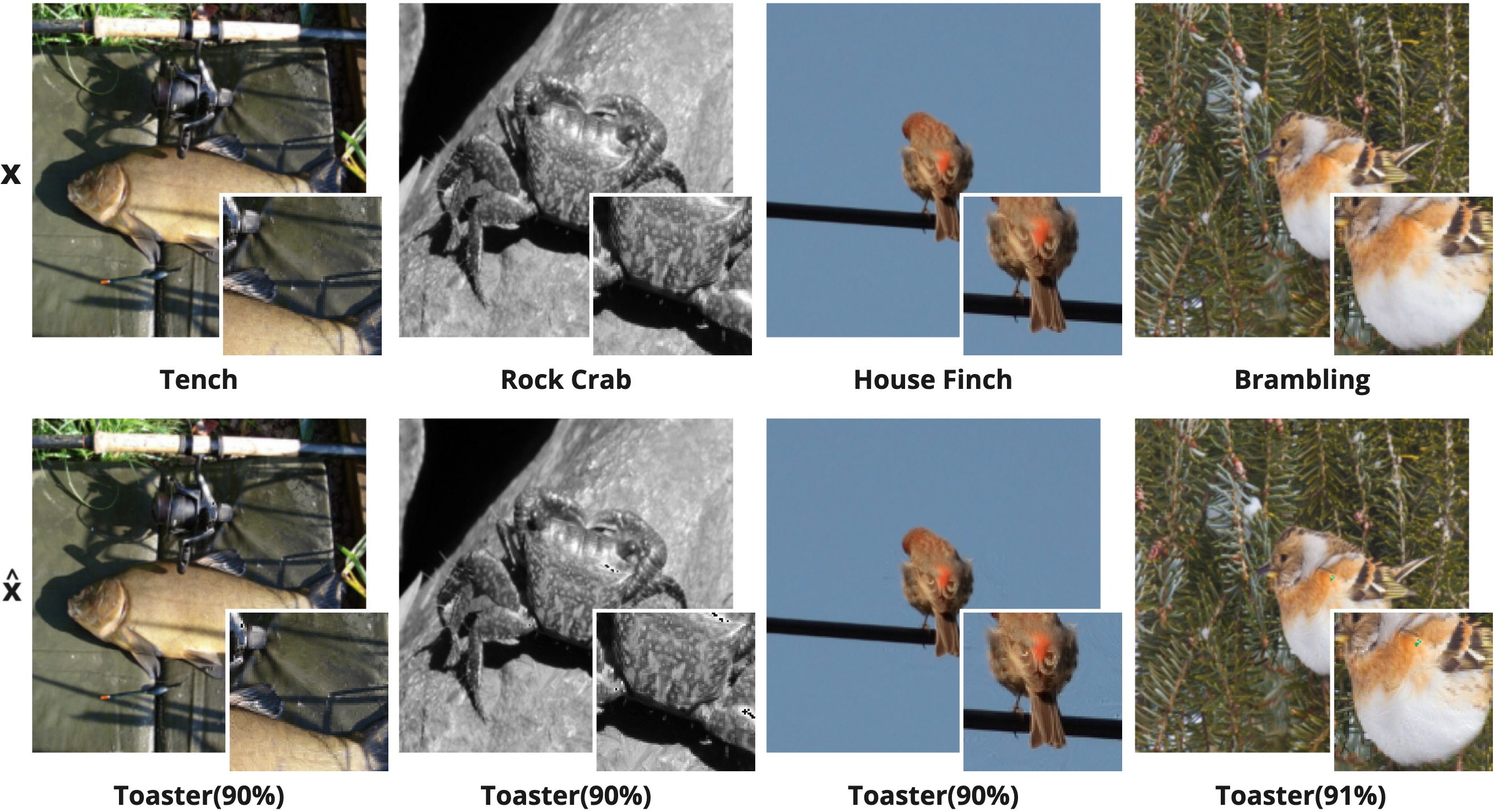}
  \vspace{-0.1in} % Adjust space if needed
  \caption{Visualizations of the original images and their adversarial counterparts produced by IAP corresponding to the target class on the ImageNet Dataset with \textbf{ResNet-50} as the victim model. \( x \) represents the benign sample, and \( \hat{x} \) represents the adversarial samples with the generated adversarial patch corresponding to the target class. The smaller images at the right-bottom corner correspond to the optimal location $(i', j')$.}
  \label{fig:res50img}
  \vspace{-0.1in} 
\end{figure}

\shortsection{Effect of Patch Size}
We evaluate the impact of patch size on attack efficacy and imperceptibility. We hypothesize that increasing the patch size would enhance attack performance and imperceptibility, as the perturbations would disperse over a larger area while remaining less salient. The results support this hypothesis, with a $99.4\%$ attack success rate (ASR) for a patch covering $14\%$ of the image, compared to $72.2\%$ ASR for $2\%$ coverage. For patch sizes of $4\%$ or more, the ASR reached $90.7\%$ or higher. These findings also show improved imperceptibility with larger patch sizes as highlighted in Table \ref{table: swint_3_results} and Figure \ref{fig:res50img}.

\shortsection{Effect of Regularization Coefficient}
We study the effect of the regularization coefficient $w_3$ in the human-oriented distance metric (Equation \ref{eq:7}), part of the total loss function (Equation \ref{eq:8}). We hypothesize that increasing $w_3$  would improve imperceptibility at the cost of slightly reducing attack performance. Our results support this hypothesis as shown in Table \ref{table: swinb_0_results}. As $w_3$ increases, the attack success rate slightly decreases while imperceptibility improves. However, beyond a certain point, the trend reverses due to the destabilizing effect of large $w_3$ values, which cause the loss function to be dominated by the regularization term, requiring more iterations for successful attacks and reducing imperceptibility.

\shortsection{Effect of Update Rule}
We compare our proposed update rule, which allows for longer iterations with no perturbation magnitude constraints while maintaining imperceptibility, to the widely used Adam Optimizer update rule. We hypothesize that Adam, optimized for attack success, would yield a higher success rate. However, Adam’s updates alter each color channel separately, potentially changing the pixel’s base color, whereas our method preserves it. While Adam achieves a slightly higher attack success rate, IAP completely outperformed it in terms of imperceptibility as demonstrated in Table \ref{table: swinb_2_results}. Figure \ref{fig:5} visualizes and compares the adversarial patches generated by both approaches.

\begin{figure}[t]
  \centering
  \includegraphics[width=1\linewidth]{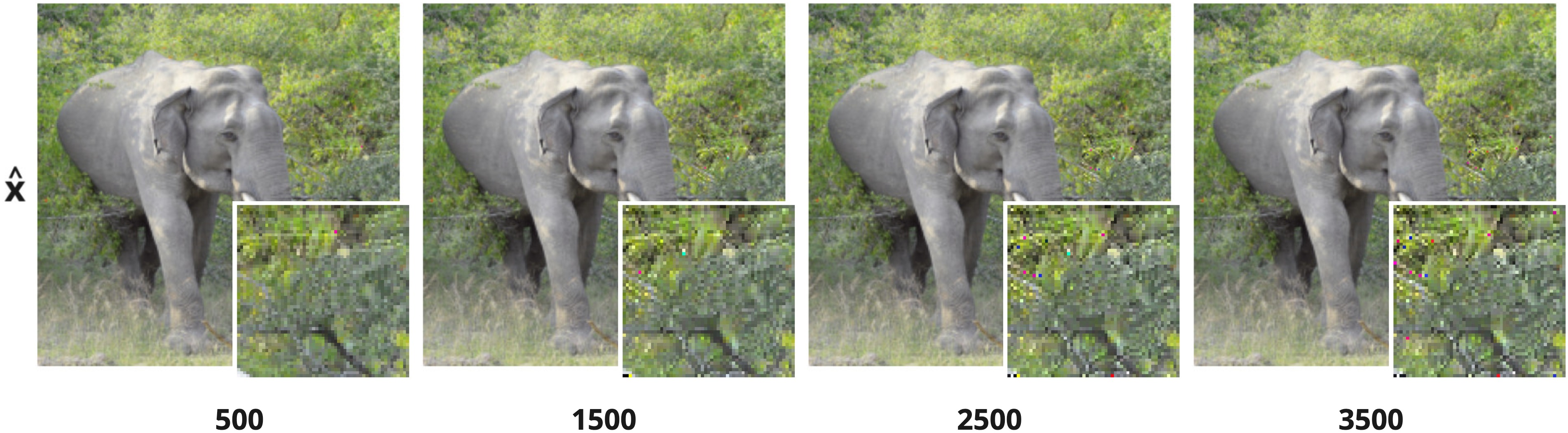}
  \vspace{-0.1in} % Adjust space if needed
  \caption{Visualizations of the impact of the number of update iterations on attack imperceptibility. \( \hat{x} \) represents the adversarial samples with the generated adversarial patch. The smaller images at the right-bottom corner correspond to the optimal location $(i', j')$. The x-axis represents the number of update iterations.}
  \label{fig: NIabla}
\end{figure}

\shortsection{Effect of the number of Update Iterations}
As the number of updates increases, the patch’s appearance diverges from the original, even if the perturbations remain less salient. Despite the reduced saliency, more iterations typically improve attack success rates. In these experiments, we fix the patch size at $6\%$ to evaluate the trade-off between attack efficacy and imperceptibility. The ASR increases as the number of update iterations increases, as shown by Table \ref{table: NIabla}, with a slight reduction in imperceptibility as perturbations accumulate, as shown in Figure \ref{fig: NIabla}.

\subsection{GradCAM analysis of Attention Overlap}
\label{append:GradCAM analysis of attention overlap}

To understand the change in the attention map induced by the adversarial samples generated by IAP, we analyze the shift in the highest attention location of the attention map generated in comparison to the one generated corresponding to the benign sample. We use GradCAM~\citep{selvaraju2017grad} to measure the attention maps. Analysis of the attention maps holds critical significance because of the defense implications that it can have on adversarial patch attacks. We measure the average proportion of the number of adversarial samples for which the location of highest attention in the attention map does not come within the attack surface area. We term this measure as ``NoPatchLoc'', which is defined as follows:
\begin{equation}
\label{eq:NPL}
\text{NoPatchLoc} = \frac{1}{N}\sum_{i=0}^{N} (1-\mathbf{1}_A(x_i,y_i,Ox_i,Oy_i)),
\end{equation}
where $N$ is the total number of adversarial samples analyzed, and the indicator function is defined as follows:
\begin{equation}
\vspace{0.1in}
\resizebox{\linewidth}{!}{
$\mathbf{1}_A(x_i, y_i, O_{x_i}, O_{y_i}) =
\begin{cases} 
1, & \text{if } O_{x_i} \leq x_i < O_{x_i} + s \text{ and } O_{y_i} \leq y_i < O_{y_i} + s \\ 
0, & \text{otherwise}
\end{cases},
$}
\end{equation}
where $s$ denotes the patch size, $(O_{x_i}, O_{y_i})$ is the optimal location identified by our method to locate the adversarial patch, and $(x_i,y_i)$ is the coordinate of the highest attention location. Table~\ref{table: gradatt} demonstrates the NoPatchLoc measures obtained from the generated adversarial samples corresponding to their respective victim models. As evident, except for ResNet-50, where the measure is $53.70\%$, the highest attention location remains consistently outside the attack region for more than $70\%$ of the perturbed samples across all other architectures. The highest occurrence is observed for the Swin Transformer Base, achieving $81.30\%$. This provides strong evidence that accounts for the strong stealth capabilities of our method, as highlighted by the performance against defense methods. 

\begin{figure}[t]
    \centering
    \begin{subfigure}[b]{0.9\linewidth}
        \centering
        \includegraphics[width=\textwidth]{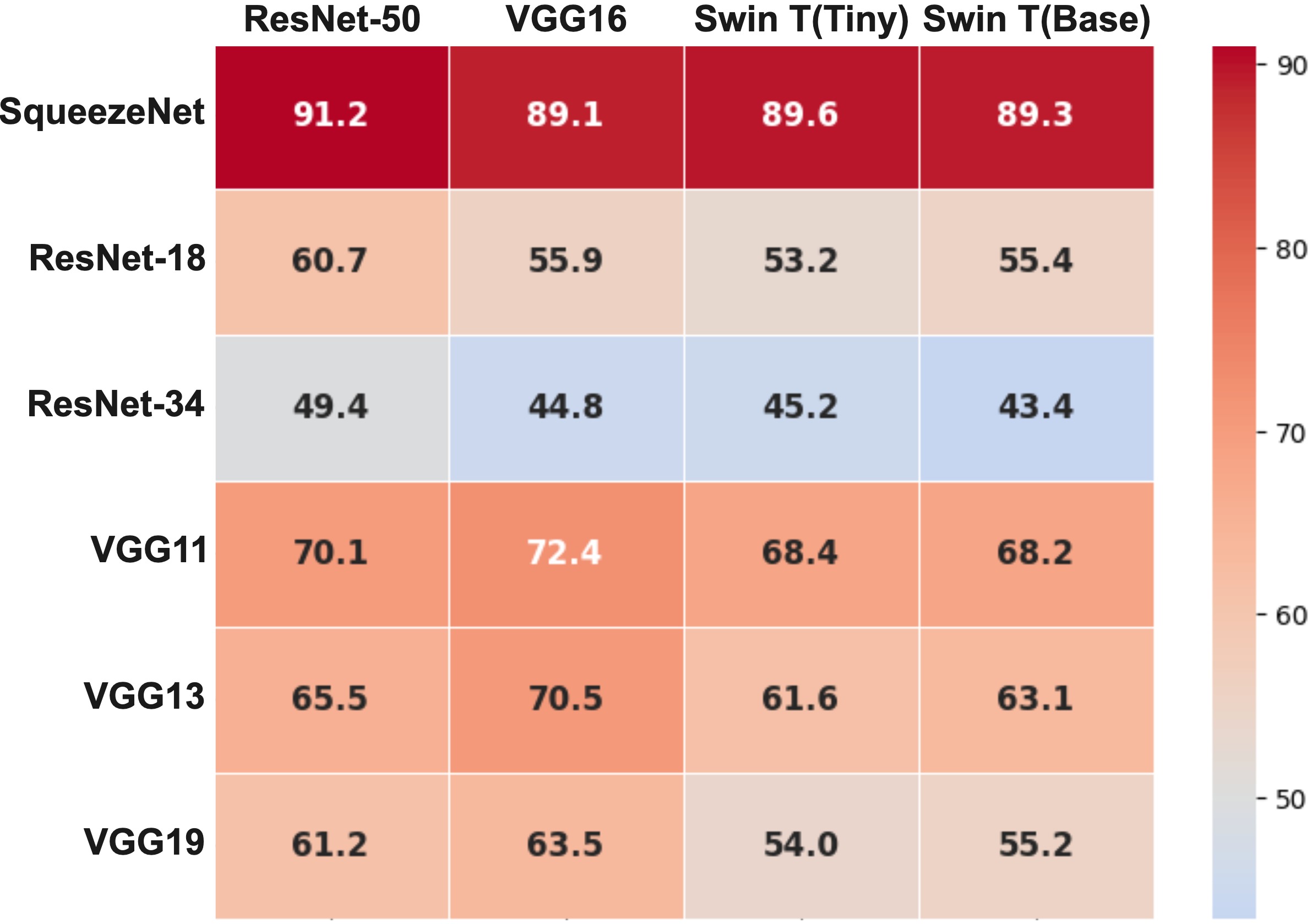}
    \end{subfigure}
    \vspace{-0.05in}
    \caption{Transferability of IAP measured by ASR (\%) on ImageNet. The first row represents the substitute model, and the first column represents the target models.}
    \label{fig:transfer}
    \vspace{-0.05in}
\end{figure}

\begin{table}[t]
\centering
\small
\begin{tabular}{cc}
    \toprule
    \midrule
    \textbf{Model} & \textbf{NoPatchLoc(\%)}\\ 

    \midrule
    \midrule
    VGG16 & $72.15$\\
    ResNet-50 & $53.70$\\
    Swin Transformer Tiny& $73.11$\\
    Swin Transformer Base& $81.30$\\
    \midrule
    Average & $\bm{70.07}$\\
    \midrule
    \bottomrule
\end{tabular}
\vspace{-0.05in}
\caption{Assessment of whether the GradCAM's highest attention location overlaps with the adversarial patch location.}
\label{table: gradatt}
\vspace{-0.1in}
\end{table}

\subsection{Transferability}
\label{append:transferability}

We assess the transferability of our general method in the untargeted scenario without incorporating any adaptations specifically aimed at enhancing attack transferability. Using a substitute model approach, we generate adversarial samples on the previously considered victim models and evaluate their transferability across a set of target models: SqueezeNet, ResNet-18, ResNet-34, VGG11, VGG13, and VGG19. Given that no specific adaptation scheme is used, our method achieves reasonable ASR as shown in Figure \ref{fig:transfer}. The results indicate that transferability is influenced by the architectural similarity between the substitute and target models, as well as their relative model sizes.

\subsection{Black-box Adaptation}
\label{append:black-box adaptation}

While IAP is initially designed as a white-box method, it can be successfully adapted to black-box settings. We ran additional experiments on ImageNet using the following black-box variation of IAP. Specifically, we first approximate the Grad-CAM localization map using a surrogate model (i.e., ResNet-50) for patch placement. Subsequently, we employ a hybrid approach for perturbation optimization, where we initialize the perturbations based on the same surrogate model and refine them using NES, a query-based attack algorithm.
The results are shown in Table \ref{table: bbox} in the main paper, where our black-box IAP variant achieves high (untargeted) attack success rates across different target models. We test $500$ samples per model with a patch size of $84$ and other parameters fixed. Based on white-box convergence trends, we run $400$ surrogate iterations followed by $200$ query-based steps, requiring at most $12,000$ queries. 

\subsection{Physical-World Applicability}
\label{append:physical}
\begin{figure}[t]
    \centering
    \begin{subfigure}[b]{0.98\linewidth}
        \centering
        \includegraphics[width=\linewidth]{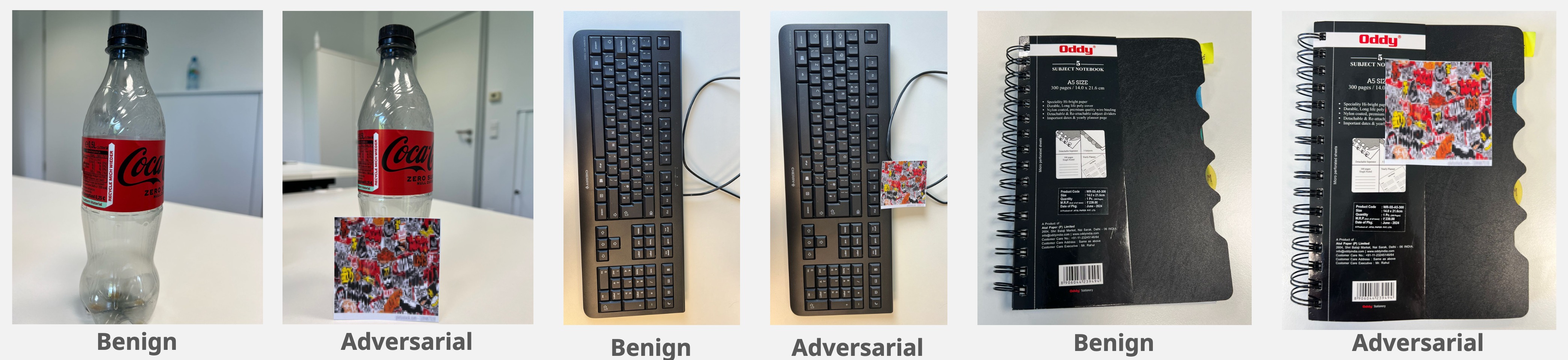}
    \end{subfigure}
    \vspace{-0.1in}
    \caption{Illustrative images of physical-world applications of IAP.}
    \label{fig:physical}
    % \vspace{-0.1in}
\end{figure}

Additionally, we examine IAP's generalizability to physical-world, untargeted attack settings on $5$ object classes. Patches are generated using our optimization scheme, initialized from a reference sticker image like PS-GANs.
To ensure location invariance, each patch is trained by randomly placing it across four proposed ``optimal'' positions from different models. Printed patches are tested on $5$ images per object under varying viewpoints (see Figure \ref{fig:physical} for illustrative examples), achieving an average ASR of $70\%$, showing the potential of IAP's adaptability to physical domains.

\subsection{Flexibility in Patch Shape}
\label{append:flexibility in patch shape}

We also run additional experiments to study whether our method for generating invisible adversarial patches is shape-agnostic. Results are shown in Table \ref{table: Shape}, where we evaluate the performance of IAP using a circular patch with a diameter of $84$ pixels ($11\%$ image area). Under our best-performing setup with Swin Transformer Base as the target model, IAP achieves a high ASR of $99.2\%$ while preserving imperceptibility with LPIPS as low as $0.085$. We believe similar results can also be achieved for other typical patch shapes, since our attack framework supports arbitrary binary masks.

\begin{table}[t]
\centering
\small
\resizebox{\linewidth}{!}{
\begin{tabular}{lccccccc}
    \toprule
    \multirow{2}{*}{\textbf{Shape}} & \multirow{2}{*}{\textbf{ASR}} & \multirow{2}{*}{\textbf{Scale}} & \multicolumn{5}{c}{\textbf{Imperceptibility metric}}\\ 
    \cmidrule{4-8}
    {} &{} & {} & \textbf{SSIM ($\uparrow$)} & \textbf{UIQ ($\uparrow$)} & \textbf{SRE} ($\uparrow$) & \textbf{CLIP} ($\uparrow$) & \textbf{LPIPS} ($\downarrow$)\\

    \midrule
    \multirow{2}{*}{Circle}  & \multirow{2}{*}{$99.2\%$} & Local & $0.95$ & $0.88$ & $27.10$ & $91.46$ & $0.085$\\
    \cmidrule{3-8}
    {} &{} & Global & $0.99$ & $0.98$ & $37.79$ & $99.13$ & $0.016$ \\
    
    \bottomrule
\end{tabular}}
\caption{Performance of IAP in ASR and various imperceptibility metrics with a circular patch shape and patch size of $11\%$.}
\vspace{-0.1in}
  \label{table: Shape}
\end{table}

\newpage
\clearpage

%vgg-16

\clearpage
\newpage

\begin{table}[t]
    \centering
    \small
    \resizebox{\linewidth}{!}{
    \begin{tabular}{cclccccc}
        \toprule
        \multirow{2}{*}{\textbf{Method}} & \multirow{2}{*}{\textbf{ASR(\%)}} & \multirow{2}{*}{\textbf{Scale}} & \multicolumn{5}{c}{\textbf{Imperceptibility metric}}\\ 
        \cmidrule{4-8}
        {} &{} & {} & \textbf{SSIM ($\uparrow$)} & \textbf{UIQ ($\uparrow$)} & \textbf{SRE} ($\uparrow$) & \textbf{CLIP} ($\uparrow$) & \textbf{LPIPS} ($\downarrow$)\\

        \midrule
        \multirow{2}{*}{Google Patch}  & \multirow{2}{*}{\textbf{100}} & Local & 0.002 & 0.000 & 11.93 & 32.50 & 0.760\\
        \cmidrule{3-8}
        {} & {} & Global & 0.830 & 0.820 & 18.73 & 73.10 & 0.190 \\

        \midrule
        \multirow{2}{*}{LaVAN}  &\multirow{2}{*}{93.6} & Local & 0.002 & 0.000 & 11.13 & 33.20 & 0.790\\
        \cmidrule{3-8}
        {} & {} & Global & 0.820 & 0.810 & 20.30 & 76.32 & 0.230 \\
        
        \midrule
        \multirow{2}{*}{GDPA}  &\multirow{2}{*}{89.2} & Local & 0.310 & 0.300 & 19.90 & 56.25 & 0.610\\
        \cmidrule{3-8}
        {} & {} & Global & 0.890 & 0.880 & 28.00 & 84.00 & 0.130 \\

        \midrule
        \multirow{2}{*}{MPGD}  & \multirow{2}{*}{96.5} & Local & 0.810 & 0.800 & 26.44 & 73.91 & 0.320\\
        \cmidrule{3-8}
        {} &{} & Global & 0.940 & 0.920 & 32.80 & 94.00 & 0.090 \\

        \midrule
        \multirow{2}{*}{Ours}  & \multirow{2}{*}{\textbf{99.1}} & Local & \textbf{0.900} & \textbf{0.860} & \textbf{28.94} & \textbf{72.70} & \textbf{0.230}\\
        \cmidrule{3-8}
        {} &{} & Global & \textbf{0.985} & \textbf{0.960} & \textbf{36.42} & \textbf{95.10} & \textbf{0.060} \\
        
        \bottomrule
    \end{tabular}}
    \vspace{-0.1in}
    \caption{Detailed comparison of attack efficacy through ASR ($\%$) and imperceptibility with \textbf{VGG16} as the victim model on the \textbf{ImageNet} dataset. For SSIM, UIQ, SRE, and CLIP scores, the higher ($\uparrow$) the better, while the lower ($\downarrow$) the better for LIPIPS.}
   \label{table: VGG16results}
\end{table}

\begin{table}[t]
    \centering
    \small
    \resizebox{\linewidth}{!}{
    \begin{tabular}{cclccccc}
        \toprule
        \multirow{2}{*}{\textbf{Method}} & \multirow{2}{*}{\textbf{ASR(\%)}} & \multirow{2}{*}{\textbf{Scale}} & \multicolumn{5}{c}{\textbf{Imperceptibility metric}}\\ 
        \cmidrule{4-8}
        {} &{} & {} & \textbf{SSIM ($\uparrow$)} & \textbf{UIQ ($\uparrow$)} & \textbf{SRE} ($\uparrow$) & \textbf{CLIP} ($\uparrow$) & \textbf{LPIPS} ($\downarrow$)\\

        \midrule
        \multirow{2}{*}{Google Patch}  & \multirow{2}{*}{99.1} & Local & 0.010 & 0.000 & 14.20 & 33.00 & 0.740\\
        \cmidrule{3-8}
        {} & {} & Global & 0.820 & 0.810 & 22.90 & 74.10 & 0.180 \\

        \midrule
        \multirow{2}{*}{LaVAN}  &\multirow{2}{*}{\textbf{100}} & Local & 0.010 & 0.000 & 14.20 & 33.30 & 0.780\\
        \cmidrule{3-8}
        {} & {} & Global & 0.820 & 0.810 & 23.40 & 76.10 & 0.180 \\
        
        \midrule
        \multirow{2}{*}{GDPA}  &\multirow{2}{*}{93.7} & Local & 0.350 & 0.330 & 19.80 & 65.20 & 0.570\\
        \cmidrule{3-8}
        {} & {} & Global & 0.920 & 0.910 & 28.40 & 87.10 & 0.090 \\

        \midrule
        \multirow{2}{*}{MPGD}  & \multirow{2}{*}{97.8} & Local & 0.790 & 0.780 & 25.30 & 76.20 & 0.240\\
        \cmidrule{3-8}
        {} &{} & Global & 0.950 & 0.930 & 33.60 & 93.30 & 0.050 \\

        \midrule
        \multirow{2}{*}{Ours}  & \multirow{2}{*}{\textbf{99.5}} & Local & \textbf{0.940} & \textbf{0.910} & \textbf{28.34} & \textbf{84.54} & \textbf{0.120}\\
        \cmidrule{3-8}
        {} &{} & Global & \textbf{0.990} & \textbf{0.970} & \textbf{37.23} & \textbf{96.52} & \textbf{0.020} \\
        
        \bottomrule
    \end{tabular}}
    \vspace{-0.1in}
    \caption{Detailed comparison of attack efficacy through ASR ($\%$) and imperceptibility with \textbf{ResNet-50} as the victim model on the ImageNet dataset. For SSIM, UIQ, SRE, and CLIP scores, the higher ($\uparrow$) the better, while the lower ($\downarrow$) the better for LIPIPS.}
   \label{table: Resnet50results}
\end{table}

\begin{table}[t]
    \centering
    \resizebox{\linewidth}{!}{
    \begin{tabular}{cclccccc}
        \toprule
        \multirow{2}{*}{\textbf{Method}} & \multirow{2}{*}{\textbf{ASR(\%)}} & \multirow{2}{*}{\textbf{Scale}} & \multicolumn{5}{c}{\textbf{Imperceptibility metric}}\\ 
        \cmidrule{4-8}
        {} &{} & {} & \textbf{SSIM ($\uparrow$)} & \textbf{UIQ ($\uparrow$)} & \textbf{SRE} ($\uparrow$) & \textbf{CLIP} ($\uparrow$) & \textbf{LPIPS} ($\downarrow$)\\

        \midrule
        \multirow{2}{*}{Google Patch}  & \multirow{2}{*}{\textbf{99.8}} & Local & 0.002 & 0.000 & 11.80 & 32.80 & 0.770\\
        \cmidrule{3-8}
        {} & {} & Global & 0.830 & 0.820 & 18.94 & 73.90 & 0.150 \\

        \midrule
        \multirow{2}{*}{LaVAN}  &\multirow{2}{*}{99.7} & Local & 0.005 & 0.000 & 14.13 & 33.10 & 0.780\\
        \cmidrule{3-8}
        {} & {} & Global & 0.820 & 0.810 & 23.30 & 76.32 & 0.170 \\

        \midrule
        \multirow{2}{*}{GDPA}  &\multirow{2}{*}{83.7} & Local & 0.390 & 0.360 & 20.20 & 63.65 & 0.540\\
        \cmidrule{3-8}
        {} & {} & Global & 0.900 & 0.890 & 28.21 & 85.75 & 0.100 \\

        \midrule
        \multirow{2}{*}{MPGD}  & \multirow{2}{*}{98.8} & Local & 0.800 & 0.790 & 25.50 & 80.54 & 0.190\\
        \cmidrule{3-8}
        {} &{} & Global & 0.940 & 0.920 & 33.11 & 95.80 & 0.050 \\
 
        \midrule
        \multirow{2}{*}{Ours}  & \multirow{2}{*}{\textbf{99.6}} & Local & \textbf{0.980} & \textbf{0.940} & \textbf{31.74} & \textbf{90.41} & \textbf{0.060}\\
        \cmidrule{3-8}
        {} &{} & Global & \textbf{0.996} & \textbf{0.980} & \textbf{40.67} & \textbf{98.61} & \textbf{0.008} \\
        
        \bottomrule
    \end{tabular}}
    \vspace{-0.1in}
    \caption{Detailed comparison of ASR ($\%$) and imperceptibility with \textbf{Swin Transformer Tiny} as the victim model on the ImageNet dataset. For SSIM, UIQ, SRE, and CLIP scores, the higher ($\uparrow$) the better, while the lower ($\downarrow$) the better for LIPIPS.}
    \label{table: SwinTresults}
\end{table}

\begin{table}[t]
    \centering
    \small
    \resizebox{\linewidth}{!}{
    \begin{tabular}{cclccccc}
        \toprule
        \multirow{2}{*}{\textbf{Method}} & \multirow{2}{*}{\textbf{ASR(\%)}} & \multirow{2}{*}{\textbf{Scale}} & \multicolumn{5}{c}{\textbf{Imperceptibility metric}}\\ 
        \cmidrule{4-8}
        {} &{} & {} & \textbf{SSIM ($\uparrow$)} & \textbf{UIQ ($\uparrow$)} & \textbf{SRE} ($\uparrow$) & \textbf{CLIP} ($\uparrow$) & \textbf{LPIPS} ($\downarrow$)\\

        \midrule
        \multirow{2}{*}{Google Patch}  & \multirow{2}{*}{97.9} & Local & 0.003 & 0.000 & 10.74 & 32.90 & 0.770\\
        \cmidrule{3-8}
        {} & {} & Global & 0.830 & 0.820 & 17.61 & 73.20 & 0.170 \\

        \midrule
        \multirow{2}{*}{LaVAN}  &\multirow{2}{*}{\textbf{100}} & Local & 0.004 & 0.000 & 13.10 & 33.19 & 0.780\\
        \cmidrule{3-8}
        {} & {} & Global & 0.820 & 0.810 & 23.30 & 76.35 & 0.180 \\
        
        \midrule
        \multirow{2}{*}{GDPA}  &\multirow{2}{*}{85.1} & Local & 0.360 & 0.345 & 20.40 & 61.25 & 0.540\\
        \cmidrule{3-8}
        {} & {} & Global & 0.880 & 0.870 & 28.00 & 85.10 & 0.110 \\

        \midrule
        \multirow{2}{*}{MPGD}  & \multirow{2}{*}{70.5} & Local & 0.800 & 0.800 & 25.30 & 74.30 & 0.200\\
        \cmidrule{3-8}
        {} &{} & Global & 0.940 & 0.920 & 33.00 & 92.10 & 0.050 \\

        \midrule
        \multirow{2}{*}{Ours}  & \multirow{2}{*}{\textbf{99.4}} & Local & \textbf{0.970} & \textbf{0.910} & \textbf{31.30} & \textbf{89.33} & \textbf{0.070}\\
        \cmidrule{3-8}
        {} &{} & Global & \textbf{0.994} & \textbf{0.970} & \textbf{40.10} & \textbf{98.43} & \textbf{0.010} \\
        
        \bottomrule
    \end{tabular}}
    \vspace{-0.1in}
    \caption{Detailed comparison of ASR ($\%$) and imperceptibility with \textbf{Swin Transformer Base} as the victim model on the ImageNet dataset. For SSIM, UIQ, SRE, and CLIP scores, the higher ($\uparrow$) the better, while the lower ($\downarrow$) the better for LIPIPS.}
    \label{table: SwinBresults}
\end{table}

\begin{table}[t]
    \centering
    \small
    \resizebox{\linewidth}{!}{
    \begin{tabular}{cclccccc}
        \toprule
        \multirow{2}{*}{\textbf{y$_{targ}$}} &  \multirow{2}{*}{\textbf{ASR(\%)}} & \multirow{2}{*}{\textbf{Scale}} & \multicolumn{5}{c}{\textbf{Imperceptibility metric}}\\ 
        \cmidrule{4-8}
        {} &{} & {} & \textbf{SSIM ($\uparrow$)} & \textbf{UIQ ($\uparrow$)} & \textbf{SRE} ($\uparrow$) & \textbf{CLIP} ($\uparrow$) & \textbf{LPIPS} ($\downarrow$)\\

        \midrule
        \multirow{2}{*}{Ipod}  & \multirow{2}{*}{99.6} & Local & 0.95 & 0.92 & 28.9 & 86.8 & 0.115\\
        \cmidrule{3-8}
        {}  & {} & Global & 0.99 & 0.97 & 37.8 & 97.1 & 0.018 \\

        \midrule
        \multirow{2}{*}{Baseball} & \multirow{2}{*}{99.3} & Local & 0.94 & 0.91 & 28.5 & 85.0 & 0.118\\
        \cmidrule{3-8}
        {}  & {} & Global & 0.99 & 0.97 & 37.4 & 96.7 & 0.019 \\
        
        \midrule
        \multirow{2}{*}{Toaster}  & \multirow{2}{*}{99.5} & Local & 0.94 & 0.91 & 28.3 & 84.5 & 0.120\\
        \cmidrule{3-8}
        {} &{} & Global & 0.990 & 0.970 & 37.23 & 96.52 & 0.020 \\
        
        \bottomrule
    \end{tabular}
    }
    \vspace{-0.1in}
    \caption{Detailed evaluation of attack efficacy through ASR ($\%$) and imperceptibility for different target classes within the ImageNet Dataset. For SSIM, UIQ, SRE, and CLIP scores, the higher ($\uparrow$), the better, while the lower ($\downarrow$), the better for LPIPS.}
    \label{table: multiTarget results}
\end{table}

\begin{table}[t]
    \centering
    \small
    \resizebox{\linewidth}{!}{
    \begin{tabular}{cclccccc}
        \toprule
        \multirow{2}{*}{\textbf{Method}} & \multirow{2}{*}{\textbf{ASR(\%)}} & \multirow{2}{*}{\textbf{Scale}} & \multicolumn{5}{c}{\textbf{Imperceptibility metric}}\\ 
        \cmidrule{4-8}
        {} &{} & {} & \textbf{SSIM ($\uparrow$)} & \textbf{UIQ ($\uparrow$)} & \textbf{SRE} ($\uparrow$) & \textbf{CLIP} ($\uparrow$) & \textbf{LPIPS} ($\downarrow$)\\

        \midrule
        \multirow{2}{*}{Google Patch}  & \multirow{2}{*}{\textbf{100}} & Local & 0.000 & 0.000 & 11.95 & 36.82 & 0.890\\
        \cmidrule{3-8}
        {} & {} & Global & 0.812 & 0.820 & 19.46 & 68.22 & 0.270 \\

        \midrule
        \multirow{2}{*}{LaVAN}  &\multirow{2}{*}{\textbf{100}} & Local & 0.006 & 0.000 & 15.85 & 36.55 & 0.865\\
        \cmidrule{3-8}
        {} & {} & Global & 0.820 & 0.825 & 24.18 & 71.84 & 0.220 \\
       
        \midrule
        \multirow{2}{*}{GDPA}  &\multirow{2}{*}{96.12} & Local & 0.240 & 0.220 & 21.00 & 57.96 & 0.660\\
        \cmidrule{3-8}
        {} & {} & Global & 0.870 & 0.865 & 29.00 & 75.66 & 0.151 \\

        \midrule
        \multirow{2}{*}{MPGD}  & \multirow{2}{*}{88.9} & Local & 0.620 & 0.533 & 28.30 & 65.30 & 0.400\\
        \cmidrule{3-8}
        {} &{} & Global & 0.960 & 0.935 & 36.70 & 86.70 & 0.087 \\

        \midrule
        \multirow{2}{*}{Ours}  & \multirow{2}{*}{\textbf{100}} & Local & \textbf{0.930} & \textbf{0.880} & \textbf{31.81} & \textbf{66.50} & \textbf{0.207}\\
        \cmidrule{3-8}
        {} &{} & Global & \textbf{0.990} & \textbf{0.980} & \textbf{40.11} & \textbf{88.57} & \textbf{0.039} \\
        
        \bottomrule
    \end{tabular}}
    \caption{Detailed evaluation and comparison of attack efficacy through ASR ($\%$) and imperceptibility with \textbf{VGG16} as the victim model on the VGG Face dataset for the Target class \textbf{``A. J. Buckley''}. For SSIM, UIQ, SRE, and CLIP scores, the higher ($\uparrow$) the better, while the lower ($\downarrow$) the better for LPIPS.}
    \label{table: VGG16_0_results}
\end{table}

\begin{table}[t]
    \centering
    \small
    \resizebox{\linewidth}{!}{
    \begin{tabular}{cclccccc}
        \toprule
        \multirow{2}{*}{\textbf{Method}} & \multirow{2}{*}{\textbf{ASR(\%)}} & \multirow{2}{*}{\textbf{Scale}} & \multicolumn{5}{c}{\textbf{Imperceptibility metric}}\\ 
        \cmidrule{4-8}
        {} &{} & {} & \textbf{SSIM ($\uparrow$)} & \textbf{UIQ ($\uparrow$)} & \textbf{SRE} ($\uparrow$) & \textbf{CLIP} ($\uparrow$) & \textbf{LPIPS} ($\downarrow$)\\

        \midrule
        \multirow{2}{*}{Google Patch}  & \multirow{2}{*}{\textbf{99.9}} & Local & 0.000 & 0.000 & 11.76 & 36.43 & 0.860\\
        \cmidrule{3-8}
        {} & {} & Global & 0.810 & 0.820 & 19.36 & 68.22 & 0.270 \\

        \midrule
        \multirow{2}{*}{LaVAN}  &\multirow{2}{*}{99.5} & Local & 0.005 & 0.000 & 15.64 & 36.52 & 0.850\\
        \cmidrule{3-8}
        {} & {} & Global & 0.820 & 0.825 & 24.06 & 71.56 & 0.220 \\

        \midrule
        \multirow{2}{*}{GDPA}  &\multirow{2}{*}{99.50} & Local & 0.220 & 0.190 & 21.46 & 55.50 & 0.685\\
        \cmidrule{3-8}
        {} & {} & Global & 0.850 & 0.840 & 55.50 & 63.41 & 0.190 \\

        \midrule
        \multirow{2}{*}{MPGD}  & \multirow{2}{*}{86.85} & Local & 0.650 & 0.550 & 27.80 & 65.20 & 0.420\\
        \cmidrule{3-8}
        {} &{} & Global & 0.950 & 0.930 & 36.10 & 86.60 & 0.090 \\

        \midrule
        \multirow{2}{*}{Ours}  & \multirow{2}{*}{\textbf{98.8}} & Local & \textbf{0.924} & \textbf{0.870} & \textbf{31.94} & \textbf{68.24} & \textbf{0.200}\\
        \cmidrule{3-8}
        {} &{} & Global & \textbf{0.990} & \textbf{0.980} & \textbf{40.08} & \textbf{88.70} & \textbf{0.039} \\
        
        \bottomrule
    \end{tabular}}
    \caption{Detailed evaluation and comparison of attack efficacy through ASR ($\%$) and imperceptibility with \textbf{VGG16} as the victim model on the VGG Face dataset for the Target class \textbf{``Aamir Khan''}. For SSIM, UIQ, SRE, and CLIP scores, the higher ($\uparrow$) the better, while the lower ($\downarrow$) the better for LPIPS.}
    \label{table: VGG16_2_results}
\end{table}

\begin{table}[t]
    \centering
    \small
    \resizebox{\linewidth}{!}{
    \begin{tabular}{cclccccc}
        \toprule
        \multirow{2}{*}{\textbf{Method}} & \multirow{2}{*}{\textbf{ASR(\%)}} & \multirow{2}{*}{\textbf{Scale}} & \multicolumn{5}{c}{\textbf{Imperceptibility metric}}\\ 
        \cmidrule{4-8}
        {} &{} & {} & \textbf{SSIM ($\uparrow$)} & \textbf{UIQ ($\uparrow$)} & \textbf{SRE} ($\uparrow$) & \textbf{CLIP} ($\uparrow$) & \textbf{LPIPS} ($\downarrow$)\\

        \midrule
        \multirow{2}{*}{Google Patch}  & \multirow{2}{*}{\textbf{100}} & Local & 0.000 & 0.000 & 10.76 & 36.65 & 0.860\\
        \cmidrule{3-8}
        {} & {} & Global & 0.810 & 0.820 & 18.27 & 68.70 & 0.290 \\

        \midrule
        \multirow{2}{*}{LaVAN}  &\multirow{2}{*}{\textbf{100}} & Local & 0.003 & 0.000 & 11.89 & 36.45 & 0.870\\
        \cmidrule{3-8}
        {} & {} & Global & 0.820 & 0.824 & 20.30 & 71.67 & 0.260 \\

        \midrule
        \multirow{2}{*}{GDPA}  &\multirow{2}{*}{91.50} & Local & 0.476 & 0.465 & 22.85 & 60.48 & 0.53\\
        \cmidrule{3-8}
        {} & {} & Global & 0.900 & 0.890 & 29.45 & 76.00 & 0.125 \\

        \midrule
        \multirow{2}{*}{MPGD}  & \multirow{2}{*}{84.95} & Local & 0.680 & 0.564 & 27.30 & 65.10 & 0.440\\
        \cmidrule{3-8}
        {} &{} & Global & 0.940 & 0.924 & 35.88 & 85.10 & 0.094 \\

        \midrule
        \multirow{2}{*}{Ours}  & \multirow{2}{*}{\textbf{99.53}} & Local & \textbf{0.904} & \textbf{0.850} & \textbf{31.40} & \textbf{65.80} & \textbf{0.217}\\
        \cmidrule{3-8}
        {} &{} & Global & \textbf{0.985} & \textbf{0.980} & \textbf{39.61} & \textbf{87.72} & \textbf{0.042} \\
        
        \bottomrule
    \end{tabular}}
    \caption{Detailed evaluation and comparison of attack efficacy through ASR ($\%$) and imperceptibility with \textbf{VGG16} as the victim model on the VGG Face dataset for the Target class \textbf{``Aaron Staton''}. For SSIM, UIQ, SRE, and CLIP scores, the higher ($\uparrow$) the better, while the lower ($\downarrow$) the better for LPIPS.}
   \label{table: VGG16_3_results} 
\end{table}

\begin{table}[t]
    \centering
    \small
    \resizebox{\linewidth}{!}{
    \begin{tabular}{cclccccc}
        \toprule
        \multirow{2}{*}{\textbf{Method}} & \multirow{2}{*}{\textbf{ASR(\%)}} & \multirow{2}{*}{\textbf{Scale}} & \multicolumn{5}{c}{\textbf{Imperceptibility metric}}\\ 
        \cmidrule{4-8}
        {} &{} & {} & \textbf{SSIM ($\uparrow$)} & \textbf{UIQ ($\uparrow$)} & \textbf{SRE} ($\uparrow$) & \textbf{CLIP} ($\uparrow$) & \textbf{LPIPS} ($\downarrow$)\\

        \midrule
        \multirow{2}{*}{Google Patch}  & \multirow{2}{*}{98.0} & Local & 0.010 & 0.000 & 17.52 & 38.81 & 0.730\\
        \cmidrule{3-8}
        {} & {} & Global & 0.830 & 0.820 & 24.25 & 63.13 & 0.210 \\

        \midrule
        \multirow{2}{*}{LaVAN}  &\multirow{2}{*}{\textbf{100}} & Local & 0.007 & 0.000 & 16.80 & 36.81 & 0.840\\
        \cmidrule{3-8}
        {} & {} & Global & 0.840 & 0.826 & 25.12 & 71.64 & 0.200 \\

        \midrule
        \multirow{2}{*}{GDPA}  &\multirow{2}{*}{99.5} & Local & 0.310 & 0.250 & 22.00 & 53.00 & 0.660\\
        \cmidrule{3-8}
        {} & {} & Global & 0.880 & 0.860 & 29.00 & 59.00 & 0.170 \\

        \midrule
        \multirow{2}{*}{MPGD}  & \multirow{2}{*}{78.1} & Local & 0.620 & 0.560 & 26.99 & 61.78 & 0.380\\
        \cmidrule{3-8}
        {} &{} & Global & 0.950 & 0.930 & 35.56 & 85.42 & 0.080 \\

        \midrule
        \multirow{2}{*}{Ours}  & \multirow{2}{*}{\textbf{98.8}} & Local & \textbf{0.920} & \textbf{0.880} & \textbf{32.11} & \textbf{69.40} & \textbf{0.170}\\
        \cmidrule{3-8}
        {} &{} & Global & \textbf{0.990} & \textbf{0.980} & \textbf{40.66} & \textbf{90.55} & \textbf{0.030} \\
        
        \bottomrule
    \end{tabular}}
    \caption{Detailed evaluation and comparison of attack efficacy through ASR ($\%$) and imperceptibility with \textbf{ResNet-50} as the victim model on the VGG Face dataset for the Target class \textbf{``A. J. Buckley''}. For SSIM, UIQ, SRE, and CLIP scores, the higher ($\uparrow$) the better, while the lower ($\downarrow$) the better for LPIPS.}
    
    \label{table: resnet_0_results}    
\end{table}

\begin{table}[t]
    \centering
    \small
    \resizebox{\linewidth}{!}{
    \begin{tabular}{cclccccc}
        \toprule
        \multirow{2}{*}{\textbf{Method}} & \multirow{2}{*}{\textbf{ASR(\%)}} & \multirow{2}{*}{\textbf{Scale}} & \multicolumn{5}{c}{\textbf{Imperceptibility metric}}\\ 
        \cmidrule{4-8}
        {} &{} & {} & \textbf{SSIM ($\uparrow$)} & \textbf{UIQ ($\uparrow$)} & \textbf{SRE} ($\uparrow$) & \textbf{CLIP} ($\uparrow$) & \textbf{LPIPS} ($\downarrow$)\\

        \midrule
        \multirow{2}{*}{Google Patch}  & \multirow{2}{*}{99.5} & Local & 0.001 & 0.000 & 16.47 & 38.80 & 0.800\\
        \cmidrule{3-8}
        {} & {} & Global & 0.830 & 0.820 & 21.89 & 63.13 & 0.270 \\

        \midrule
        \multirow{2}{*}{LaVAN}  &\multirow{2}{*}{\textbf{100}} & Local & 0.007 & 0.000 & 16.89 & 36.82 & 0.830\\
        \cmidrule{3-8}
        {} & {} & Global & 0.840 & 0.826 & 25.30 & 71.51 & 0.210 \\

        \midrule
        \multirow{2}{*}{GDPA}  &\multirow{2}{*}{99.70} & Local & 0.280 & 0.230 & 21.99 & 56.73 & 0.600\\
        \cmidrule{3-8}
        {} & {} & Global & 0.870 & 0.850 & 56.73 & 59.32 & 0.200 \\

        \midrule
        \multirow{2}{*}{MPGD }  & \multirow{2}{*}{70.74} & Local & 0.610 & 0.550 & 26.60 & 59.87 & 0.390\\
        \cmidrule{3-8}
        {} &{} & Global & 0.940 & 0.930 & 35.30 & 84.63 & 0.080 \\

        \midrule
        \multirow{2}{*}{Ours}  & \multirow{2}{*}{\textbf{93.0}} & Local & \textbf{0.890} & \textbf{0.830} & \textbf{30.88} & \textbf{65.75} & \textbf{0.226}\\
        \cmidrule{3-8}
        {} &{} & Global & \textbf{0.980} & \textbf{0.970} & \textbf{39.37} & \textbf{87.30} & \textbf{0.040} \\
        
        \bottomrule
    \end{tabular}}
    \caption{Detailed evaluation and comparison of attack efficacy through ASR ($\%$) and imperceptibility with \textbf{ResNet-50} as the victim model on the VGG Face dataset for the Target class \textbf{``Aamir Khan''}. For SSIM, UIQ, SRE, and CLIP scores, the higher ($\uparrow$) the better, while the lower ($\downarrow$) the better for LPIPS.}
    
    \label{table: resnet_2_results}    
\end{table}

\begin{table}[t]
    \centering
    \small
    \resizebox{\linewidth}{!}{
    \begin{tabular}{cclccccc}
        \toprule
        \multirow{2}{*}{\textbf{Method}} & \multirow{2}{*}{\textbf{ASR(\%)}} & \multirow{2}{*}{\textbf{Scale}} & \multicolumn{5}{c}{\textbf{Imperceptibility metric}}\\ 
        \cmidrule{4-8}
        {} &{} & {} & \textbf{SSIM ($\uparrow$)} & \textbf{UIQ ($\uparrow$)} & \textbf{SRE} ($\uparrow$) & \textbf{CLIP} ($\uparrow$) & \textbf{LPIPS} ($\downarrow$)\\

        \midrule
        \multirow{2}{*}{Google Patch}  & \multirow{2}{*}{80.3} & Local & 0.010 & 0.000 & 17.52 & 38.81 & 0.730\\
        \cmidrule{3-8}
        {} & {} & Global & 0.830 & 0.820 & 24.25 & 63.13 & 0.210 \\

        \midrule
        \multirow{2}{*}{LaVAN}  &\multirow{2}{*}{\textbf{97.0}} & Local & 0.010 & 0.000 & 17.45 & 41.54 & 0.750\\
        \cmidrule{3-8}
        {} & {} & Global & 0.830 & 0.820 & 22.32 & 62.68 & 0.240 \\

        \midrule
        \multirow{2}{*}{GDPA}  &\multirow{2}{*}{98.00} & Local & 0.330 & 0.280 & 22.10 & 55.68 & 0.60\\
        \cmidrule{3-8}
        {} & {} & Global & 0.880 & 0.850 & 29.12 & 57.54 & 0.200 \\

        \midrule
        \multirow{2}{*}{MPGD }  & \multirow{2}{*}{52.50} & Local & 0.610 & 0.550 & 26.83 & 60.25 & 0.380\\
        \cmidrule{3-8}
        {} &{} & Global & 0.940 & 0.930 & 35.25 & 83.42 & 0.080 \\

        \midrule
        \multirow{2}{*}{Ours}  & \multirow{2}{*}{\textbf{91.80}} & Local & \textbf{0.890} & \textbf{0.840} & \textbf{30.89} & \textbf{65.70} & \textbf{0.216}\\
        \cmidrule{3-8}
        {} &{} & Global & \textbf{0.980} & \textbf{0.970} & \textbf{39.33} & \textbf{88.32} & \textbf{0.040} \\
        
        \bottomrule
    \end{tabular}}
    \caption{Detailed evaluation and comparison of attack efficacy through ASR ($\%$) and imperceptibility with \textbf{ResNet-50} as the victim model on the VGG Face dataset for the Target class \textbf{``Aaron Staton''}. For SSIM, UIQ, SRE, and CLIP scores, the higher ($\uparrow$) the better, while the lower ($\downarrow$) the better for LPIPS.}
    
    \label{table: resnet_3_results}
\end{table}

\begin{table}[t]
    \centering
    \small
    \resizebox{\linewidth}{!}{
    \begin{tabular}{cclccccc}
        \toprule
        \multirow{2}{*}{\textbf{Method}} & \multirow{2}{*}{\textbf{ASR(\%)}} & \multirow{2}{*}{\textbf{Scale}} & \multicolumn{5}{c}{\textbf{Imperceptibility metric}}\\ 
        \cmidrule{4-8}
        {} &{} & {} & \textbf{SSIM ($\uparrow$)} & \textbf{UIQ ($\uparrow$)} & \textbf{SRE} ($\uparrow$) & \textbf{CLIP} ($\uparrow$) & \textbf{LPIPS} ($\downarrow$)\\

        \midrule
        \multirow{2}{*}{Google Patch}  & \multirow{2}{*}{98.9} & Local & 0.040 & 0.000 & 10.12 & 36.10 & 0.820\\
        \cmidrule{3-8}
        {} & {} & Global & 0.830 & 0.830 & 16.87 & 66.88 & 0.260 \\

        \midrule
        \multirow{2}{*}{LaVAN}  &\multirow{2}{*}{\textbf{100}} & Local & 0.007 & 0.000 & 16.49 & 36.50 & 0.850\\
        \cmidrule{3-8}
        {} & {} & Global & 0.840 & 0.825 & 24.75 & 71.87 & 0.210 \\

        \midrule
        \multirow{2}{*}{GDPA}  &\multirow{2}{*}{92.9} & Local & 0.330 & 0.270 & 21.85 & 62.10 & 0.570\\
        \cmidrule{3-8}
        {} & {} & Global & 0.880 & 0.870 & 29.30 & 71.76 & 0.140 \\

        \midrule
        \multirow{2}{*}{MPGD }  & \multirow{2}{*}{95.5} & Local & 0.630 & 0.540 & 27.65 & 62.48 & 0.380\\
        \cmidrule{3-8}
        {} &{} & Global & 0.950 & 0.930 & 35.72 & 86.66 & 0.070 \\

        \midrule
        \multirow{2}{*}{Ours}  & \multirow{2}{*}{\textbf{99.3}} & Local & \textbf{0.860} & \textbf{0.800} & \textbf{29.22} & \textbf{63.28} & \textbf{0.275}\\
        \cmidrule{3-8}
        {} &{} & Global & \textbf{0.980} & \textbf{0.970} & \textbf{38.00} & \textbf{87.83} & \textbf{0.048} \\
        
        \bottomrule
    \end{tabular}}
    \caption{Detailed evaluation and comparison of attack efficacy through ASR ($\%$) and imperceptibility with \textbf{Swin Transformer Tiny} as the victim model on the VGG Face dataset for the Target class \textbf{``A. J. Buckley''}. For SSIM, UIQ, SRE, and CLIP scores, the higher ($\uparrow$) the better, while the lower ($\downarrow$) the better for LPIPS.}
    
    \label{table: swint_0_results}    
\end{table}

\begin{table}[t]
    \centering
    \small
    \resizebox{\linewidth}{!}{
    \begin{tabular}{cclccccc}
        \toprule
        \multirow{2}{*}{\textbf{Method}} & \multirow{2}{*}{\textbf{ASR(\%)}} & \multirow{2}{*}{\textbf{Scale}} & \multicolumn{5}{c}{\textbf{Imperceptibility metric}}\\ 
        \cmidrule{4-8}
        {} &{} & {} & \textbf{SSIM ($\uparrow$)} & \textbf{UIQ ($\uparrow$)} & \textbf{SRE} ($\uparrow$) & \textbf{CLIP} ($\uparrow$) & \textbf{LPIPS} ($\downarrow$)\\

        \midrule
        \multirow{2}{*}{Google Patch}  & \multirow{2}{*}{99.2} & Local & 0.000 & 0.000 & 10.11 & 37.23 & 0.780\\
        \cmidrule{3-8}
        {} & {} & Global & 0.830 & 0.820 & 17.22 & 67.50 & 0.230 \\

        \midrule
        \multirow{2}{*}{LaVAN}  &\multirow{2}{*}{\textbf{100}} & Local & 0.006 & 0.000 & 16.31 & 36.57 & 0.850\\
        \cmidrule{3-8}
        {} & {} & Global & 0.840 & 0.825 & 24.71 & 71.73 & 0.210 \\

        \midrule
        \multirow{2}{*}{GDPA}  &\multirow{2}{*}{100} & Local & 0.340 & 0.300 & 19.85 & 60.84 & 0.600\\
        \cmidrule{3-8}
        {} & {} & Global & 0.910 & 0.910 & 29.82 & 80.01 & 0.100 \\

        \midrule
        \multirow{2}{*}{MPGD }  & \multirow{2}{*}{94.87} & Local & 0.640 & 0.550 & 27.68 & 62.69 & 0.370\\
        \cmidrule{3-8}
        {} &{} & Global & 0.950 & 0.930 & 35.80 & 86.97 & 0.070 \\

        \midrule
        \multirow{2}{*}{Ours}  & \multirow{2}{*}{\textbf{99.3}} & Local & \textbf{0.870} & \textbf{0.820} & \textbf{29.80} & \textbf{63.00} & \textbf{0.240}\\
        \cmidrule{3-8}
        {} &{} & Global & \textbf{0.980} & \textbf{0.970} & \textbf{38.60} & \textbf{88.20} & \textbf{0.043} \\
        
        \bottomrule
    \end{tabular}}
    \caption{Detailed evaluation and comparison of attack efficacy through ASR ($\%$) and imperceptibility with \textbf{Swin Transformer Tiny} as the victim model on the VGG Face dataset for the Target class \textbf{``Aamir Khan''}. For SSIM, UIQ, SRE, and CLIP scores, the higher ($\uparrow$) the better, while the lower ($\downarrow$) the better for LPIPS.}
    
    \label{table: swint_2_results}    
\end{table}

\begin{table}[t]
    \centering
    \small
    \resizebox{\linewidth}{!}{
    \begin{tabular}{cclccccc}
        \toprule
        \multirow{2}{*}{\textbf{Method}} & \multirow{2}{*}{\textbf{ASR(\%)}} & \multirow{2}{*}{\textbf{Scale}} & \multicolumn{5}{c}{\textbf{Imperceptibility metric}}\\ 
        \cmidrule{4-8}
        {} &{} & {} & \textbf{SSIM ($\uparrow$)} & \textbf{UIQ ($\uparrow$)} & \textbf{SRE} ($\uparrow$) & \textbf{CLIP} ($\uparrow$) & \textbf{LPIPS} ($\downarrow$)\\

        \midrule
        \multirow{2}{*}{Google Patch}  & \multirow{2}{*}{99.3} & Local & 0.000 & 0.000 & 12.60 & 38.84 & 0.820\\
        \cmidrule{3-8}
        {} & {} & Global & 0.830 & 0.820 & 17.48 & 63.13 & 0.290 \\

        \midrule
        \multirow{2}{*}{LaVAN}  &\multirow{2}{*}{\textbf{100}} & Local & 0.007 & 0.000 & 16.45 & 36.67 & 0.850\\
        \cmidrule{3-8}
        {} & {} & Global & 0.840 & 0.825 & 24.82 & 72.06 & 0.210 \\

        \midrule
        \multirow{2}{*}{GDPA}  &\multirow{2}{*}{92.4} & Local & 0.310 & 0.260 & 20.19 & 54.86 & 0.65\\
        \cmidrule{3-8}
        {} & {} & Global & 0.860 & 0.840 & 27.21 & 60.54 & 0.220 \\

        \midrule
        \multirow{2}{*}{MPGD}  & \multirow{2}{*}{96.2} & Local & 0.640 & 0.550 & 27.76 & 61.90 & 0.360\\
        \cmidrule{3-8}
        {} &{} & Global & 0.950 & 0.930 & 35.85 & 87.30 & 0.070 \\

        \midrule
        \multirow{2}{*}{Ours}  & \multirow{2}{*}{\textbf{98.60}} & Local & \textbf{0.860} & \textbf{0.800} & \textbf{29.64} & \textbf{62.00} & \textbf{0.260}\\
        \cmidrule{3-8}
        {} &{} & Global & \textbf{0.980} & \textbf{0.970} & \textbf{38.34} & \textbf{87.90} & \textbf{0.046} \\
        
        \bottomrule
    \end{tabular}}
    \caption{Detailed evaluation and comparison of attack efficacy through ASR ($\%$) and imperceptibility with \textbf{Swin Transformer Tiny} as the victim model on the VGG Face dataset for the Target class \textbf{``Aaron Staton''}. For SSIM, UIQ, SRE, and CLIP scores, the higher ($\uparrow$) the better, while the lower ($\downarrow$) the better for LPIPS.}
    
    \label{table: swint_3_results}    
\end{table}

\begin{table}[t]
    \centering
    \small
    \resizebox{\linewidth}{!}{
    \begin{tabular}{cclccccc}
        \toprule
        \multirow{2}{*}{\textbf{Method}} & \multirow{2}{*}{\textbf{ASR(\%)}} & \multirow{2}{*}{\textbf{Scale}} & \multicolumn{5}{c}{\textbf{Imperceptibility metric}}\\ 
        \cmidrule{4-8}
        {} &{} & {} & \textbf{SSIM ($\uparrow$)} & \textbf{UIQ ($\uparrow$)} & \textbf{SRE} ($\uparrow$) & \textbf{CLIP} ($\uparrow$) & \textbf{LPIPS} ($\downarrow$)\\

        \midrule
        \multirow{2}{*}{Google Patch}  & \multirow{2}{*}{98.2} & Local & 0.000 & 0.000 & 11.23 & 36.51 & 0.835\\
        \cmidrule{3-8}
        {} & {} & Global & 0.830 & 0.820 & 18.23 & 67.65 & 0.240 \\

        \midrule
        \multirow{2}{*}{LaVAN}  &\multirow{2}{*}{\textbf{100}} & Local & 0.005 & 0.000 & 15.47 & 36.52 & 0.850\\
        \cmidrule{3-8}
        {} & {} & Global & 0.840 & 0.825 & 23.80 & 71.82 & 0.220 \\

        \midrule
        \multirow{2}{*}{GDPA}  &\multirow{2}{*}{77.24} & Local & 0.410 & 0.360 & 21.59 & 58.14 & 0.56\\
        \cmidrule{3-8}
        {} & {} & Global & 0.910 & 0.900 & 29.66 & 72.23 & 0.110 \\

        \midrule
        \multirow{2}{*}{MPGD}  & \multirow{2}{*}{97.9} & Local & 0.600 & 0.520 & 27.45 & 61.22 & 0.390\\
        \cmidrule{3-8}
        {} &{} & Global & 0.940 & 0.920 & 35.57 & 85.00 & 0.080 \\

        \midrule
        \multirow{2}{*}{Ours}  & \multirow{2}{*}{\textbf{99.0}} & Local & \textbf{0.860} & \textbf{0.780} & \textbf{29.8} & \textbf{63.00} & \textbf{0.300}\\
        \cmidrule{3-8}
        {} &{} & Global & \textbf{0.980} & \textbf{0.960} & \textbf{38.10} & \textbf{86.00} & \textbf{0.055} \\
        
        \bottomrule
    \end{tabular}}
    \caption{Detailed evaluation and comparison of attack efficacy through ASR ($\%$) and imperceptibility with \textbf{Swin Transformer Base} as the victim model on the VGG Face dataset for the Target class \textbf{``A. J. Buckley''}. For SSIM, UIQ, SRE, and CLIP scores, the higher ($\uparrow$) the better, while the lower ($\downarrow$) the better for LPIPS.}
    
    \label{table: swinb_0_results}    
\end{table}

\begin{table}[t]
    \centering
    \small
    \resizebox{\linewidth}{!}{
    \begin{tabular}{cclccccc}
        \toprule
        \multirow{2}{*}{\textbf{Method}} & \multirow{2}{*}{\textbf{ASR(\%)}} & \multirow{2}{*}{\textbf{Scale}} & \multicolumn{5}{c}{\textbf{Imperceptibility metric}}\\ 
        \cmidrule{4-8}
        {} &{} & {} & \textbf{SSIM ($\uparrow$)} & \textbf{UIQ ($\uparrow$)} & \textbf{SRE} ($\uparrow$) & \textbf{CLIP} ($\uparrow$) & \textbf{LPIPS} ($\downarrow$)\\

        \midrule
        \multirow{2}{*}{Google Patch}  & \multirow{2}{*}{97.2} & Local & 0.000 & 0.000 & 10.78 & 36.85 & 0.900\\
        \cmidrule{3-8}
        {} & {} & Global & 0.830 & 0.820 & 18.10 & 69.36 & 0.260 \\

        \midrule
        \multirow{2}{*}{LaVAN}  &\multirow{2}{*}{\textbf{99.3}} & Local & 0.004 & 0.000 & 15.00 & 36.49 & 0.850\\
        \cmidrule{3-8}
        {} & {} & Global & 0.840 & 0.824 & 23.40 & 71.68 & 0.220 \\

        \midrule
        \multirow{2}{*}{GDPA}  &\multirow{2}{*}{55.1} & Local & 0.160 & 0.140 & 18.36 & 65.87 & 0.700\\
        \cmidrule{3-8}
        {} & {} & Global & 0.920 & 0.920 & 30.10 & 84.10 & 0.090 \\

        \midrule
        \multirow{2}{*}{MPGD }  & \multirow{2}{*}{80.81} & Local & 0.610 & 0.530 & 27.35 & 61.22 & 0.390\\
        \cmidrule{3-8}
        {} &{} & Global & 0.940 & 0.930 & 35.47 & 85.28 & 0.080 \\

        \midrule
        \multirow{2}{*}{Ours}  & \multirow{2}{*}{\textbf{97.0}} & Local & \textbf{0.840} & \textbf{0.760} & \textbf{29.63} & \textbf{61.00} & \textbf{0.300}\\
        \cmidrule{3-8}
        {} &{} & Global & \textbf{0.970} & \textbf{0.960} & \textbf{37.84} & \textbf{86.00} & \textbf{0.055} \\
        
        \bottomrule
    \end{tabular}}
    \caption{Detailed evaluation and comparison of attack efficacy through ASR ($\%$) and imperceptibility with \textbf{Swin Transformer Base} as the victim model on the VGG Face dataset for the Target class \textbf{``Aamir Khan''}. For SSIM, UIQ, SRE, and CLIP scores, the higher ($\uparrow$) the better, while the lower ($\downarrow$) the better for LPIPS.}
    \label{table: swinb_2_results}    
\end{table}

\begin{table}[t]
    \centering
    \small
    \resizebox{\linewidth}{!}{
    \begin{tabular}{cclccccc}
        \toprule
        \multirow{2}{*}{\textbf{Method}} & \multirow{2}{*}{\textbf{ASR(\%)}} & \multirow{2}{*}{\textbf{Scale}} & \multicolumn{5}{c}{\textbf{Imperceptibility metric}}\\ 
        \cmidrule{4-8}
        {} &{} & {} & \textbf{SSIM ($\uparrow$)} & \textbf{UIQ ($\uparrow$)} & \textbf{SRE} ($\uparrow$) & \textbf{CLIP} ($\uparrow$) & \textbf{LPIPS} ($\downarrow$)\\

        \midrule
        \multirow{2}{*}{Google Patch}  & \multirow{2}{*}{98.3} & Local & 0.000 & 0.000 & 11.86 & 35.58 & 0.920\\
        \cmidrule{3-8}
        {} & {} & Global & 0.830 & 0.820 & 17.78 & 68.63 & 0.280 \\

        \midrule
        \multirow{2}{*}{LaVAN}  &\multirow{2}{*}{\textbf{99.8}} & Local & 0.006 & 0.000 & 15.73 & 36.70 & 0.850\\
        \cmidrule{3-8}
        {} & {} & Global & 0.840 & 0.825 & 24.12 & 71.96 & 0.210 \\

        \midrule
        \multirow{2}{*}{GDPA}  &\multirow{2}{*}{84.9} & Local & 0.290 & 0.260 & 19.76 & 57.25 & 0.620\\
        \cmidrule{3-8}
        {} & {} & Global & 0.910 & 0.910 & 29.72 & 78.20 & 0.100 \\

        \midrule
        \multirow{2}{*}{MPGD}  & \multirow{2}{*}{94.9} & Local & 0.630 & 0.550 & 27.75 & 61.50 & 0.370\\
        \cmidrule{3-8}
        {} &{} & Global & 0.950 & 0.930 & 35.84 & 86.54 & 0.070 \\

        \midrule
        \multirow{2}{*}{Ours}  & \multirow{2}{*}{\textbf{98.6}} & Local & \textbf{0.880} & \textbf{0.810} & \textbf{30.50} & \textbf{63.50} & \textbf{0.250}\\
        \cmidrule{3-8}
        {} &{} & Global & \textbf{0.980} & \textbf{0.970} & \textbf{38.84} & \textbf{88.40} & \textbf{0.045} \\
        
        \bottomrule
    \end{tabular}}
    \caption{Detailed evaluation and comparison of attack efficacy through ASR ($\%$) and imperceptibility with \textbf{Swin Transformer Base} as the victim model on the VGG Face dataset for the Target class \textbf{``Aaron Staton''}. For SSIM, UIQ, SRE, and CLIP scores, the higher ($\uparrow$) the better, while the lower ($\downarrow$) the better for LPIPS.}
    
    \label{table: swinb_3_results}    
\end{table}

\begin{table}[t]
    \centering
    \small
    \resizebox{\linewidth}{!}{
    \begin{tabular}{cclccccc}
        \toprule
        \multirow{2}{*}{\textbf{Patch Size(\%)}} & \multirow{2}{*}{\textbf{ASR(\%)}} & \multirow{2}{*}{\textbf{Scale}} & \multicolumn{5}{c}{\textbf{Imperceptibility metric}}\\ 
        \cmidrule{4-8}
        {} &{} & {} & \textbf{SSIM ($\uparrow$)} & \textbf{UIQ ($\uparrow$)} & \textbf{SRE} ($\uparrow$) & \textbf{CLIP} ($\uparrow$) & \textbf{LPIPS} ($\downarrow$)\\

        \midrule
        \multirow{2}{*}{2}  & \multirow{2}{*}{72.2} & Local & 0.640 & 0.530 & 21.07 & 70.00 & 0.413\\
        \cmidrule{3-8}
        {} & {} & Global & 0.992 & 0.985 & 38.10 & 98.20 & 0.014 \\

        \midrule
        \multirow{2}{*}{4}  &\multirow{2}{*}{90.7} & Local & 0.784 & 0.683 & 23.32 & 70.77 & 0.308\\
        \cmidrule{3-8}
        {} & {} & Global & 0.991 & 0.972 & 37.68 & 98.11 & 0.014 \\
        
        \midrule
        \multirow{2}{*}{6}  &\multirow{2}{*}{94.2} & Local & 0.854 & 0.756 & 25.02 & 74.74 & 0.024\\
        \cmidrule{3-8}
        {} & {} & Global & 0.991 & 0.970 & 37.86 & 98.18 & 0.013 \\

        \midrule
        \multirow{2}{*}{8}  & \multirow{2}{*}{97.3} & Local & 0.896 & 0.810 & 26.77 & 78.05 & 0.183\\
        \cmidrule{3-8}
        {} &{} & Global & 0.991 & 0.970 & 38.14 & 98.15 & 0.012 \\

        \midrule
        \multirow{2}{*}{10}  & \multirow{2}{*}{98.1} & Local & 0.920 & 0.840 & 27.90 & 80.91 & 0.152\\
        \cmidrule{3-8}
        {} &{} & Global & 0.992 & 0.970 & 38.31 & 97.97 & 0.011 \\

        \midrule
        \multirow{2}{*}{12}  & \multirow{2}{*}{99.0} & Local & 0.934 & 0.860 & 28.90 & 83.43 & 0.126\\
        \cmidrule{3-8}
        {} &{} & Global & 0.992 & 0.965 & 38.46 & 98.03 & 0.011 \\
        
        \midrule
        \multirow{2}{*}{14}  & \multirow{2}{*}{\textbf{99.4}} & Local & \textbf{0.970} & \textbf{0.910} & \textbf{31.30} & \textbf{89.33} & \textbf{0.070}\\
        \cmidrule{3-8}
        {} &{} & Global & \textbf{0.994} & \textbf{0.970} & \textbf{40.10} & \textbf{98.43} & \textbf{0.010} \\
        
        \bottomrule
    \end{tabular}}
    \caption{Impact of \textbf{patch size} on attack performance represented through ASR ($\%$) and imperceptibility with \textbf{Swin Transformer Base} as the victim model on the ImageNet dataset. For SSIM, UIQ, SRE, and CLIP scores, the higher ($\uparrow$) the better, while the lower ($\downarrow$) the better for LPIPS.}
    
    \label{table: PSabla}
\end{table}

\begin{table}[t]
    \centering
    \small
    \resizebox{\linewidth}{!}{
    \begin{tabular}{cclccccc}
        \toprule
        \multirow{2}{*}{\textbf{w$_3$}} & \multirow{2}{*}{\textbf{ASR(\%)}} & \multirow{2}{*}{\textbf{Scale}} & \multicolumn{5}{c}{\textbf{Imperceptibility metric}}\\ 
        \cmidrule{4-8}
        {} &{} & {} & \textbf{SSIM ($\uparrow$)} & \textbf{UIQ ($\uparrow$)} & \textbf{SRE} ($\uparrow$) & \textbf{CLIP} ($\uparrow$) & \textbf{LPIPS} ($\downarrow$)\\

        \midrule
        \multirow{2}{*}{0}  & \multirow{2}{*}{99.0} & Local & 0.943 & 0.873 & 29.76 & 85.54 & 0.111\\
        \cmidrule{3-8}
        {} & {} & Global & 0.992 & 0.964 & 38.59 & 97.95 & 0.017 \\

        \midrule
        \multirow{2}{*}{1}  &\multirow{2}{*}{98.9} & Local & 0.944 & 0.874 & 29.79 & 85.65 & 0.110\\
        \cmidrule{3-8}
        {} & {} & Global & 0.992 & 0.965 & 38.62 & 97.97 & 0.017 \\
        
        \midrule
        \multirow{2}{*}{4}  &\multirow{2}{*}{98.9} & Local & 0.945 & 0.875 & 29.83 & 85.66 & 0.109\\
        \cmidrule{3-8}
        {} & {} & Global & 0.992 & 0.965 & 38.67 & 97.99 & 0.017 \\

        \midrule
        \multirow{2}{*}{7}  & \multirow{2}{*}{98.8} & Local & 0.946 & 0.876 & 29.84 & 85.68 & 0.108\\
        \cmidrule{3-8}
        {} &{} & Global & 0.992 & 0.966 & 38.69 & 98.01 & 0.016 \\

        \midrule
        \multirow{2}{*}{10}  & \multirow{2}{*}{99.1} & Local & 0.945 & 0.875 & 29.83 & 85.71 & 0.108\\
        \cmidrule{3-8}
        {} &{} & Global & 0.992 & 0.965 & 38.66 & 97.98 & 0.017 \\

        \midrule
        \multirow{2}{*}{13}  & \multirow{2}{*}{99.0} & Local & 0.944 & 0.874 & 29.78 & 85.56 & 0.110\\
        \cmidrule{3-8}
        {} &{} & Global & 0.992 & 0.965 & 38.60 & 97.98 & 0.017 \\
        
        \bottomrule
    \end{tabular}}
    \caption{Impact of distance term regularization coefficient $w_3$ on attack performance represented through ASR ($\%$) and imperceptibility with \textbf{Swin Transformer Base} as the victim model on the ImageNet dataset. For SSIM, UIQ, SRE, and CLIP scores, the higher ($\uparrow$) the better, while the lower ($\downarrow$) the better for LPIPS.}

    \label{table: w3abla}
\end{table}

\begin{table}[t]
    \centering
    \small
    \resizebox{\linewidth}{!}{
    \begin{tabular}{cclccccc}
        \toprule
        \multirow{2}{*}{\textbf{Update Rule}} & \multirow{2}{*}{\textbf{ASR(\%)}} & \multirow{2}{*}{\textbf{Scale}} & \multicolumn{5}{c}{\textbf{Imperceptibility metric}}\\ 
        \cmidrule{4-8}
        {} &{} & {} & \textbf{SSIM ($\uparrow$)} & \textbf{UIQ ($\uparrow$)} & \textbf{SRE} ($\uparrow$) & \textbf{CLIP} ($\uparrow$) & \textbf{LPIPS} ($\downarrow$)\\

        \midrule
        \multirow{2}{*}{Adam}  & \multirow{2}{*}{\textbf{100}} & Local & 0.130 & 0.157 & 17.13 & 36.15 & 0.662\\
        \cmidrule{3-8}
        {} & {} & Global & 0.867 & 0.848 & 25.98 & 80.94 & 0.130 \\

        \midrule
        \multirow{2}{*}{Ours}  & \multirow{2}{*}{99.4} & Local & \textbf{0.970} & \textbf{0.910} & \textbf{31.30} & \textbf{89.33} & \textbf{0.070}\\
        \cmidrule{3-8}
        {} &{} & Global & \textbf{0.994} & \textbf{0.970} & \textbf{40.10} & \textbf{98.43} & \textbf{0.010} \\
        
        \bottomrule
    \end{tabular}}
    \caption{Impact of the \textbf{update rule} on attack performance represented through ASR ($\%$) and imperceptibility with \textbf{Swin Transformer Base} as the victim model on the ImageNet dataset. For SSIM, UIQ, SRE, and CLIP scores, the higher ($\uparrow$) the better, while the lower ($\downarrow$) the better for LPIPS.}
    
    \label{table: URabla}
\end{table}

\begin{table}[t]
    \centering
    \small
    \resizebox{\linewidth}{!}{
    \begin{tabular}{cclccccc}
        \toprule
        \multirow{2}{*}{\textbf{No. Iters}} & \multirow{2}{*}{\textbf{ASR(\%)}} & \multirow{2}{*}{\textbf{Scale}} & \multicolumn{5}{c}{\textbf{Imperceptibility metric}}\\ 
        \cmidrule{4-8}
        {} &{} & {} & \textbf{SSIM ($\uparrow$)} & \textbf{UIQ ($\uparrow$)} & \textbf{SRE} ($\uparrow$) & \textbf{CLIP} ($\uparrow$) & \textbf{LPIPS} ($\downarrow$)\\

        \midrule
        \multirow{2}{*}{500}  & \multirow{2}{*}{86.0} & Local & \textbf{0.870} & \textbf{0.770} & \textbf{25.88} & \textbf{75.87} & \textbf{0.223}\\
        \cmidrule{3-8}
        {} & {} & Global & \textbf{0.992} & \textbf{0.972} & \textbf{38.48} & \textbf{98.23} & \textbf{0.015} \\

        \midrule
        \multirow{2}{*}{1000}  &\multirow{2}{*}{94.2} & Local & 0.854 & 0.756 & 25.02 & 74.74 & 0.024\\
        \cmidrule{3-8}
        {} & {} & Global & 0.991 & 0.970 & 37.86 & 98.18 & 0.013 \\
        
        \midrule
        \multirow{2}{*}{1500}  &\multirow{2}{*}{96.2} & Local & 0.850 & 0.755 & 24.91 & 73.81 & 0.024\\
        \cmidrule{3-8}
        {} & {} & Global & 0.991 & 0.969 & 37.70 & 98.10 & 0.016 \\

        \midrule
        \multirow{2}{*}{2000}  & \multirow{2}{*}{97.3} & Local & 0.843 & 0.749 & 24.77 & 73.29 & 0.246\\
        \cmidrule{3-8}
        {} &{} & Global & 0.990 & 0.968 & 37.59 & 98.04 & 0.017 \\

        \midrule
        \multirow{2}{*}{2500}  & \multirow{2}{*}{98.0} & Local & 0.840 & 0.746 & 24.67 & 72.87 & 0.249\\
        \cmidrule{3-8}
        {} &{} & Global & 0.990 & 0.967 & 37.50 & 98.02 & 0.017 \\

        \midrule
        \multirow{2}{*}{3000}  & \multirow{2}{*}{98.5} & Local & 0.836 & 0.743 & 24.48 & 72.78 & 0.252\\
        \cmidrule{3-8}
        {} &{} & Global & 0.990 & 0.966 & 37.41 & 97.98 & 0.017 \\
        
        \midrule
        \multirow{2}{*}{3500}  & \multirow{2}{*}{\textbf{98.6}} & Local & 0.834 & 0.741 & 24.65 & 72.49 & 0.254\\
        \cmidrule{3-8}
        {} &{} & Global & 0.990 & 0.969 & 37.40 & 97.92 & 0.017 \\
        
        \bottomrule
    \end{tabular}}
    \caption{Impact of \textbf{number of update iterations} on attack performance, represented through ASR ($\%$) and imperceptibility with \textbf{Swin Transformer Base} as the victim model on the ImageNet dataset. For SSIM, UIQ, SRE, and CLIP scores, the higher ($\uparrow$) the better, while the lower ($\downarrow$) the better for LPIPS. Patch size is kept fixed at $6$\%.}

    \label{table: NIabla}
\end{table}

% \ifarxiv \clearpage \appendix \input{12_appendix} \fi

\end{document}